\documentclass[letterpaper]{article}
\usepackage{aaai}
\usepackage{times}
\usepackage{helvet}
\usepackage{courier}

\usepackage{amsmath}
\usepackage{amssymb}
\usepackage{graphicx}
\usepackage[noend]{algpseudocode}
\usepackage{algorithm}
\usepackage{bbm}
\usepackage{booktabs}
\usepackage{todonotes}
\usepackage{url}
\usepackage{amsthm}
\usepackage{subcaption}

\newcommand{\E}[2]{\mathbb E_{#1}\left[#2\right]}
\newcommand\mtiny[1]{\mbox{\tiny\ensuremath{#1}}}
\newcommand{\cov}[0]{\text{cov}}
\newcommand{\V}[2]{\mathbb V_{#1}\left[#2\right]}
\newtheorem{theorem}{Theorem}

\begin{document}
%

\title{There is no Double-Descent in Random Forests}

\author{
Sebastian Buschjäger, Katharina Morik \\
Artificial Intelligence Group, TU Dortmund, Germany \\
\{sebastian.buschjaeger, katharina.morik\}@tu-dortmund.de
}
\maketitle
\begin{abstract}
Random Forests (RFs) are among the state-of-the-art in machine learning and offer excellent performance with nearly zero parameter tuning. Remarkably, RFs seem to be impervious to overfitting even though their basic building blocks are well-known to overfit. 
Recently, a broadly received study argued that a RF exhibits a so-called double-descent curve: First, the model overfits the data in a u-shaped curve and then, once a certain model complexity is reached, it suddenly improves its performance again. 
In this paper, we challenge the notion that model capacity is the correct tool to explain the success of RF and argue that the algorithm which trains the model plays a more important role than previously thought. We show that a RF does not exhibit a double-descent curve but rather has a single descent. Hence, it does not overfit in the classic sense. We further present a RF variation that also does not overfit although its decision boundary approximates that of an overfitted DT. Similar, we show that a DT which approximates the decision boundary of a RF will still overfit. Last, we study the diversity of an ensemble as a tool the estimate its performance. To do so, we introduce Negative Correlation Forest (NCForest) which allows for precise control over the diversity in the ensemble. We show, that the diversity and the bias indeed have a crucial impact on the performance of the RF. Having too low diversity collapses the performance of the RF into a a single tree, whereas having too much diversity means that most trees do not produce correct outputs anymore. However, in-between these two extremes we find a large range of different trade-offs with all roughly equal performance. Hence, the specific trade-off between bias and diversity does not matter as long as the algorithm reaches this good trade-off regime.
\end{abstract}

\begin{figure}
\centering
\includegraphics[height=6cm,keepaspectratio]{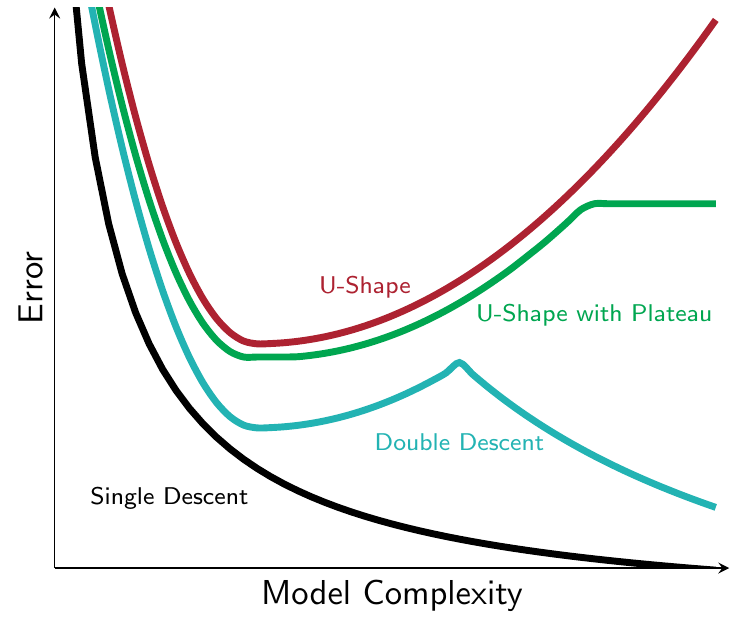}
\caption{Bias-Variance trade-off: A single descent occurs if the error decreases while the model complexity increases. A U-Shaped curved can be seen when overfitting occurs and the error increases again after a certain model complexity is reached, either with or without a plateau. A double-descent can be seen where the loss suddenly again decreases if the model complexity if further increased beyond the point of overfitting.}
\label{fig:Bias-Variance}
\end{figure}

One of the core theoretical foundations of machine learning is the bias-variance trade-off. It states that a smaller, less complex model with comparable empirical error will generalize better than a large, very complex model \cite{shalevshwartz/bendavid/2014}. Hence, one should always strive for a good balance between model complexity and empirical error. The last years of Deep Neural Network research challenged this widely accepted notion by using larger and larger models with more and more parameters that seemingly do not overfit. One particular remarkable observation is that DNNs seem to exhibit a double U-shaped curved sometimes dubbed `double descent` (see Figure \ref{fig:Bias-Variance}): With increasing model complexity, the training error approaches 0 quickly and the test-error also shrinks up to a point where overfitting starts and the test-error rises again. However, further increasing the model complexity, e.g. beyond what is reasonable given the available training data suddenly leads to a drop in test-error again. 

The reasons for this behavior are not entirely discovered yet and the question is, whether SGD-like learning algorithms or the model architecture of Deep nets are the decisive factor for this phenomenon \cite{zhang/etal/2021}. Similar, we may ask if such a phenomenon exists for other, non Deep Learning models and other learning algorithms such as Random Forests. RFs are known to be extremely resilient against overfitting with nearly no parameter tuning (see e.g. \cite{oshiro/etal/2012}) and at the same time behave better when the individual trees overfit their bootstrap samples as much as possible \cite{Breiman/2000,biau/2012,denil/2014,biau/Scornet/2016}. 
Belkin et al. showed in \shortcite{belkin/etal/2019} that there seems to be a universal double descent across different types of models. In particular, the authors show that RFs also exhibit a double-descent which could explain the miraculous performance of RFs in practice. 
We found that the discussion of the double-descent phenomenon is not properly placed in the context of existing research. In particular, the authors suggest to use the \emph{total} number of decision nodes in the Random Forests as their complexity measure, whereas learning theory implies that the \emph{average} number of decision nodes in the forest would be a more appropriate measure.
Similarly, training over-parameterized decision trees is virtually impossible because we cannot train larger trees then what we have data available. Last, the diversity among the trees inside the ensemble has long been cited as one of the corner-stones of a RF's good performance and hence cannot be ignored in this discussion.

We emphasize that we couple existing theoretical knowledge with empirical evidence and leave a more thorough theoretical analysis of this subject for future research. This paper is outlined as follows:

\begin{itemize}
    \item Section \ref{sec:Background} introduces our notation, some theoretical background of tree ensembles and the RF algorithm.
    \item In section \ref{sec:complexity} we discuss how the complexity of a RF is measured in terms of its Rademacher complexity. We present experiments which show that a RF has a single-descent curve rather than a double-descent whereas DTs overfit in a `classical' u-shaped curve. We discuss under which circumstances a DT overfits with a regular u-shaped curve and in which circumstances its overfitting plateaus. 
    \item Section \ref{sec:data-augmentation} presents experiments which show that the Rademacher complexity is not an appropriate estimator for the performance of a forest. In particular we present a variation of the RF algorithm that approximates the decision boundary of an overfitted DT, but itself does \emph{not} overfit. Similar, we show that a DT which approximates a good, not overfitted RF still overfits.
    \item In section \ref{sec:bias-variance} we study the diversity of tree ensembles as an estimators for a forests performance. We present the Negative Correlation Forest (NCForest) algorithm that allows us to smoothly interpolate between different level of diversity among the trees. We show that -- given a sufficiently small bias -- the diversity of the forest can be an accurate predictor of the forests generalization performance.
    \item In section \ref{sec:conclusion} we conclude the paper.
\end{itemize}

\section{Theoretical Background}
\label{sec:Background}

We consider a supervised learning setting, in which we assume that training and test points are drawn i.i.d. according to some distribution $\mathcal D$ over the input space $\mathcal X$ and labels $\mathcal Y$. For training, we have given a labeled sample $\mathcal{S} = \{(x_i,y_i)|i=1,\dots,N\}$, where $x_i \in \mathcal X \subseteq \mathbb R^d$ is a $d$-dimensional feature-vector and $y_i\in \mathcal Y \subseteq \mathbb R^C$ is the corresponding target vector. For regression problems we have $C=1$ and $\mathcal Y = \mathbb R$. For binary classification with $C = 2$ we use $\mathcal Y = \{-1,+1\}$ and for multiclass problems with $C\ge 3$ classes we encode each label as a one-hot vector $y = (0,\dots,0,1,0,\dots,0)$ which contains a `$1$' at coordinate $c$ for label $c \in \{0,\dots,C-1\}$. In this paper we are interested in the overfitting behavior of ensembles and more specifically of Random Forests. In this context we refer to overfitting as the u-shaped curved depicted in Figure \ref{fig:Bias-Variance} in which the test error increases again after a certain complexity, but \emph{not} the fact that in many practical applications there is a gap between the test and training error. We assume a convex combination of $M$ classifier $h_i \in \mathcal H$ each scaled by their respective weights $w_i\in [0,1]$:
$$
f(x) = \sum_{i=1}^M w_i h_i(x)
$$
with $\sum_{i=1}^M w_i = 1$.
For concreteness, we assume that $\mathcal H$ is the model class of axis-aligned decision trees and $f$ is a random forest. An axis-aligned DT partitions the input space $\mathcal X$ into increasingly smaller d-dimensional hypercubes called leaves and uses independent predictions for each leaf node in the tree. Formally, we represent a tree as a directed graph with a root node and two children per node. Each node in the tree belongs to a sub-cube $\mathcal I \subseteq \mathcal X$ and all children of each node recursively partition the region of their parent node into $2$ non-overlapping smaller cubes. Inner nodes perform an axis-aligned split $1\{x_k \le t\}$ where $k$ is a feature index and $t$ is a split-threshold. Each node is associated with split function $s(x) \colon \mathcal X \to \{0,1\}$ that is `1' if $x$ belongs to the hyper-cube of that node and '0' if not. 
To compute $k$ and $t$ the gini score (CART algorithm \cite{breiman/etal/1984}) or the information gain (ID3 algorithm \cite{quinlan/1986}) is minimized which measure the impurity of a split. The induction starts with the root node and the entire dataset. Then the optimal splitting is computed and the training data is split into the left part ($1\{x_k \le t\}$) and the right part ($1\{x_k > t\}$). This splitting is repeated recursively until either a node is `pure' (it contains only examples form one class) or another abort criterion, e.g. a maximum number of leaf nodes, is reached. The predictions on the leaf nodes are computed by estimating the class probabilities of all observations in that specific leaf. To classify a new example one starts at the root node and traverses the tree according to the comparison $1\{x_k \le t\}$ in each inner node. Let $L$ be the total number of leaf nodes in the tree and let $L_i = (n_1,n_2,\dots)$ be the nodes visited depending on the outcome of $1\{x_k \le t\}$ then the prediction function of a tree is given by
$$
h(x) = \sum_{i = 1}^L \widehat y_i s_i(x)  = \sum_{i=1}^L \widehat y_i \prod_{l \in L_i} s_{i,l}(x)
$$
where $\widehat y_i \in \mathbb R^C$ is the (constant) prediction value per leaf and $s_{i,l}$ are the split functions of the individual hypercubes.

A Random Forest extends a single DT by training a set of $M$ axis-aligned decision trees on bootstrap samples using $d_i \ll d$ randomly sampled features and weighting them equally. Algorithm \ref{fig:RF} summarizes the RF algorithm.

\begin{algorithm}
	\begin{algorithmic}[1]
	\For{$i=1,\dots,M$}
	    \State $X_i \gets \text{bootstrap\_sample}(X)$
	    \State $h_i \gets \text{new\_root\_node}()$
	    \State $nodes \gets [(h_i, X_i)]$
	    \While{$\text{len(nodes)}>0$}
    	   \State $n, X_n \gets \text{nodes.pop()}$
    	   \State $d_i \gets \text{sample\_features}(d)$
    	   \State $split \gets \text{compute\_best\_split}(X_n, d_i)$
    	   \State $n.\text{set\_split}(X_n, split)$
    	   \State $X_l, X_r \gets \text{split\_data}(X_n, split)$
    	   \State $c_l, c_r \gets \text{new\_children}(n)$
    	   \If{tree\_not\_done}
    	   \State $nodes.\text{append}(c_l, X_l)$
    	   \State $nodes.\text{append}(c_r, X_r)$
    	   \EndIf
	    \EndWhile
	    \State $trees.\text{append}(h_i)$
	    \State $weights.\text{append}(1/M)$
	\EndFor
	\end{algorithmic}
	\caption{Random Forest algorithm.}
	\label{fig:RF}
\end{algorithm}


\section{There is no double-descent in RF}
\label{sec:complexity}

In statistical learning theory the generalization error of a model $f$ is bounded in terms of its empirical error $\frac{1}{N}\sum_{i=1}^N \ell(f(x),y)$ given some loss function $\ell$ and a complexity measure $\mathcal R$ for the trained model. For concreteness consider a binary classification problem with $\mathcal Y = \{-1,+1\}$ and let $f \colon \mathcal X \to [-1,+1]$ be a prediction model and $\rho > 0$ be the classification margin. We denote the binary classification error of $f$ on $\mathcal D$ with respect to $\rho$ with $L_{\rho}(f)$ and the empirical classification error of $f$ wrt. $\rho$ on $\mathcal S$ with $\widehat L_{\rho,\mathcal S}(f)$:
\begin{align}
    L_{\rho}(f) &= \underset{{(x,y)\sim \mathcal D}}{\mathbb{E}}\left[{1}\{y f(x)\le \rho\}\right] \\
    \widehat{L}_{\rho,\mathcal S}(f) &= \underset{{(x,y)\sim \mathcal S}}{\mathbb{E}}\left[{1}\{y f(x)\le \rho\}\right] = \frac{1}{N}\sum_{i=1}^N {1}\{y_i f(x_i)\le \rho\}
\end{align}

Intuitively, a large margin indicates how convinced we are in our predictions. For example consider the two predictions $f(x_1) = 0.9$ and $f(x_2) = 0.1$, which both can be considered to be class `+1`. Now using $\rho = 0.8$ means that we only accept predictions greater $0.8$ as `+1`, so that the prediction for $x_2$ will be regarded as wrong in any case (regardless of the actual label $y_2$). Using $\rho = 0$ indicates that we do not care if the prediction is $0.9$ or $0.1$ -- both are considered equally to belong to class `+1'. The following theorem bounds the generalization error of a convex combination of classifiers in terms of their individual Rademacher complexities. 

\begin{theorem}[Convex combination of classifiers \cite{cortes/etal/2014}]
	\label{th:ConvexRademacher}
	Let $\mathcal H = \bigcup_{j=1}^k \mathcal H_{j}$ denote a set of base classifiers and let $f = \sum_{i=1}^M w_i h_i$ with $w_i\in [0,1], \sum_{i=1}^M w_i = 1$ be the convex combination of classifiers $h_i \in \mathcal H$. Furthermore, let $\mathcal R(h_i)$ be the Rademacher complexity of the i-th classifier. Then, for a fixed margin $\rho > 0$ and for any $\delta > 0$, with probability at least $1-\delta$ over the choice of sample $\mathcal{S}$ of size $N$ drawn i.i.d. according to $\mathcal D$, the following inequality holds:
	\begin{align*}
	L_0(f) &\leq \widehat L_{\mathcal{S},\rho}(f) + \frac{4}{\rho}\sum_{i=1}^M w_i \mathcal R(h_i) + C(N,k)
    \end{align*}
    where $C(N,k,\rho)$ is a constant depending on $N, k, \rho$ which tends to $0$ for $N\to\infty$ for any $k$ and any $\rho$.
\end{theorem}

Theorem \ref{th:ConvexRademacher} offers two interesting insights: First, the Rademacher complexity of a convex combination of classifiers does \emph{not} increase and thus, an ensemble is not more likely to overfit than each of its individual base learners. Second, the individual Rademacher complexities of each base learner are scaled by their respective weights. For $k = 1$, where all classifiers have the same complexity, this bound recovers the well-known result from \cite{Koltchinskii/Panchenko/2002}. 

The key question now becomes how to compute the Rademacher complexity of the trees inside the forest. It is well-known that the Rademacher complexity $\mathcal R(h_i)$ is related to the VC-dimension $D(h_i)$ via 
\begin{equation}
\mathcal R(h_i) \le \sqrt{\frac{2 D(h_i)}{N}}.
\end{equation}
Interestingly, the exact VC-Dimension of decision trees is unknown. Asian etal. performed an exhaustive search to compute the VC dimension of trees with depth up to $4$ in \shortcite{asian/etal/2009}, but so far no general formula has been discovered. However, there exist some useful bounds. A decision tree $h_i$ with $n_i$ nodes trained on $d_i$ binary features has a VC-dimension of at most \cite{mansour/1997}:
\begin{equation}
D(h_i) \le (2 n_i + 1) \log_2(d_i + 1)
\end{equation}
Leboeuf et al. extend this bound for continuous features in \shortcite{leboeuf/etal/2020} by introducing the concept of partition functions into the VC-framework. They are able to show that the VC-dimension of a decision tree trained on $d_i$ continuous features is of order $\mathcal O \left(n_i \log(n_i + d_i) \right)$. Unfortunately, the expression discovered by the authors is computationally expensive so that experiments with larger trees are impractical\footnote{The authors provide a simplified version of their expression which works well for trees with less than $100$ leaf nodes on our test system, but anything beyond that would take too long.}. For our analysis in this paper we are interested in the asymptotic behavior of Decision Trees and Random Forests. Hence we use the following \emph{asymptotic} Rademacher complexity:
\begin{equation}
\widehat {\mathcal R}(h_i) = \sqrt{\frac{2n_i \log(n_i + d_i)}{N}}
\end{equation}


The previous discussion highlights two things: First, the complexity of an RF does not increase when adding more trees but it \emph{averages}. Second, the complexity of a tree largely depends on the number of features $d_i$ and the total number of nodes $n_i$ (ignoring any factors). Belkin et al. empirically showed in \shortcite{belkin/etal/2019} that Random Forest exhibit a double-descent curve. Similar to our discussion here, the authors introduce the number of nodes as a measure of complexity for single trees, but then use the \emph{total} number of nodes in the forest throughout their discussion. While we acknowledge that this is a very intuitive definition of complexity it is not consistent with our above discussion and the results in learning theory. Hence, we propose to use the \emph{average} (asymptotic) Rademacher complexity as a capacity measure. We argue, that with this adapted definition, there is no double-descent occurring in Random Forests but rather a single descent in which we fit the training data better and better the more capacity is given to the model. 


\subsection{A RF shows a single descent curve}

We validate our hypothesis experimentally. To do so we train various RF models with different complexities and compare their overfitting behavior on five different datasets depicted in table \ref{tab:datasets}. By today's standards these datasets are small to medium size which allows us to quickly train and evaluate different configurations, but large enough to train large trees. The code for our experiments is available under \url{https://github.com/sbuschjaeger/rf-double-descent}.

\begin{table}[]
\centering
\begin{tabular}{@{}lrlr@{}}
\toprule
Dataset & N  & C & d \\ \midrule
Adult   & 32~562 & 2 &  108   \\
Bank    & 45~211 & 2 &  51   \\
EEG     & 14~980 & 2 &  14   \\
Magic   & 19~019 & 2 &  10   \\
Nomao   & 34~465 & 2 &  174   \\ \bottomrule
\end{tabular}
\caption{\label{tab:datasets}Datasets used for our experiments. We performed minimal pre-processing on each dataset removing instances which contain \texttt{NaN} values and computed a one-hot encoding for categorical features. Each dataset is available under \texttt{https://archive.ics.uci.edu/ml/datasets}.}
\end{table}
 
Our experimental protocol is as follows:  Oshiro et al. showed in \cite{oshiro/etal/2012} empirically on a variety of datasets that the prediction of a RF stabilizes between $128$ and $256$ trees and adding more trees to the ensemble does not yield significantly better results. Hence, we train the `base' Random Forests with $M = 256$ trees. To control the complexity of a Random Forest, we limit the maximum number of leaf nodes $n_l$ of the individual trees to $\{2,4,8,16,32,64,\dots,16384\}$. In all our experiments we perform a 5-fold cross validation and report the average error across these runs. 

\begin{figure*}
\begin{subfigure}[c]{\dimexpr0.30\textwidth+20pt\relax}
    \makebox[20pt]{\raisebox{60pt}{\rotatebox[origin=c]{90}{Adult}}}%
    \includegraphics[width=5.5cm, height=4cm]{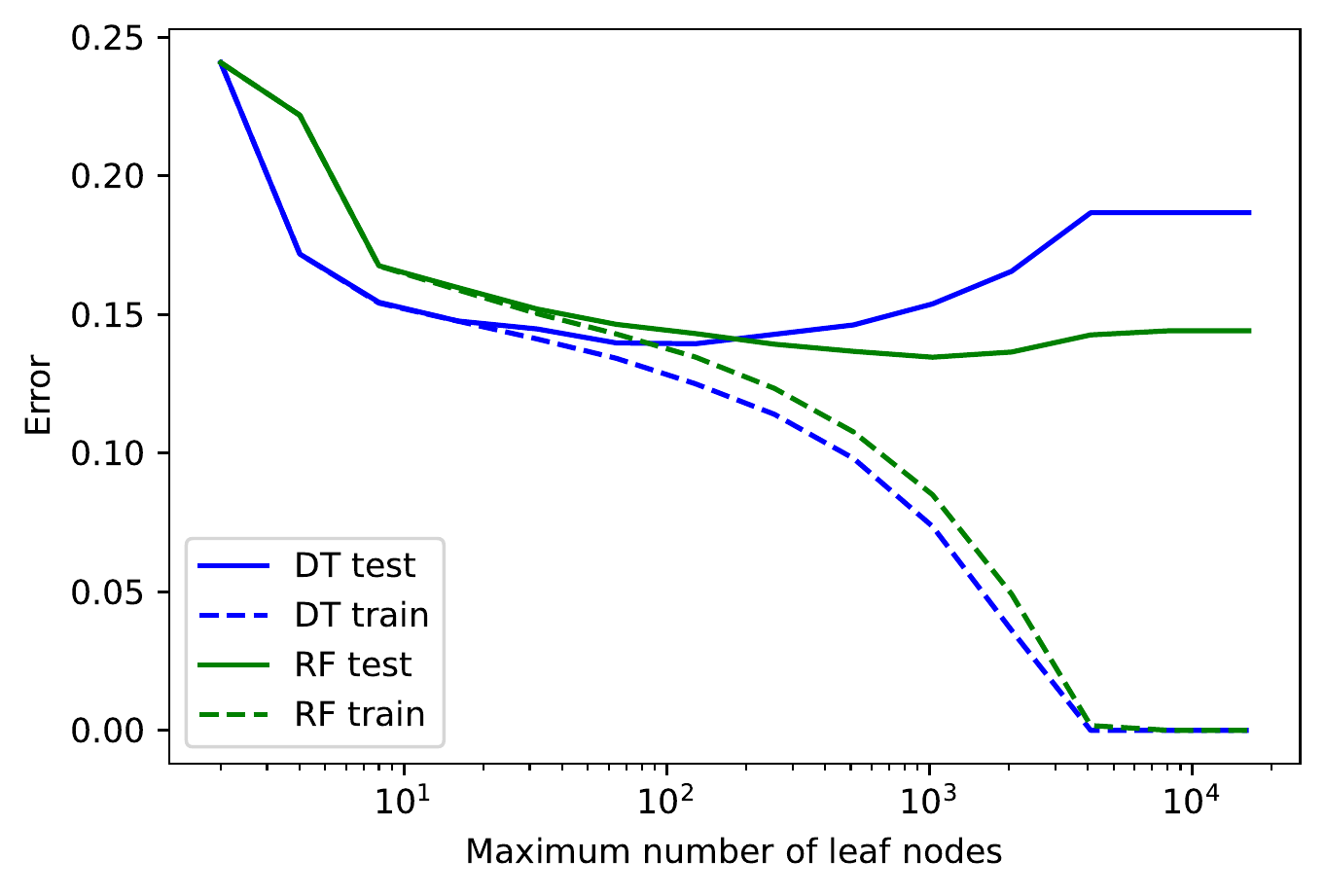}
    \makebox[20pt]{\raisebox{60pt}{\rotatebox[origin=c]{90}{Bank}}}%
    \includegraphics[width=5.5cm, height=4cm]{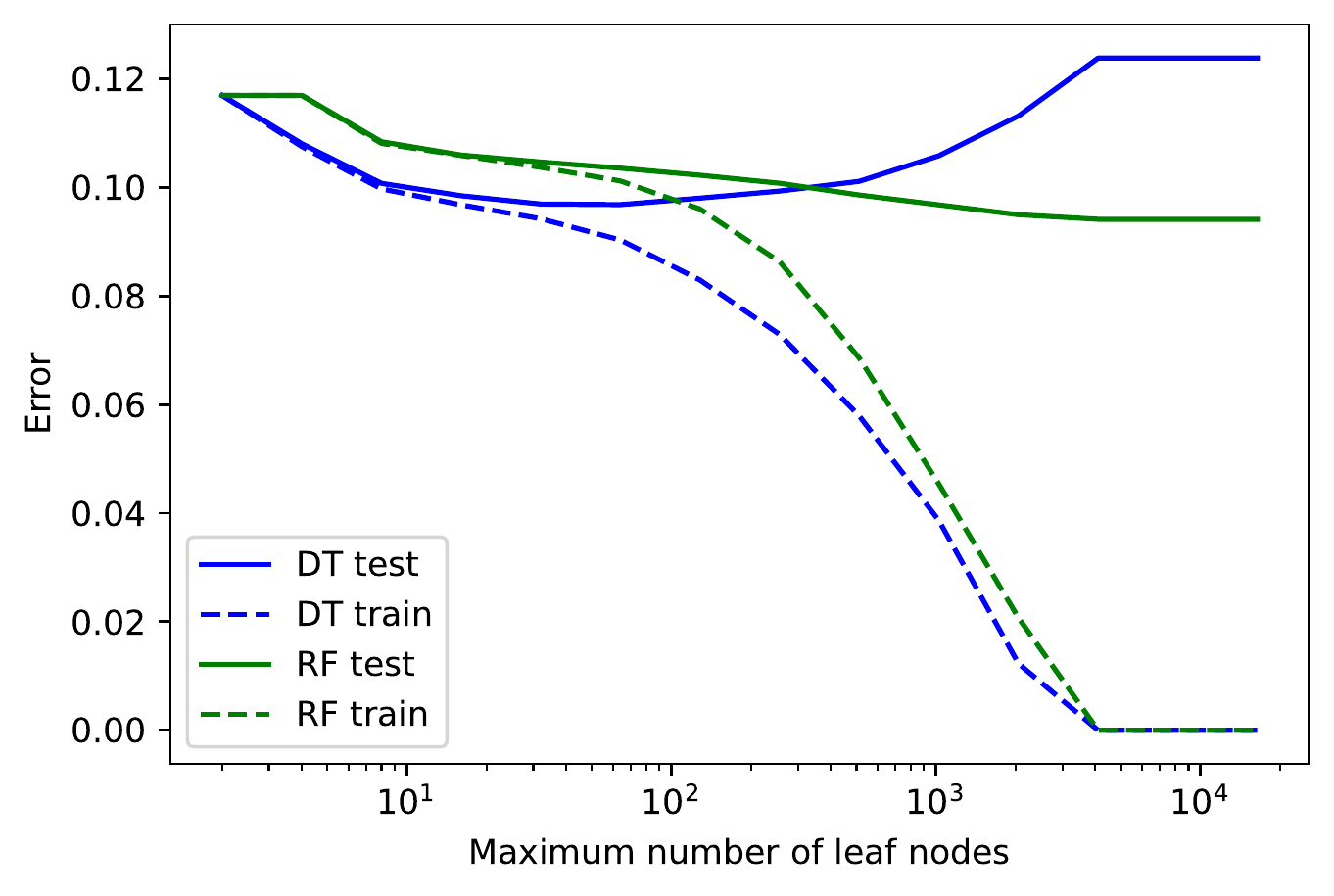}
    \makebox[20pt]{\raisebox{60pt}{\rotatebox[origin=c]{90}{EEG}}}%
    \includegraphics[width=5.5cm, height=4cm]{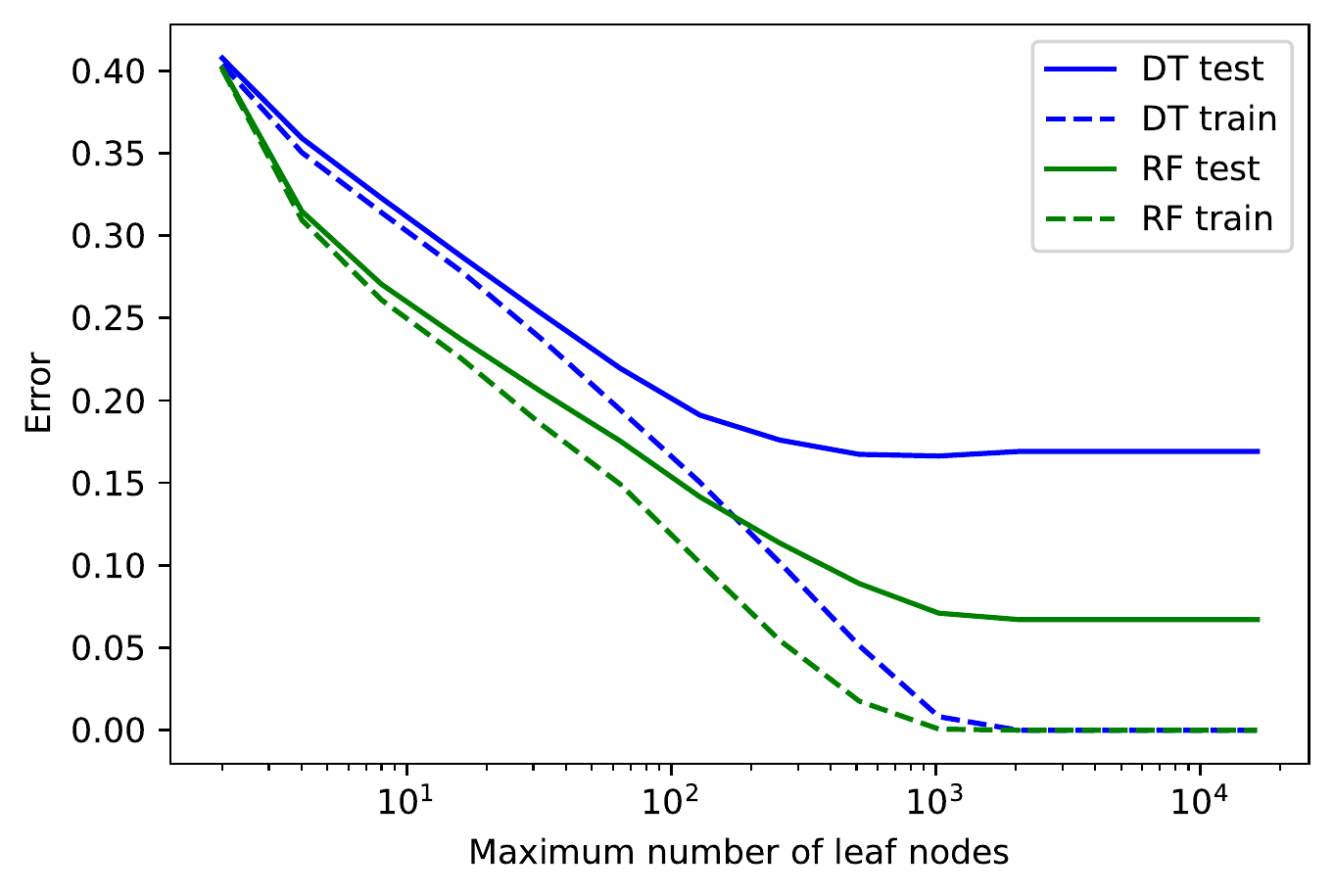}
    \makebox[20pt]{\raisebox{60pt}{\rotatebox[origin=c]{90}{Magic}}}%
    \includegraphics[width=5.5cm, height=4cm]{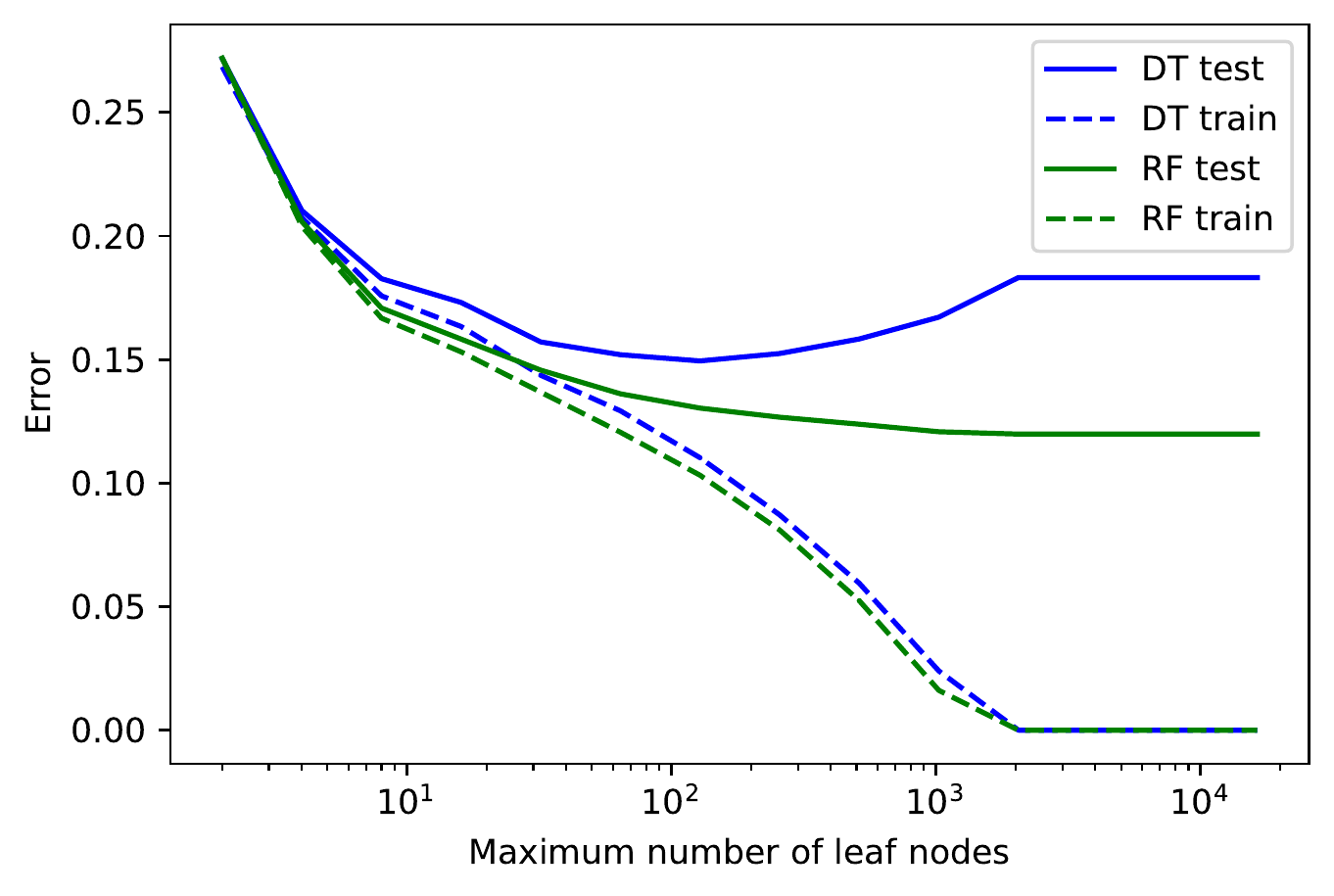}
    \makebox[20pt]{\raisebox{60pt}{\rotatebox[origin=c]{90}{Nomao}}}%
    \includegraphics[width=5.5cm, height=4cm]{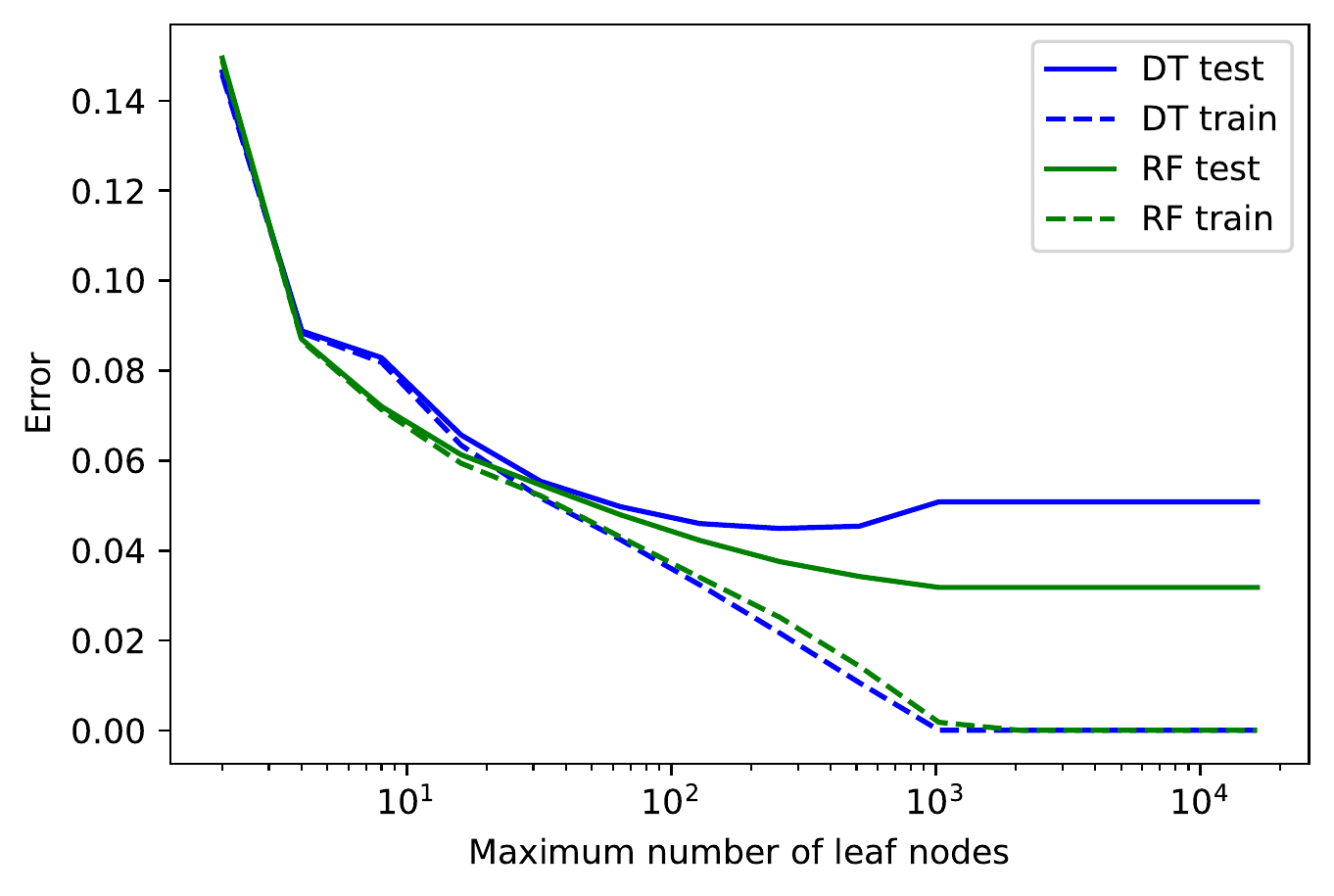}
    \caption{Test and Training error.}\label{fig:1a}
\end{subfigure}
\begin{subfigure}[c]{0.30\textwidth}
    \includegraphics[width=5.5cm, height=4cm]{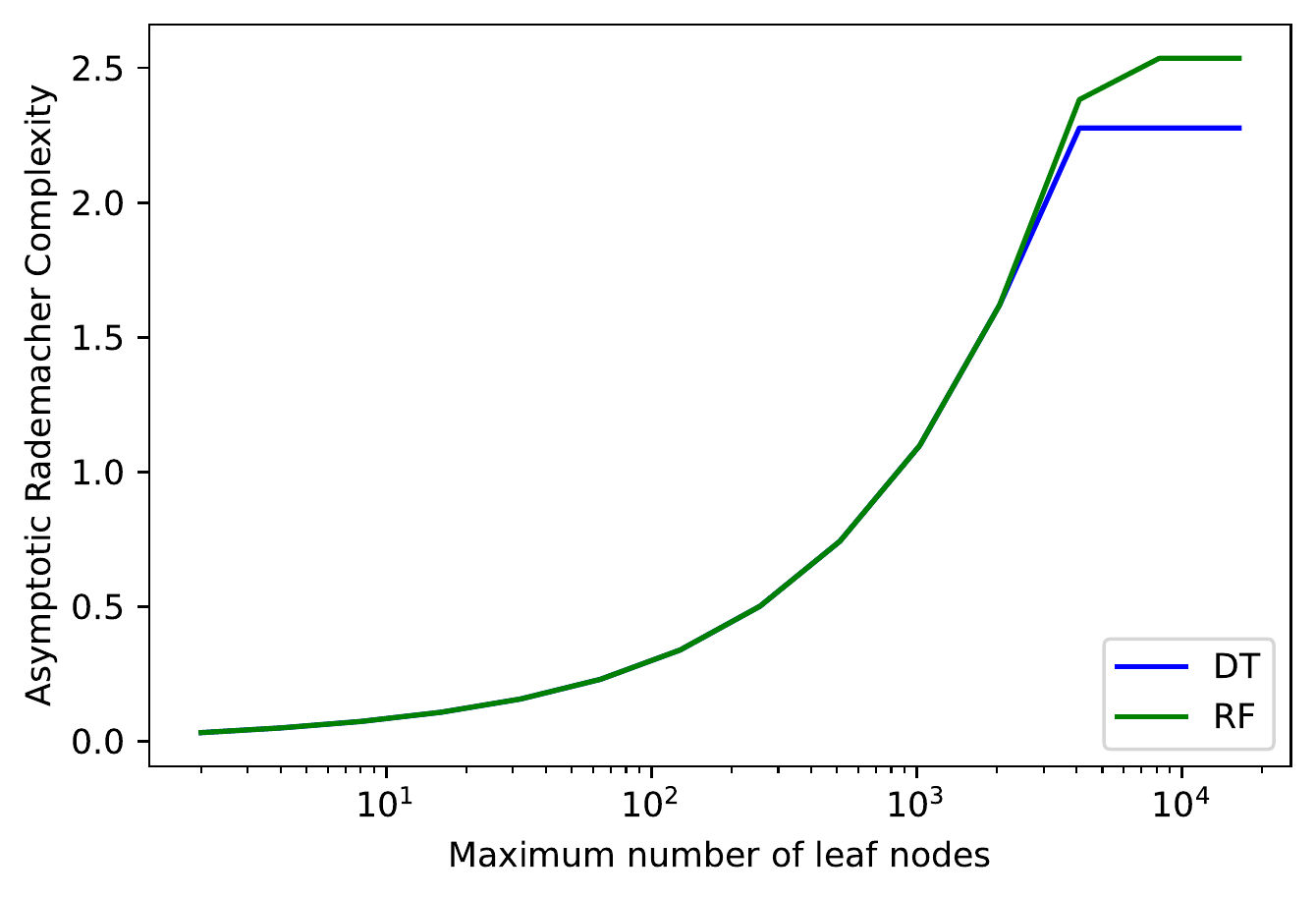}
    \includegraphics[width=5.5cm, height=4cm]{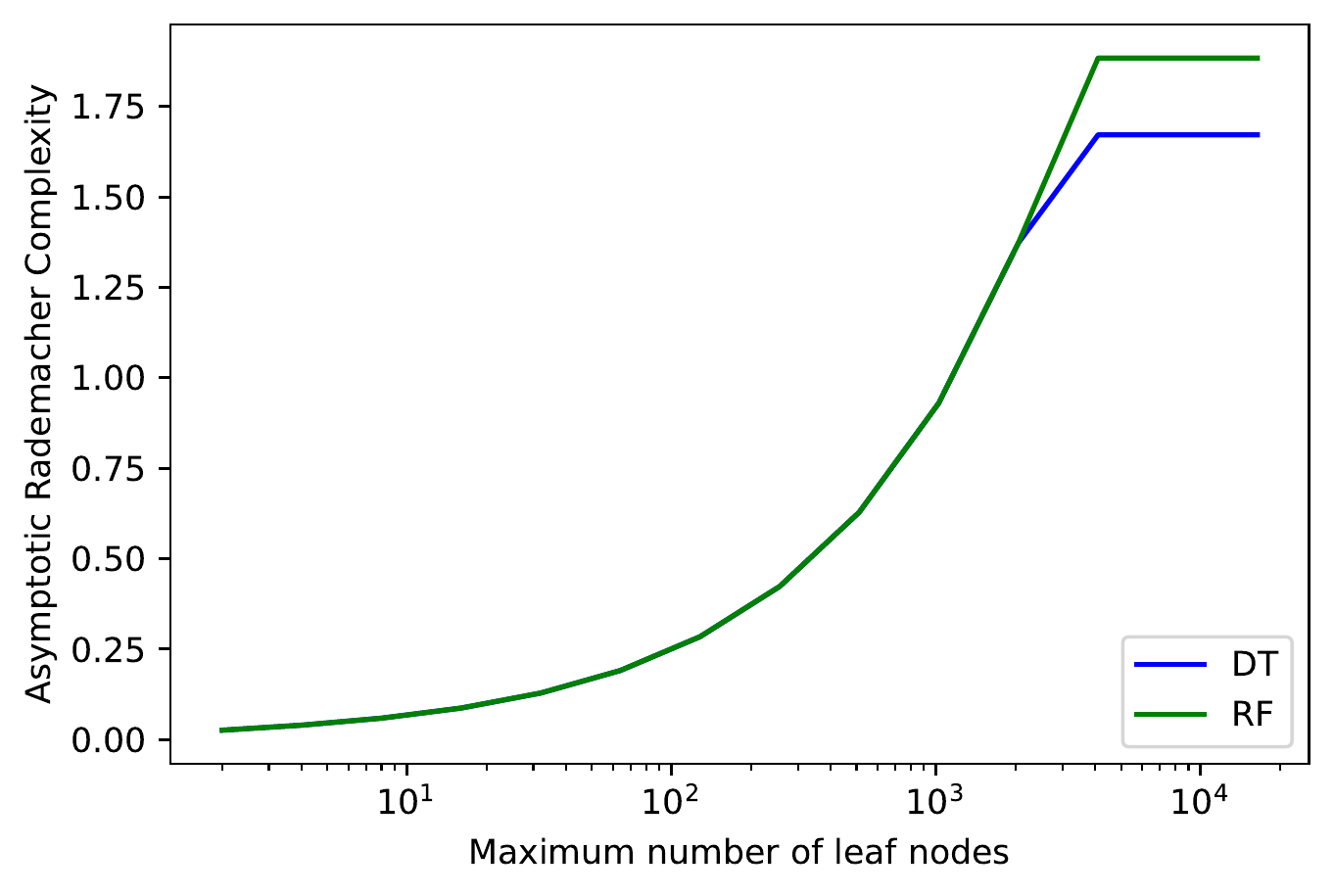}
    \includegraphics[width=5.5cm, height=4cm]{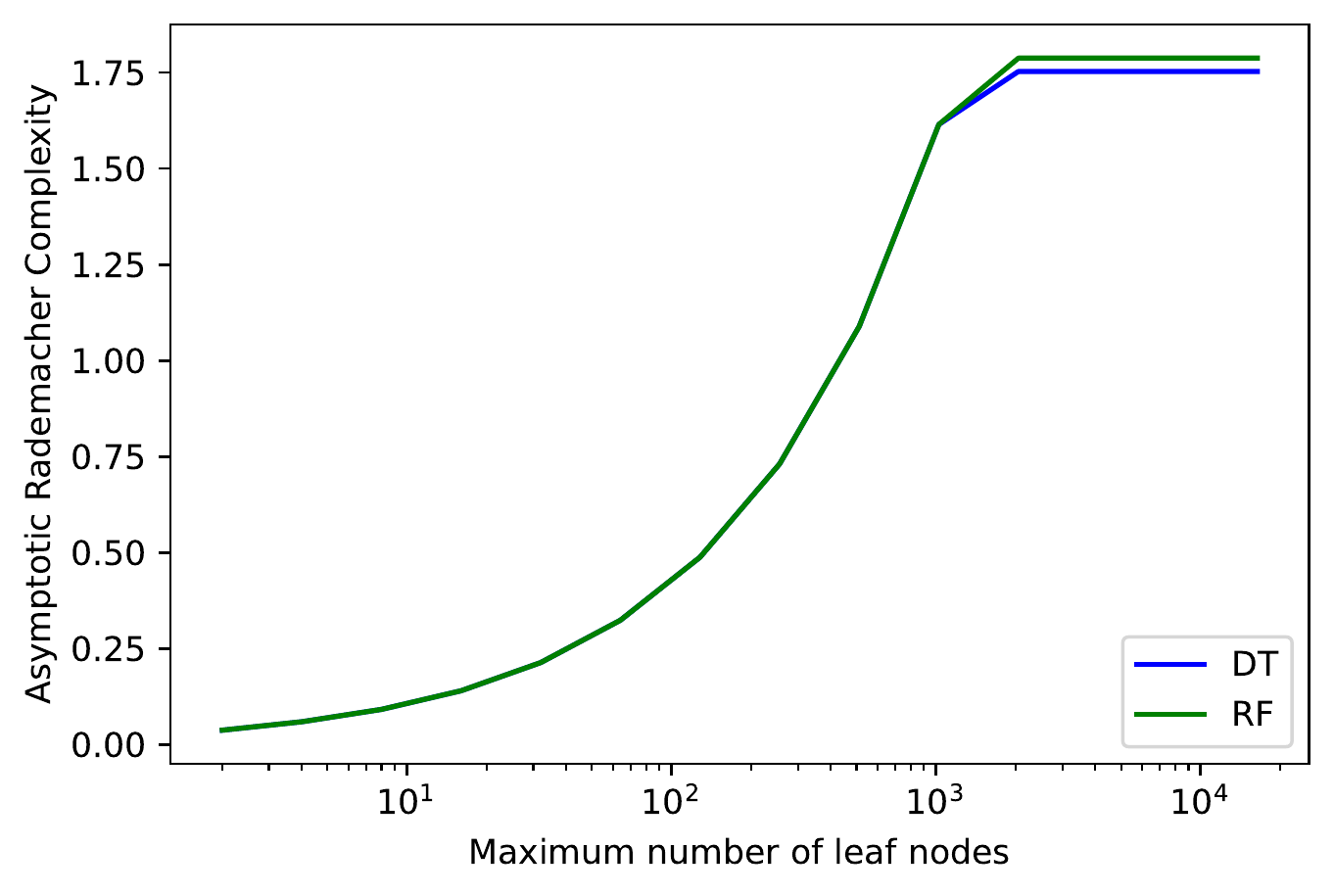}
    \includegraphics[width=5.5cm, height=4cm]{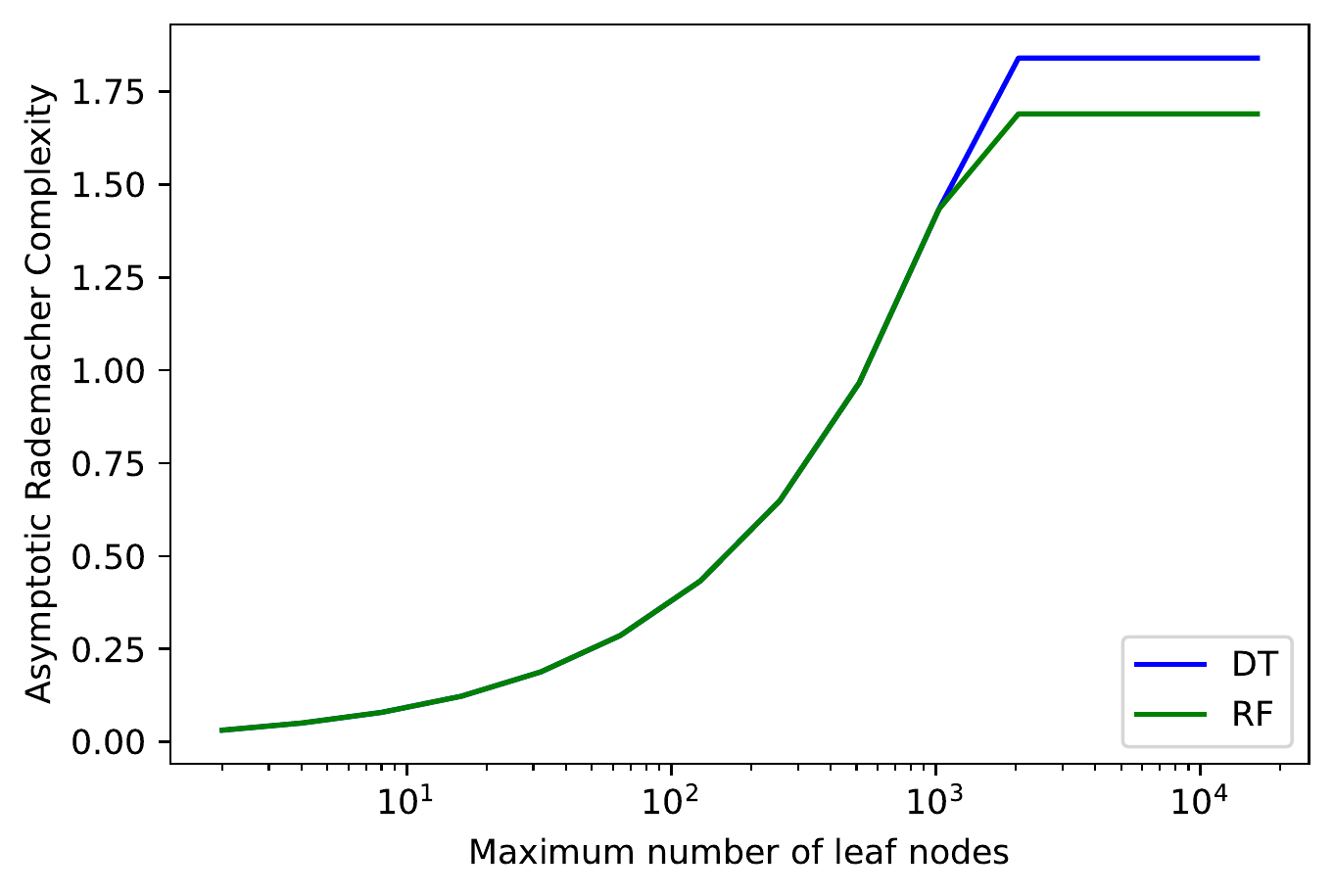}
    \includegraphics[width=5.5cm, height=4cm]{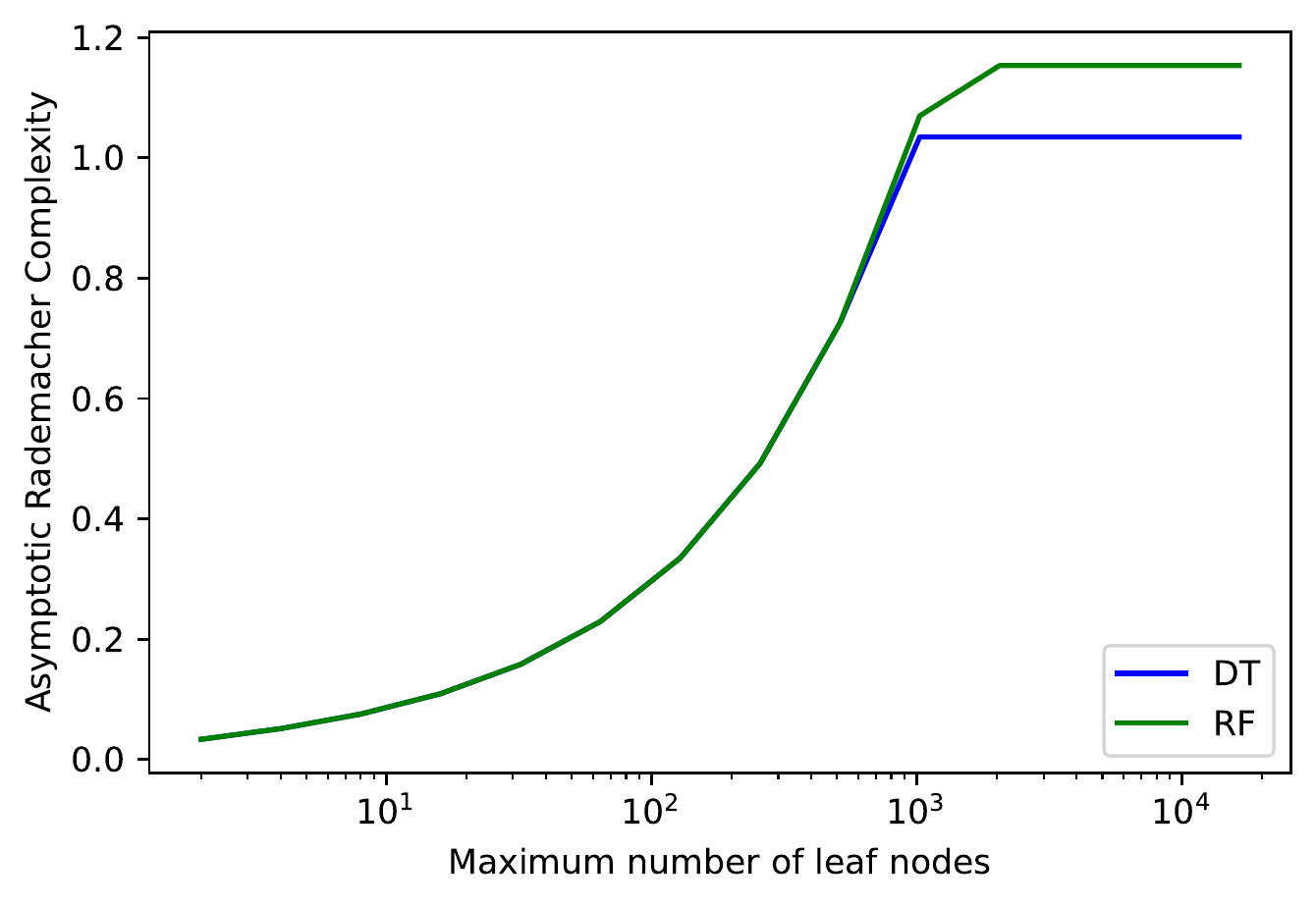}
    \caption{Average Rademacher complexity.}\label{fig:1b}
\end{subfigure}
\begin{subfigure}[c]{0.30\textwidth}
    \includegraphics[width=5.5cm, height=4cm]{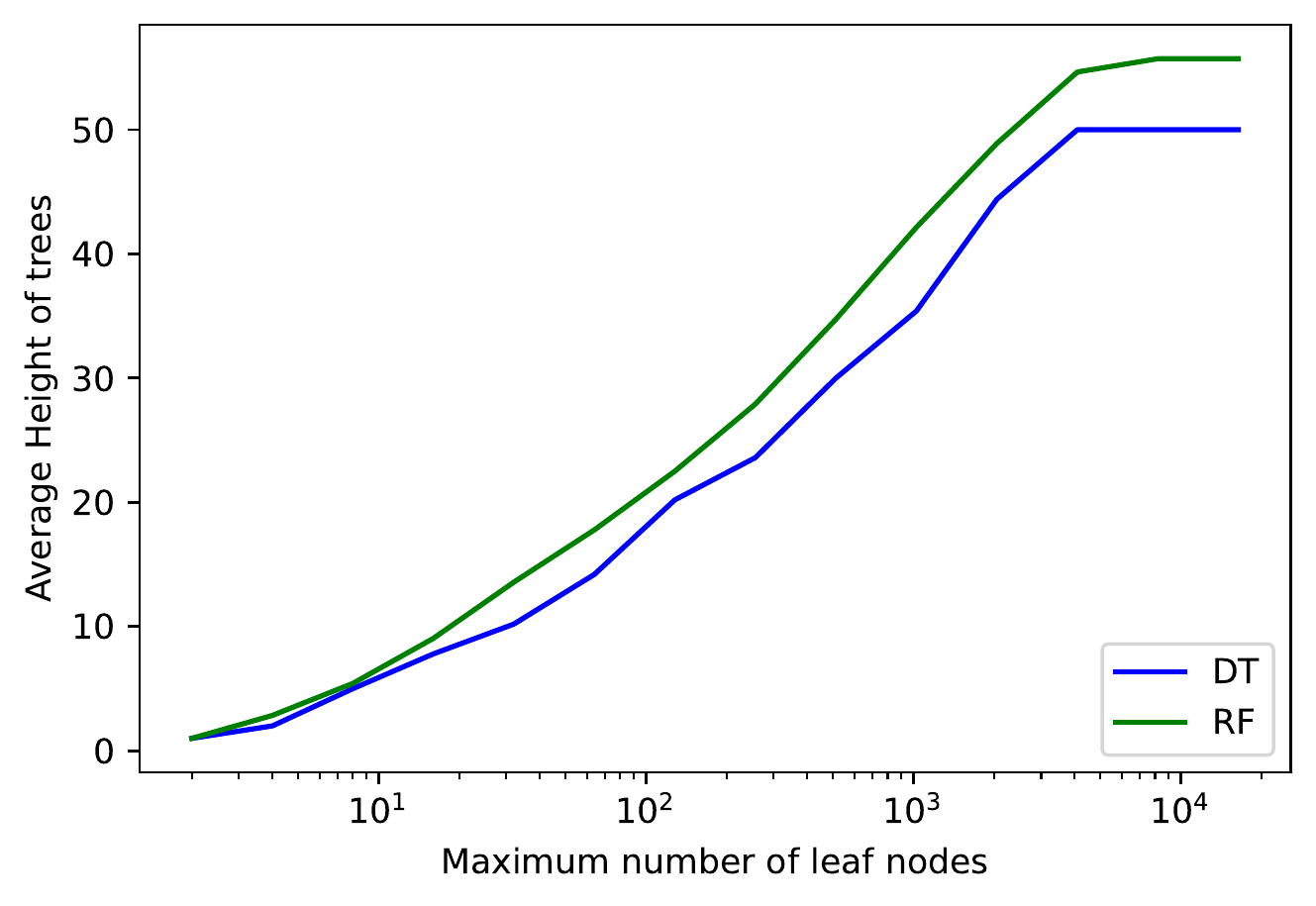}
    \includegraphics[width=5.5cm, height=4cm]{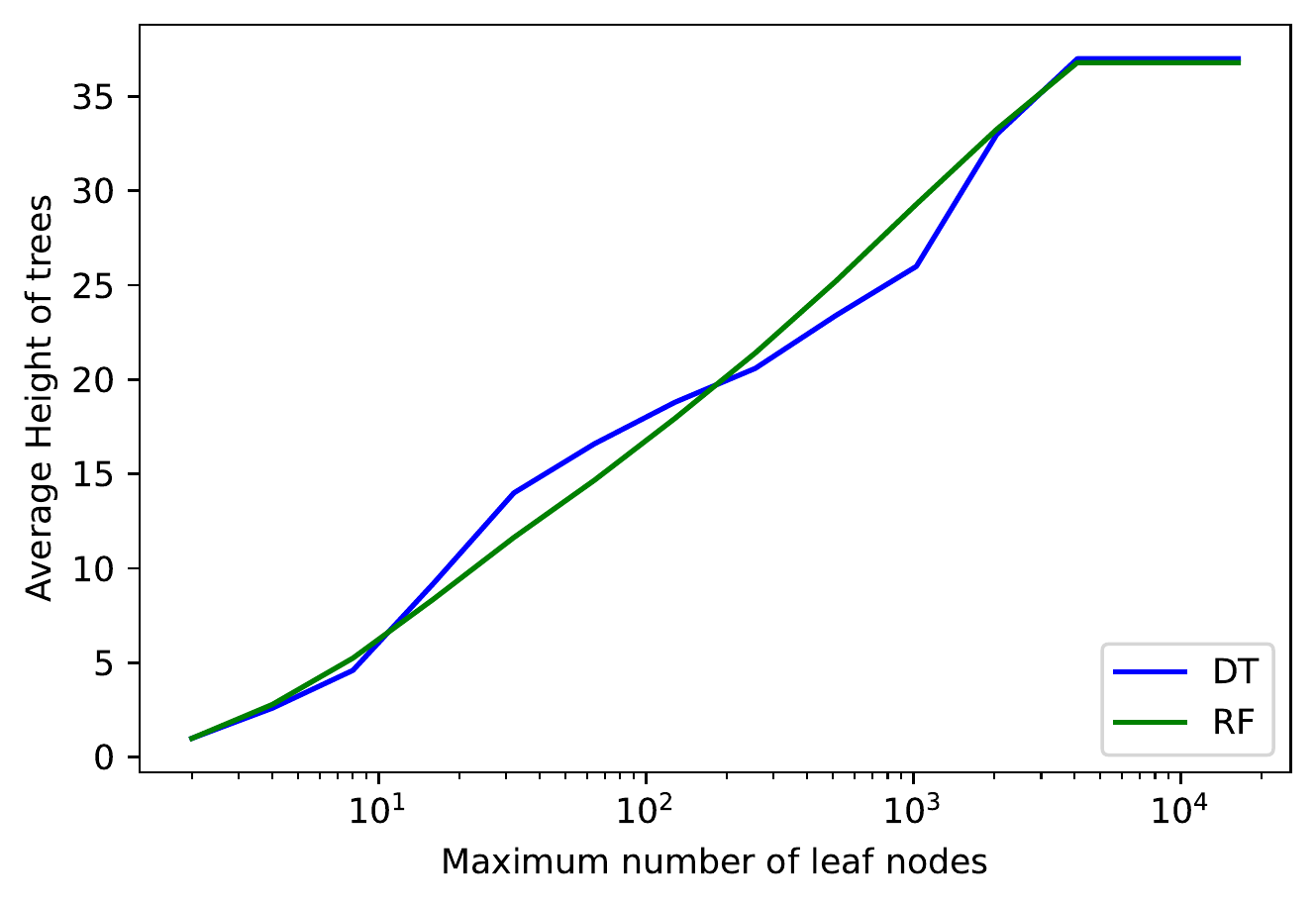}
    \includegraphics[width=5.5cm, height=4cm]{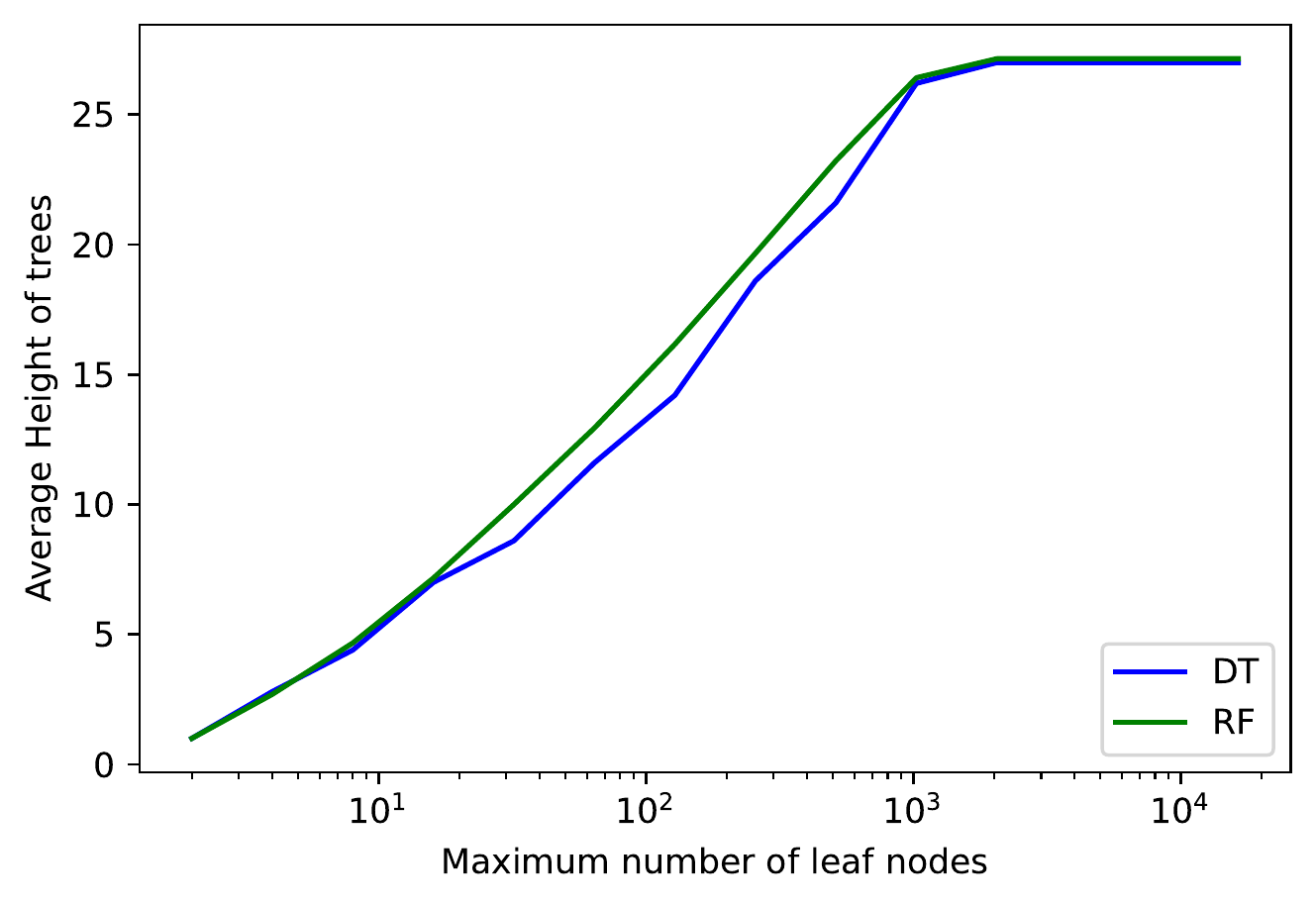}
    \includegraphics[width=5.5cm, height=4cm]{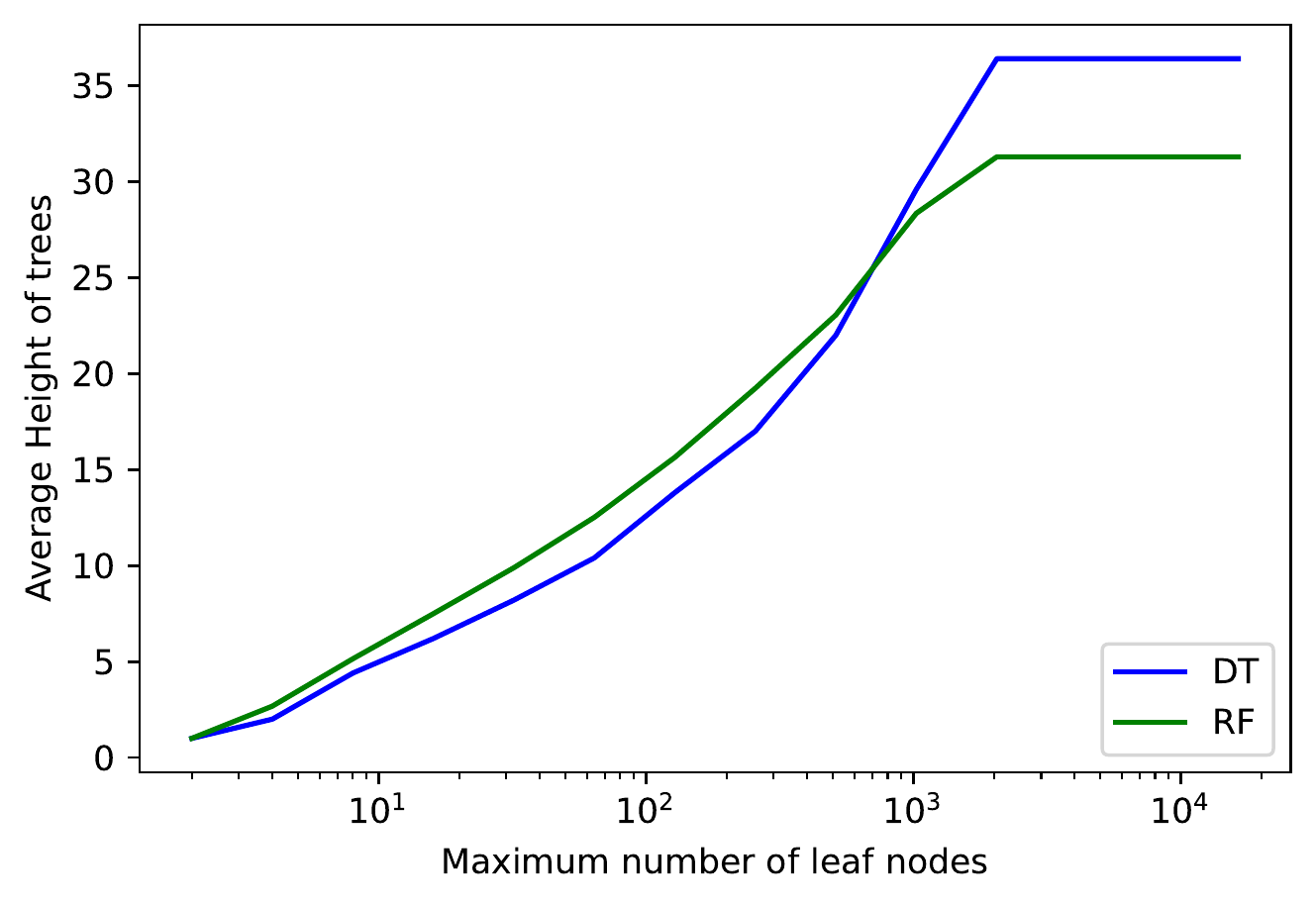}
    \includegraphics[width=5.5cm, height=4cm]{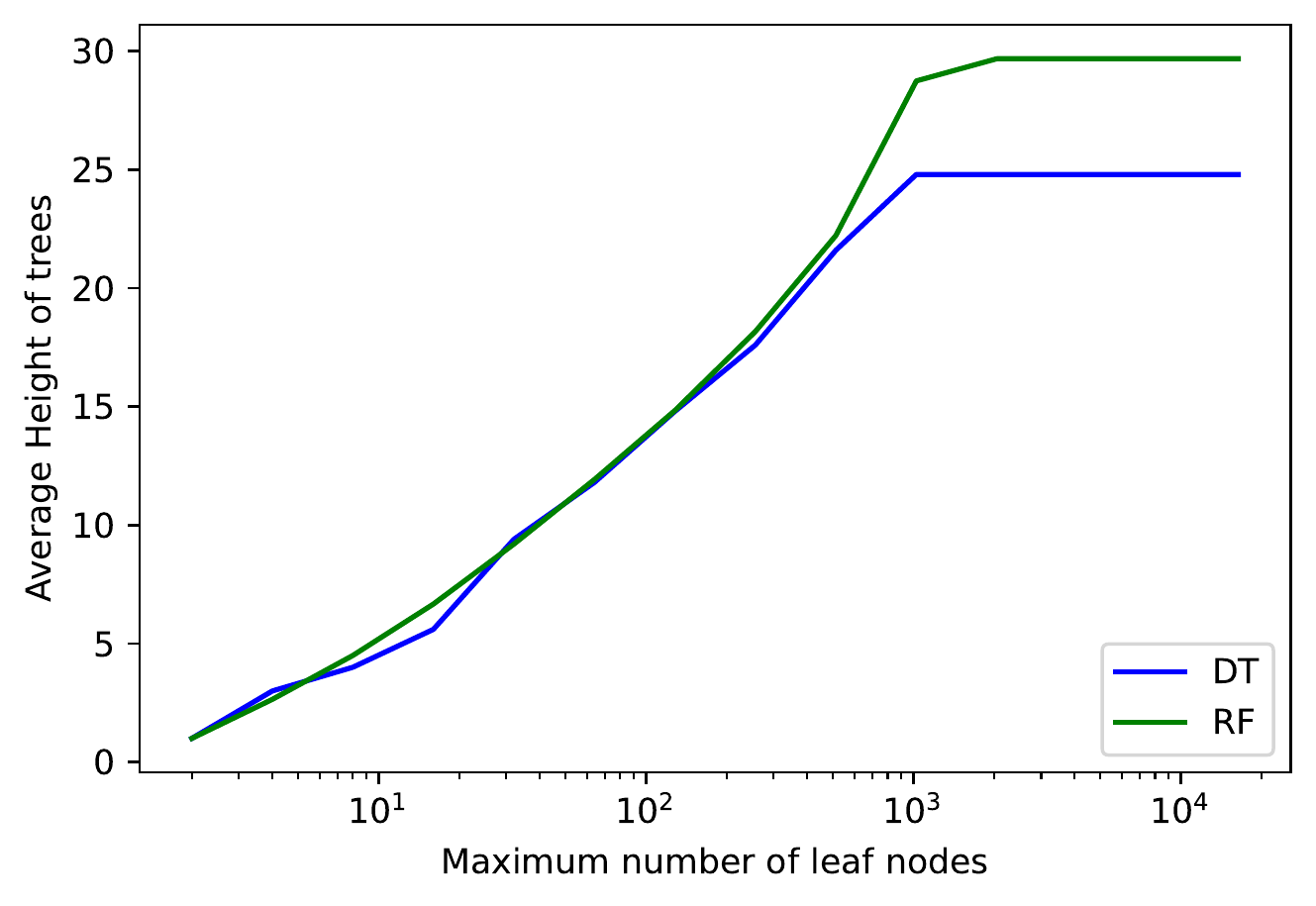}
    \caption{Average tree height.}\label{fig:1c}
\end{subfigure}
\caption{Test and training error of RF and DT (first column), the average Rademacher complexity (second column) and the average height of the trees (third column) over the maximum number of leaf nodes $n_l$. Each row depicts one dataset. Results are averaged over a 5-fold cross validation. Solid lines are the test error and dashed lines are the training error. Best viewed in color.} \label{fig:H1}
\end{figure*}

Figure \ref{fig:1a} shows the results of this experiment. Solid lines show the test error and dashed lines are the training error. Note the logarithmic scale on the x-axis. It can be clearly seen that for both, RF and DT, the training error decreases towards 0 for larger $n_l$ values. On the adult, bank, magic and nomao dataset we see the `classic' u-shaped overfitting curve for a DT in which the error first improves and then suddenly increases again. On the EEG dataset the DT shows a single-descent in which it never overfits, but its test error is much higher than that of a RF. On all datasets, the DT seems to reach a plateau after a certain number of maximum leaf nodes. Looking at the RF we see a single-descent curve on all but the adult dataset in which the RF fits the data better and better with larger $n_l$. Only on the adult dataset there are signs of small overfitting for the RF.

When there is no double-descent in Random Forests, then why are they performing better than single trees? Interestingly, the above discussion may already offers a reasonable explanation of this behavior. First, a Random Forest uses both feature sampling as well as bootstrapping for training new trees. When done with care\footnote{In scikit learn \cite{Pedregosa/etal/2001} the implementation may evaluate more than $d_i$ features if no sufficient split has been found.}, then feature sampling can reduce the number of features to $d_i \ll d$ so that $\log_2(d_i+1)$ also becomes smaller and thereby also reduces $\mathcal R$. Second, bootstrap sampling samples data points with replacement. Given a dataset with $N$ observations, there are only 
$$
1 - \lim_{N\to\infty} \left(1 - \frac{1}{N}\right)^N = 1 - e^{-1} \approx 0.632
$$
unique data points per individual bootstrap sample in the limit. Thus, the effective size of each bootstrap sample reduces to roughly $N_i = 0.632 \cdot N$ which can lead to smaller trees because the entire training set is smaller and easier to learn due to duplicate observations. Last and maybe most important, tree induction algorithms such as CART or ID3 are adaptive in a sense, that the tree-structure is data-dependent. In the worst-case, a complete tree is build in which single observations are isolated in the leaf nodes so that every leaf-node contains exactly one example. However, it is impossible to grow a tree beyond isolating single observations because there simply is not any data left to split. Subsequently, the Rademacher complexity cannot grown beyond this point and is limited by an inherent, data-dependent limit. We summarize these arguments into the following hypothesis.

\subsection{The maximum Rademacher complexity of RF and DT is bounded by the data}


Figure \ref{fig:1b} shows the Rademacher complexity of the DT and RF for the previous experiment. As one can see the Rademacher complexity for both models on all datasets steeply increases until they both converge against a maximum from which they then continue to plateau. So indeed, both models have an inherent maximum Rademacher complexity given by the data as expected. Contrary to the above discussion, however, the RF has a \emph{larger} Rademacher complexity than a DT on all but the magic dataset. For a better understanding we look at the average height of trees produced by the two algorithms in Figure \ref{fig:1c}. Here we can see that RF -- on average -- has larger trees than the DT given the same number of maximum leaf nodes. We hypothesize that due to the feature and bootstrap sampling sub-optimal features are chosen during the splits. Hence, a RF requires more splits in total to achieve a small loss leading to larger trees with larger Rademacher complexities.

Combining both experiments leads to a mixed explanation why RF seems to be so resilient to overfitting: For trees trained via greedy algorithms such as CART one cannot (freely) over-parameterize the final model because its complexity is inherently bounded by the provided data. Even if one allows for more leaf nodes, the algorithm simply cannot make use of more parameters. A similar argument holds for a Random Forest: Adding more trees does not increase the Rademacher complexity as implied by Theorem \ref{th:ConvexRademacher}. Thus one can add more and more trees without the risk of overfitting. Similar, increasing $n_l$ only increases the Rademacher complexity up to the inherent limit given by the data. Thus even if one allows for more parameters, a RF cannot make use of them. Its Rademacher complexity is inherently bounded by the data. However, as shown in Figure \ref{fig:1a} - \ref{fig:1c} there does not seem to be a direct, data-independent connection between the maximum number of leaf nodes and this inherent maximum Rademacher complexity.

\section{Complexity does not predict the performance of a RF}
\label{sec:data-augmentation}
The above discussion already shows that the Rademacher complexity of a forest does not seem to be an accurate predictor for the generalization error of the ensemble. In this section we further challenge the notion that complexity is a predictor of the performance of a tree ensemble and construct ensembles with large complexities that do not overfit and trees with small complexities that do overfit.
For example, we could conceive a very complex tree by simply introducing unnecessary comparisons e.g. by comparing against infinity $1\{x_k \le \infty\}$. This comparison is always true effectively adding a useless decision node to the tree and a forest of such trees would have a huge Rademacher complexity while having the same performance as before. Clearly these trees would neither be in the spirit of DT learning nor really useful in practice. 
Hence we will now look a small variation of this idea. We study the performance of a DT which approximates the decision boundary of a good, not overfitted RF and similar we study a RF which approximates the decision boundary of a bad, overfitted DT. It is conceivable that the DT `inherits' the positive properties of the RF and likewise that such a RF has all the negative properties of the original DT. 
Algorithm \ref{fig:DATraining} summarizes this approach. We first train a regular reference model e.g a RF or DT given our the data $\mathcal S$ in line 2. Then, we sample $N \cdot T$ points along the decision boundary of the model by using augmented copies of the training data. Specifically, we copy the training data $T$ times and add Gaussian Noise to the observations in these copies as shown in line $4$. Then we apply the reference model (RF or DT) to this augmented data in line $5$ and use its predictions as the new label for fitting the actual model in line $7$. 

\begin{algorithm}
	\begin{algorithmic}[1]
	\State $\mathcal S \gets X, y$
	\State $g \gets \texttt{train\_ref}(\mathcal S)$ \Comment{\parbox[t]{0.45\linewidth}{Train a DT or RF}}
	\For{$i=1,\dots,T$} \Comment{\parbox[t]{0.45\linewidth}{Generate training data}}
	    \State $X_i \gets X + \mathcal N(0, \varepsilon)$ \Comment{\parbox[t]{0.45\linewidth}{Augment data}}
	    \State $y_i \gets g(X_i)$ \Comment{\parbox[t]{0.45\linewidth}{Apply original model}}
	    \State $\mathcal S = \mathcal S \cup (X_i, y_i)$
	\EndFor
	\State $f \gets \texttt{train\_model}(\mathcal S)$ \Comment{\parbox[t]{0.45\linewidth}{Train a RF or DT}}
	\end{algorithmic}
	\caption{Training with Data Augmentation.}
	\label{fig:DATraining}
\end{algorithm}

\subsection{A RF that approximates a DT does not overfit}

\begin{figure*}

\begin{subfigure}[c]{\dimexpr0.30\textwidth+20pt\relax}
    \makebox[20pt]{\raisebox{60pt}{\rotatebox[origin=c]{90}{Adult}}}%
    \includegraphics[width=5.5cm, height=4cm]{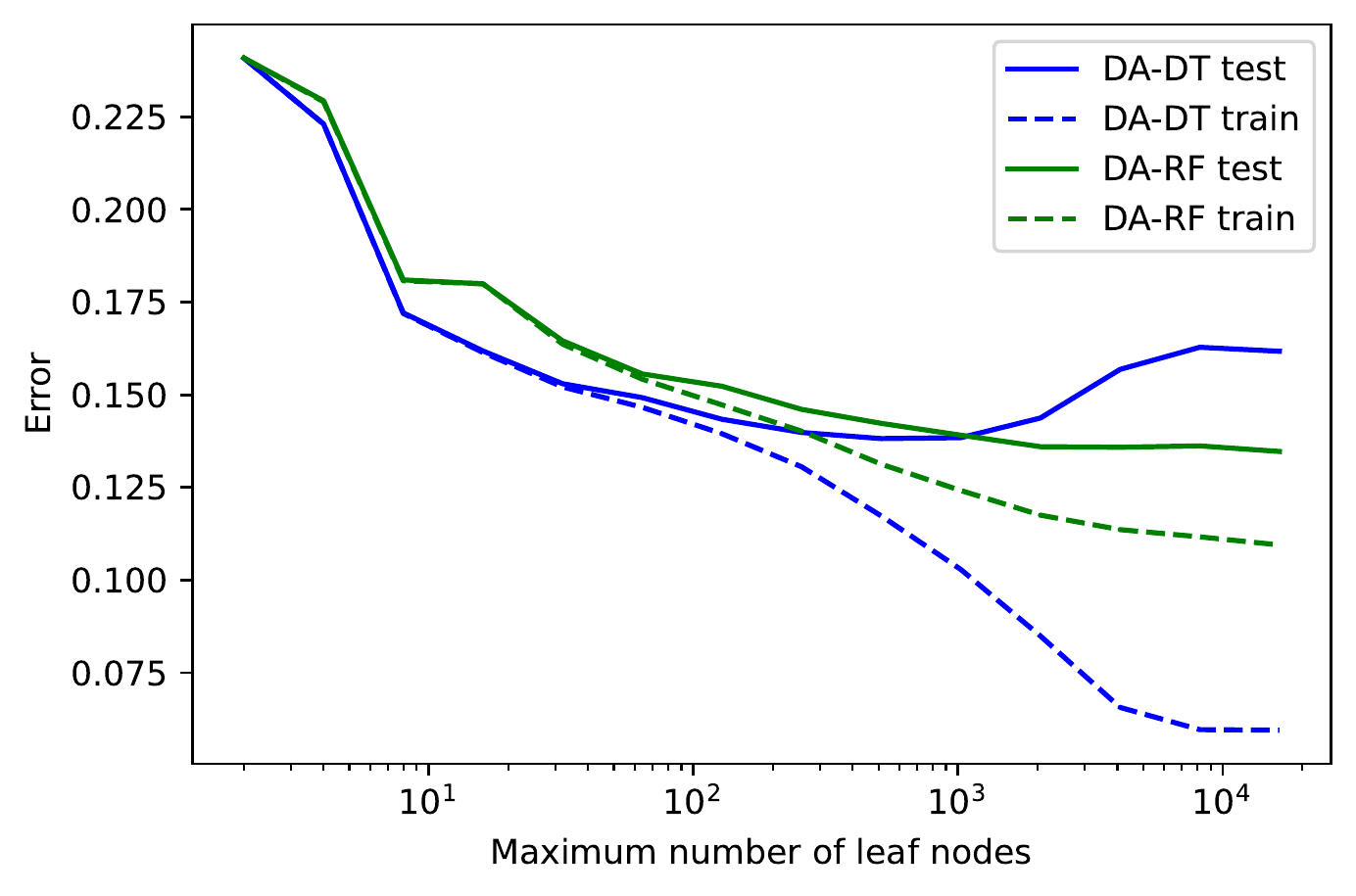}
    \makebox[20pt]{\raisebox{60pt}{\rotatebox[origin=c]{90}{Bank}}}%
    \includegraphics[width=5.5cm, height=4cm]{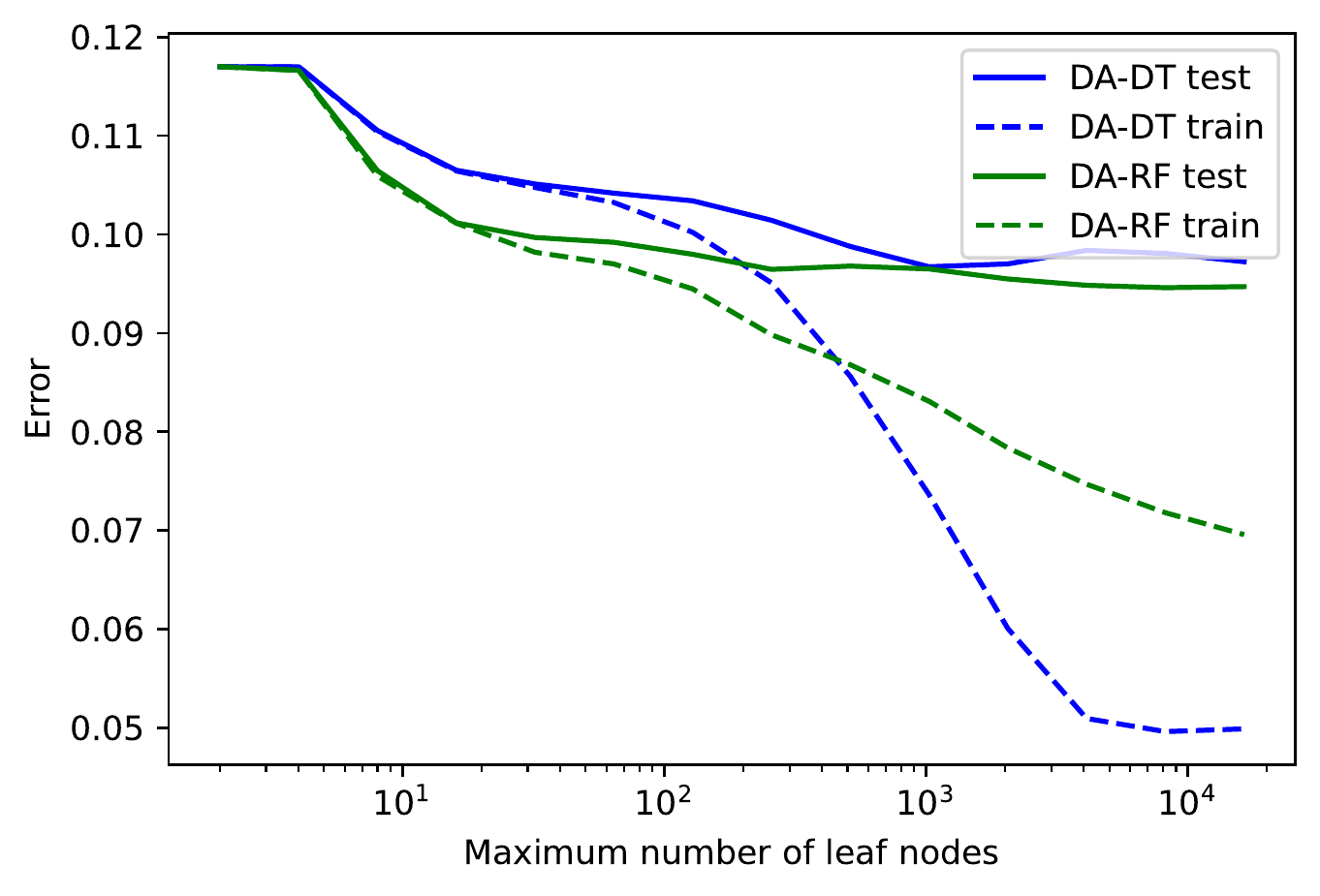}
    \makebox[20pt]{\raisebox{60pt}{\rotatebox[origin=c]{90}{EEG}}}%
    \includegraphics[width=5.5cm, height=4cm]{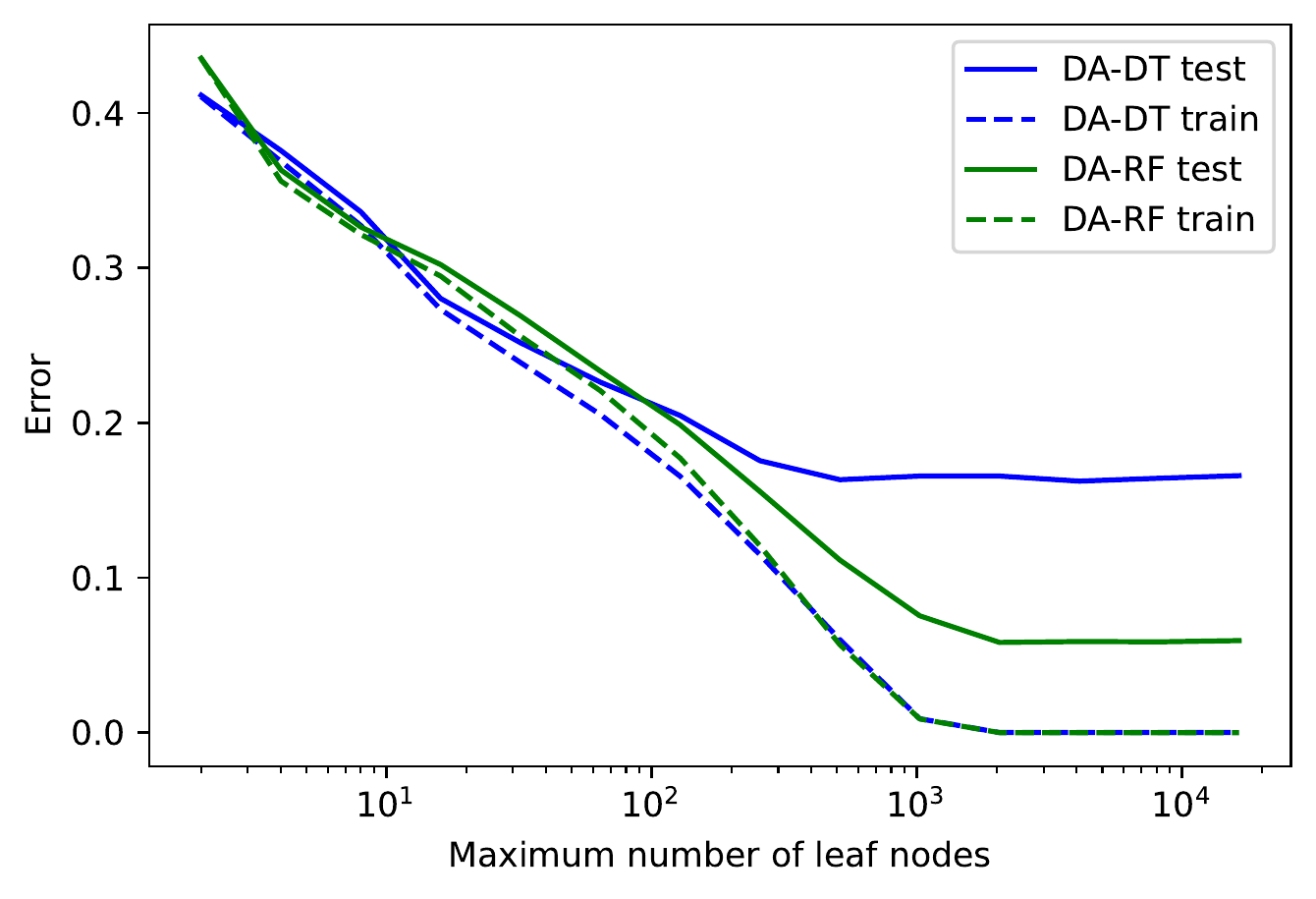}
    \makebox[20pt]{\raisebox{60pt}{\rotatebox[origin=c]{90}{Magic}}}%
    \includegraphics[width=5.5cm, height=4cm]{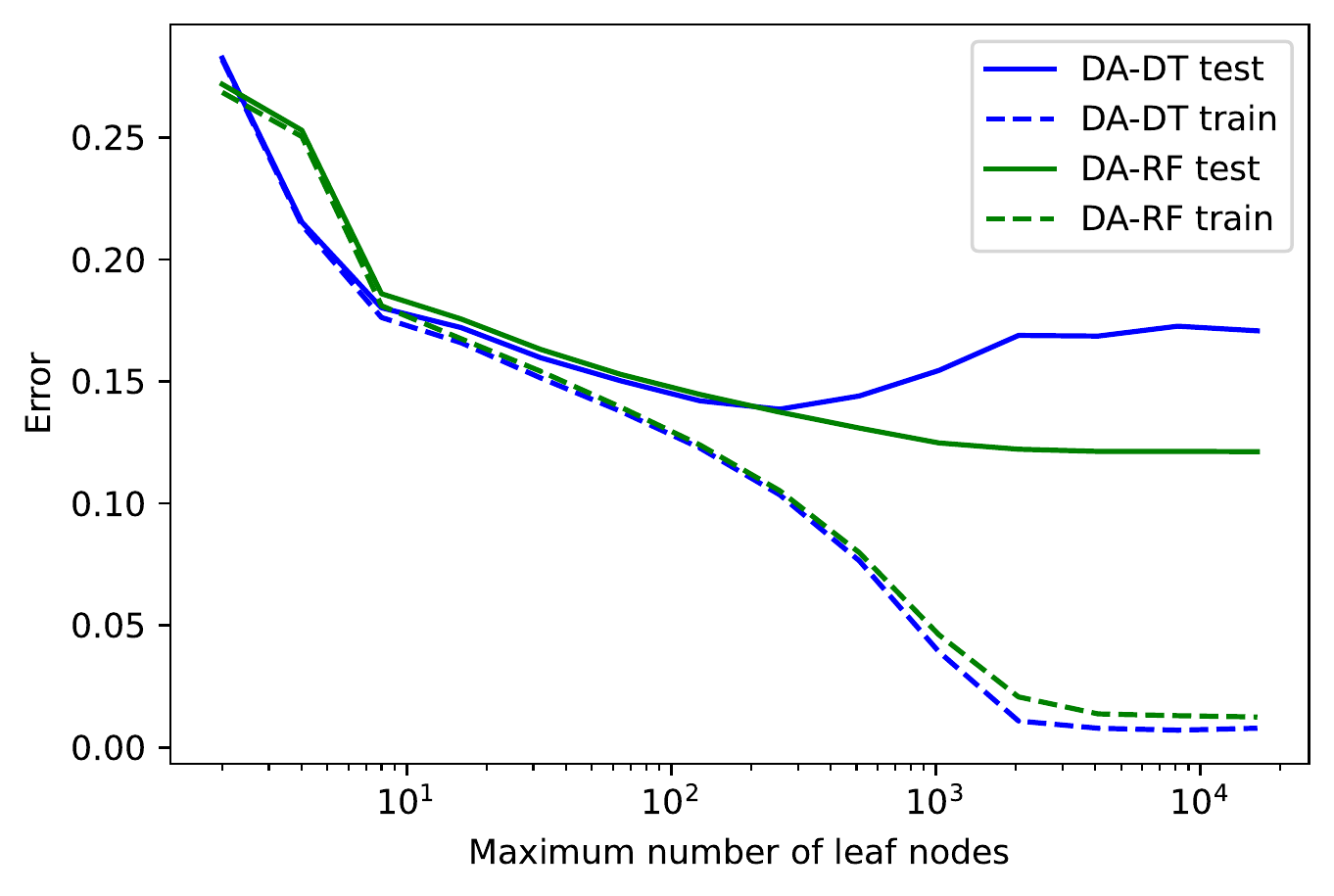}
    \makebox[20pt]{\raisebox{60pt}{\rotatebox[origin=c]{90}{Nomao}}}%
    \includegraphics[width=5.5cm, height=4cm]{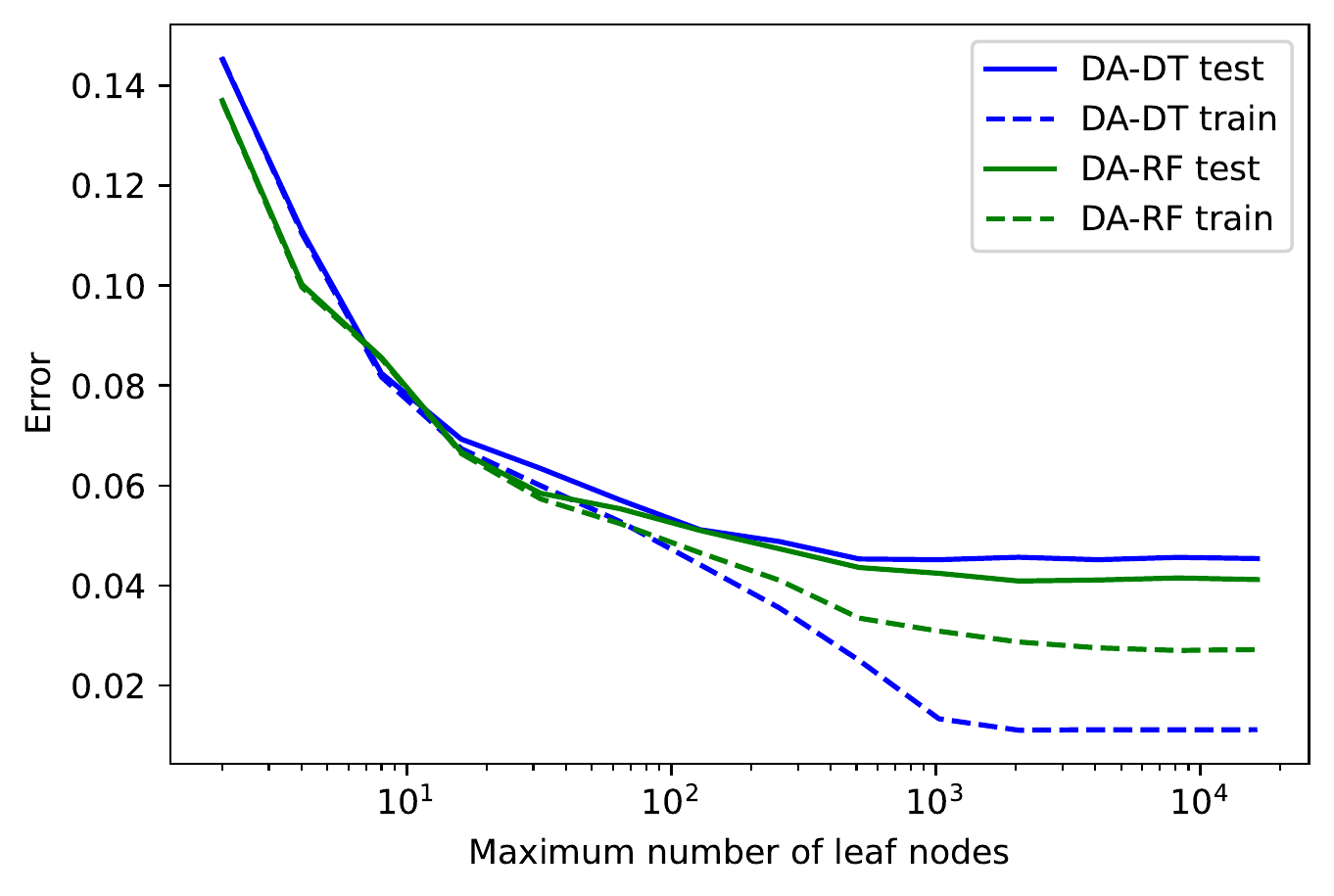}
    \caption{Test and training error.}\label{fig:2a}
\end{subfigure}
\begin{subfigure}[c]{0.30\textwidth}
    \includegraphics[width=5.5cm, height=4cm]{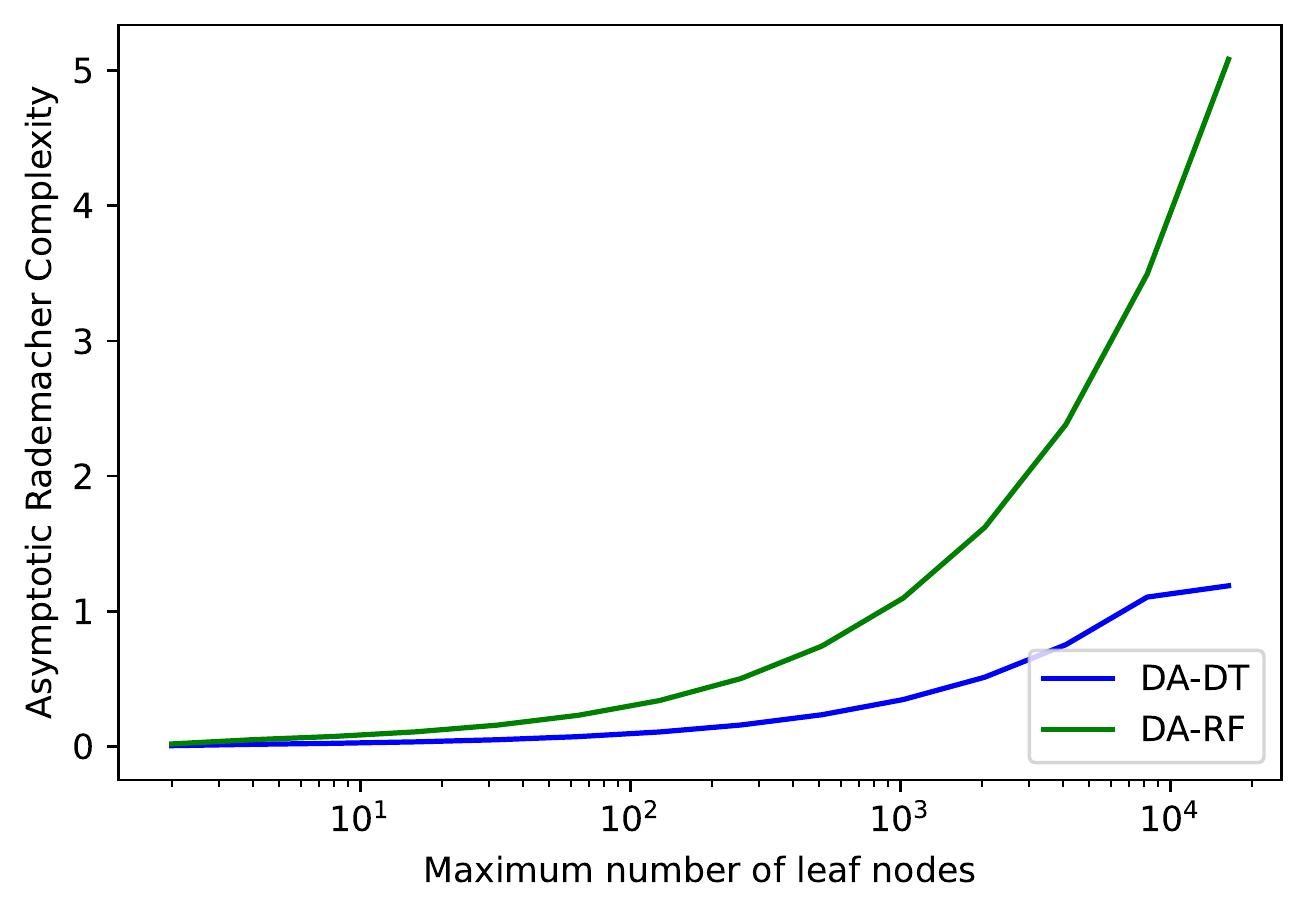}
    \includegraphics[width=5.5cm, height=4cm]{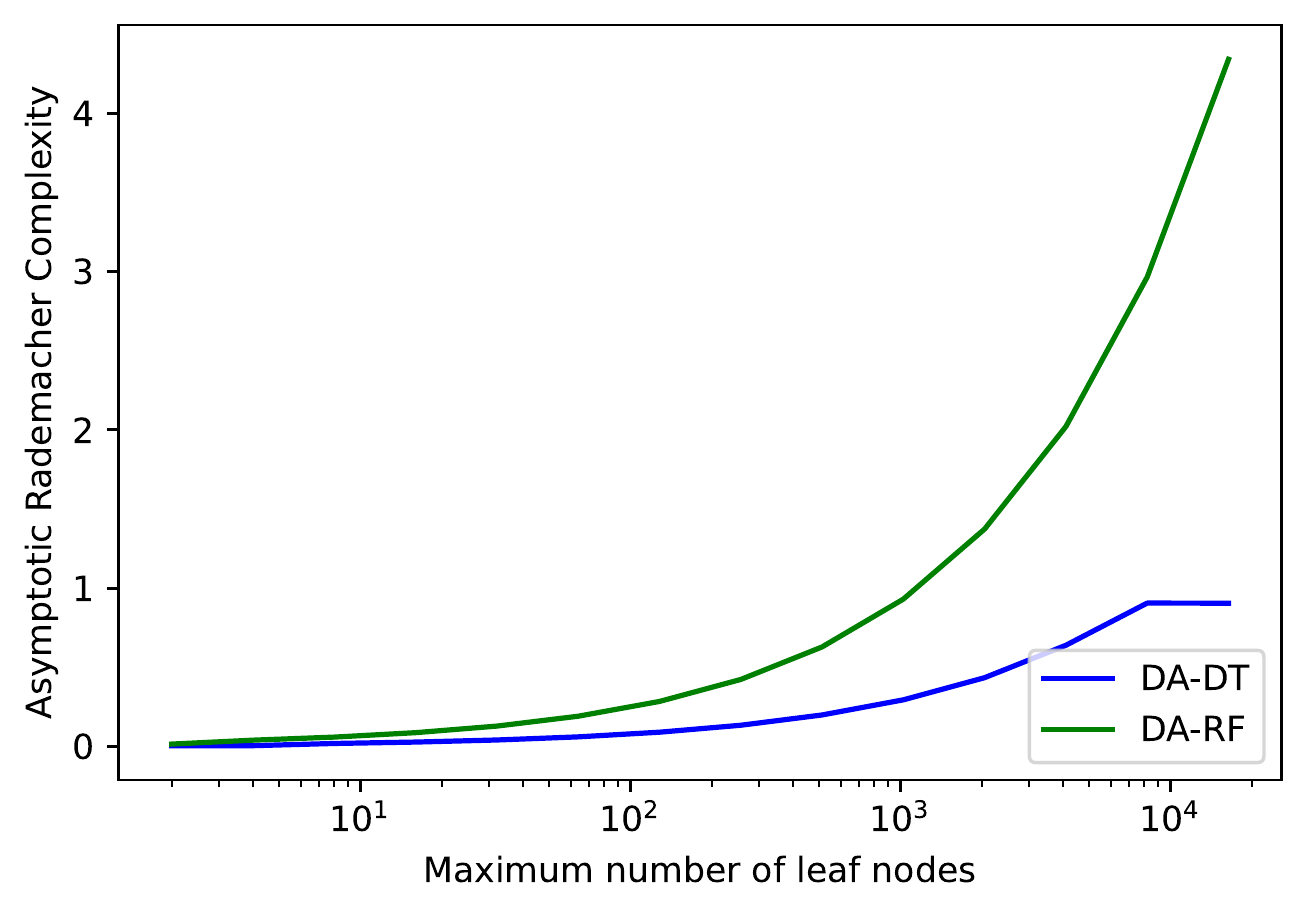}
    \includegraphics[width=5.5cm, height=4cm]{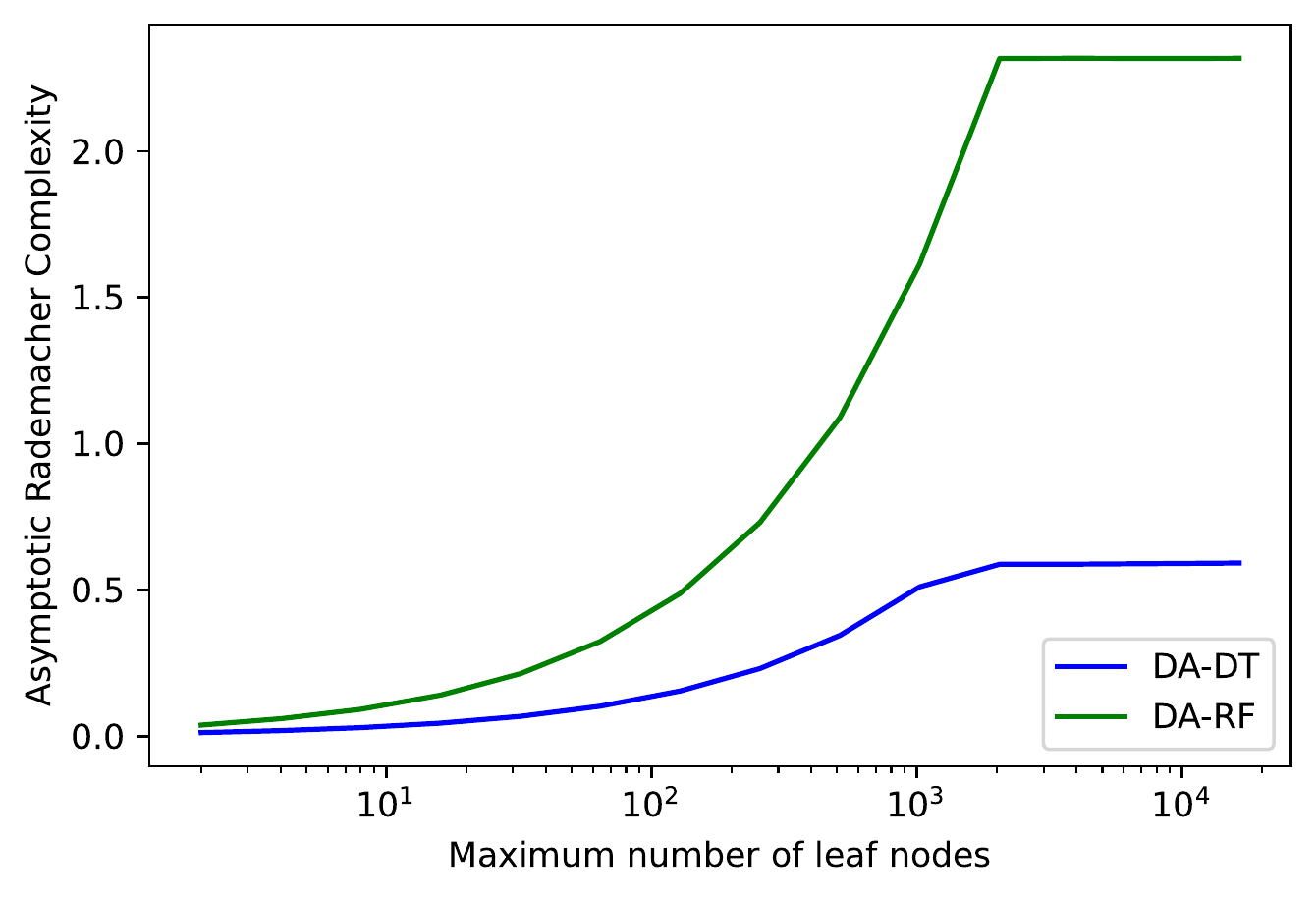}
    \includegraphics[width=5.5cm, height=4cm]{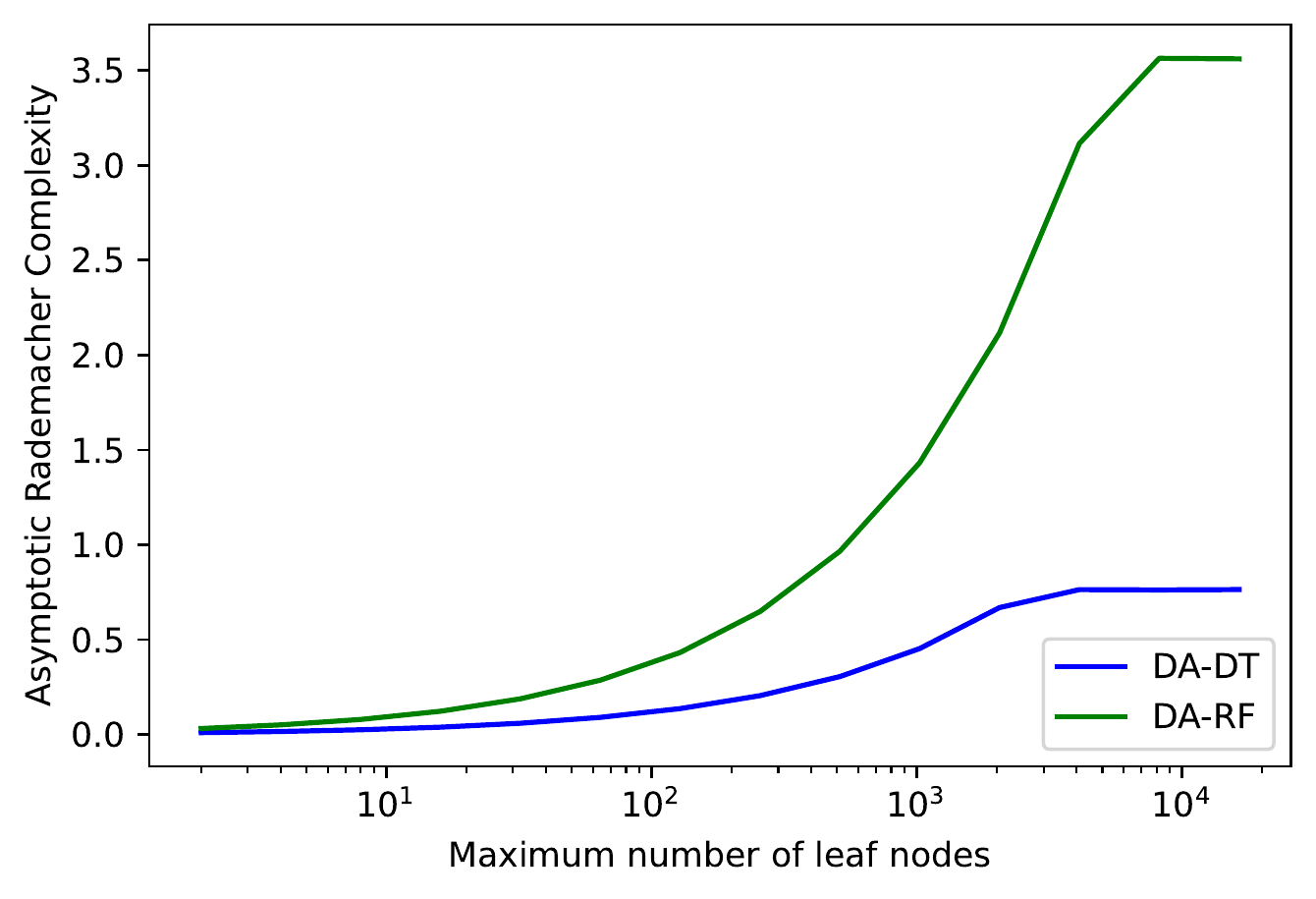}
    \includegraphics[width=5.5cm, height=4cm]{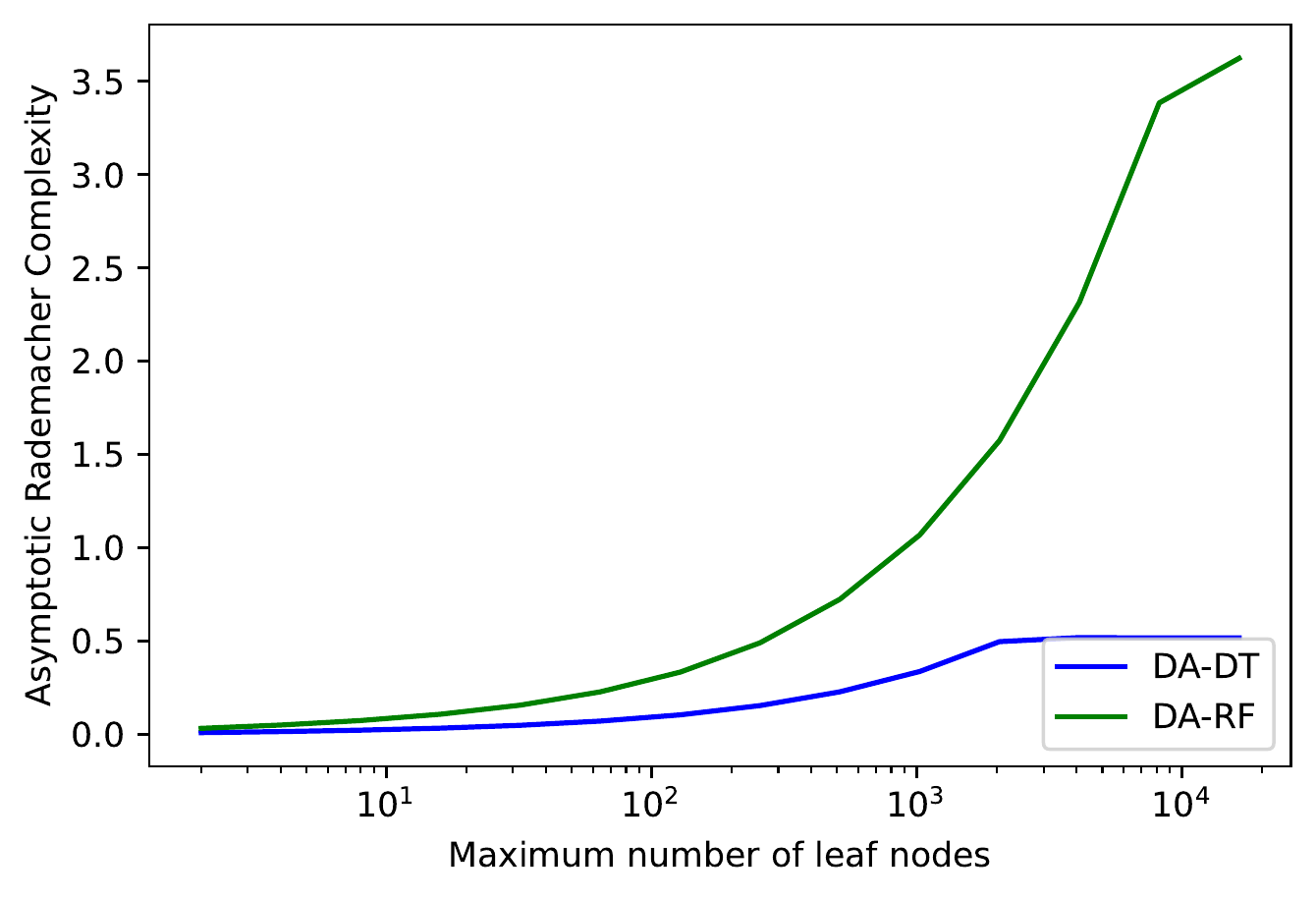}
    \caption{Average Rademacher Complexity.}\label{fig:2b}
\end{subfigure}
\begin{subfigure}[c]{0.30\textwidth}
    \includegraphics[width=5.5cm, height=4cm]{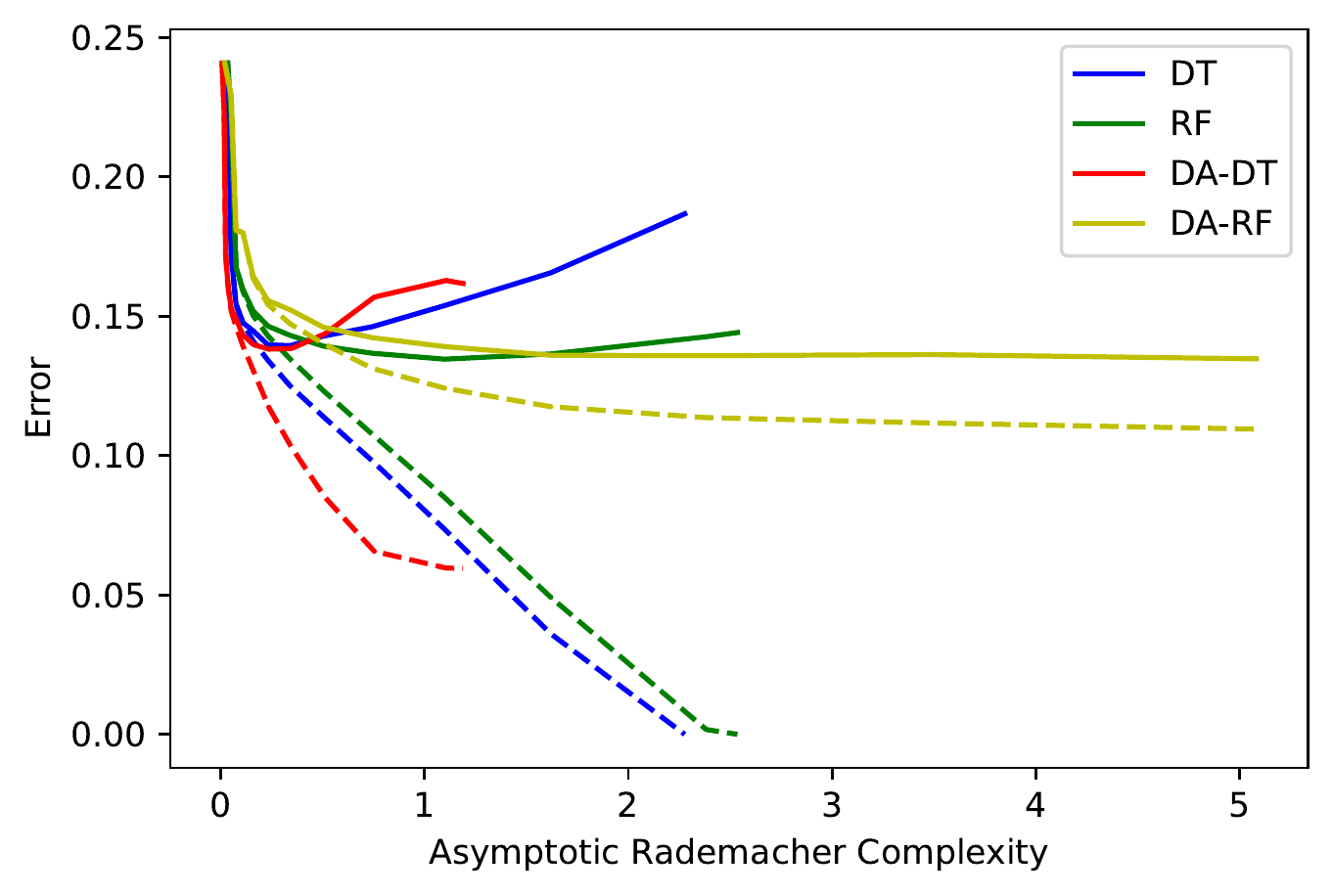}
    \includegraphics[width=5.5cm, height=4cm]{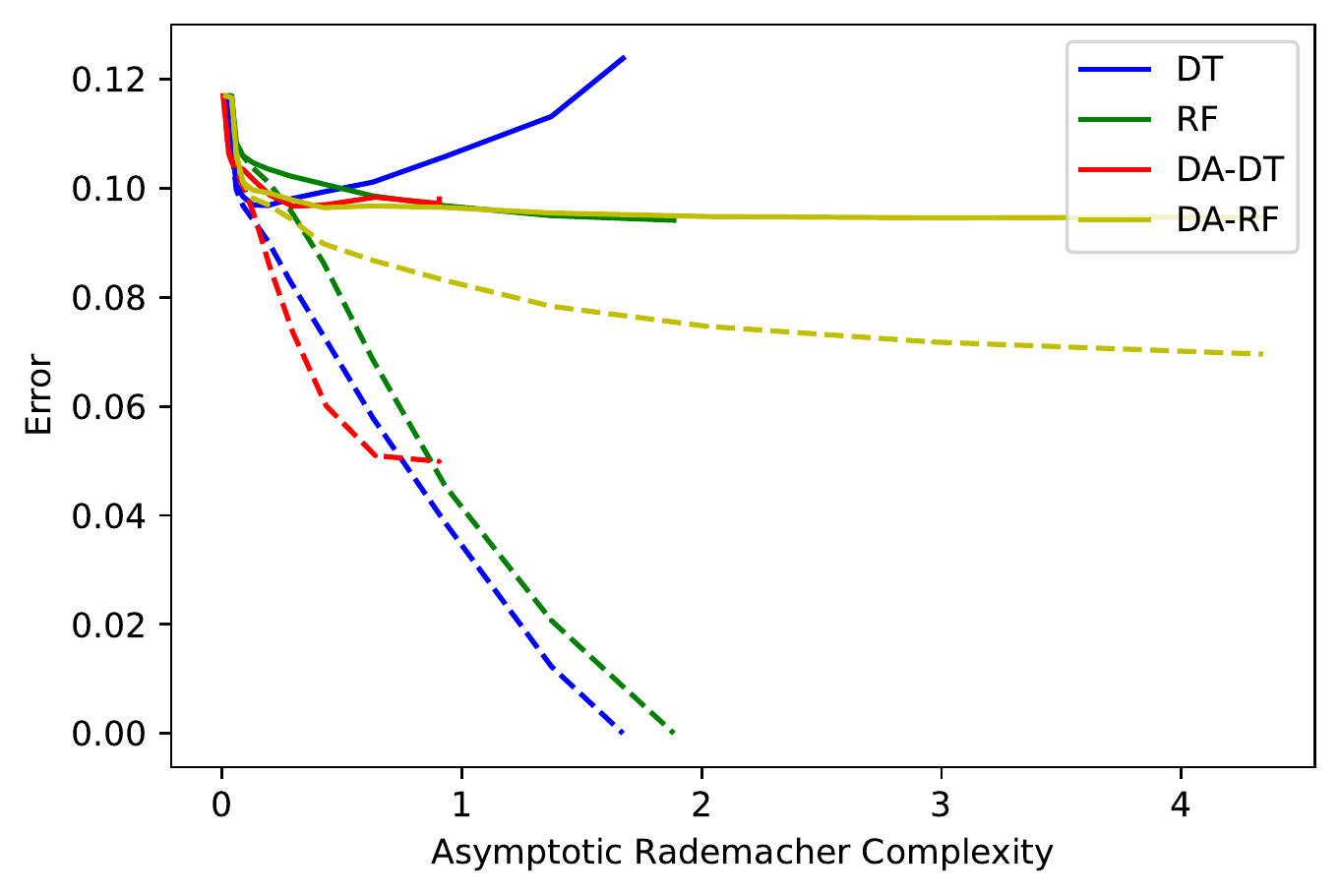}
    \includegraphics[width=5.5cm, height=4cm]{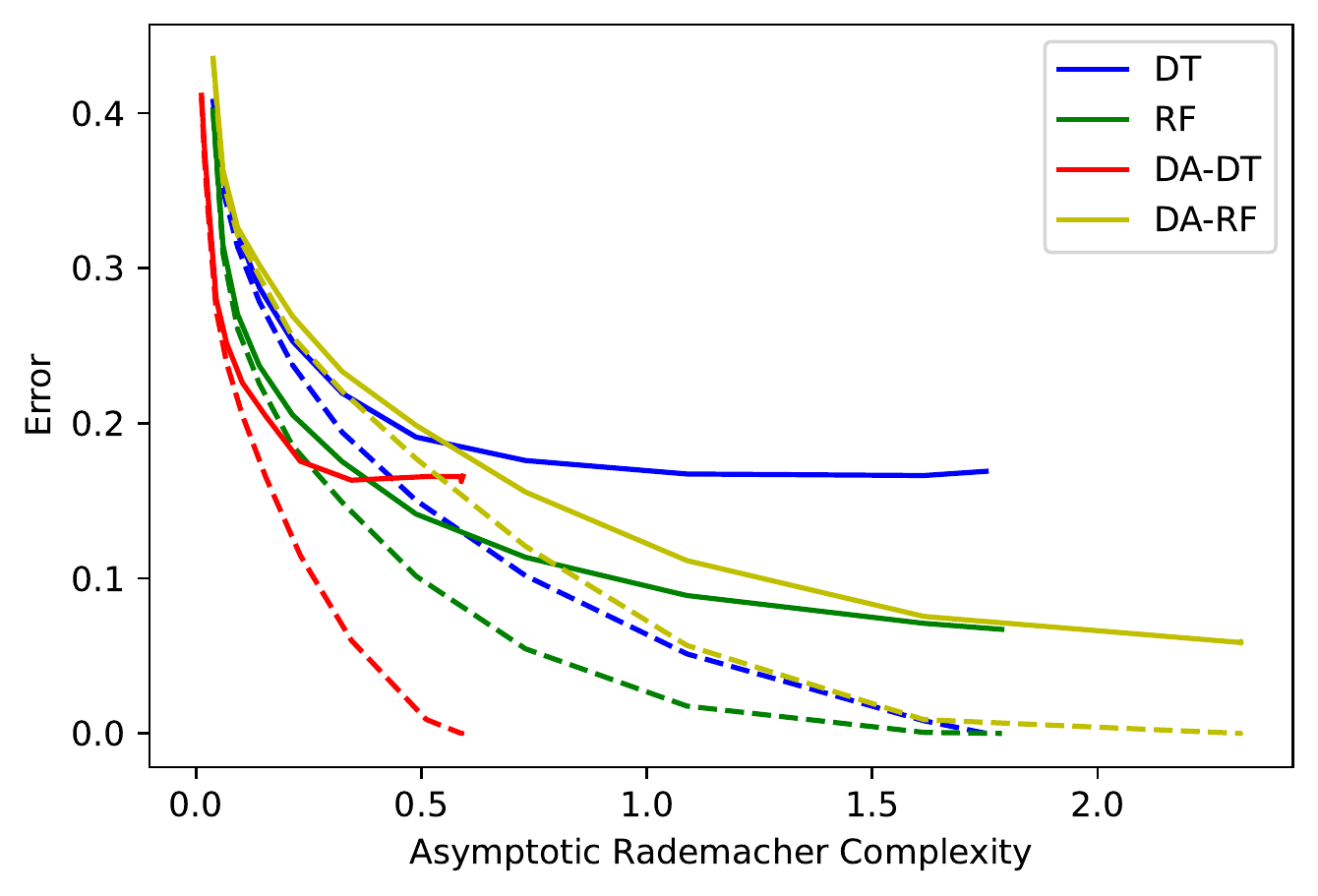}
    \includegraphics[width=5.5cm, height=4cm]{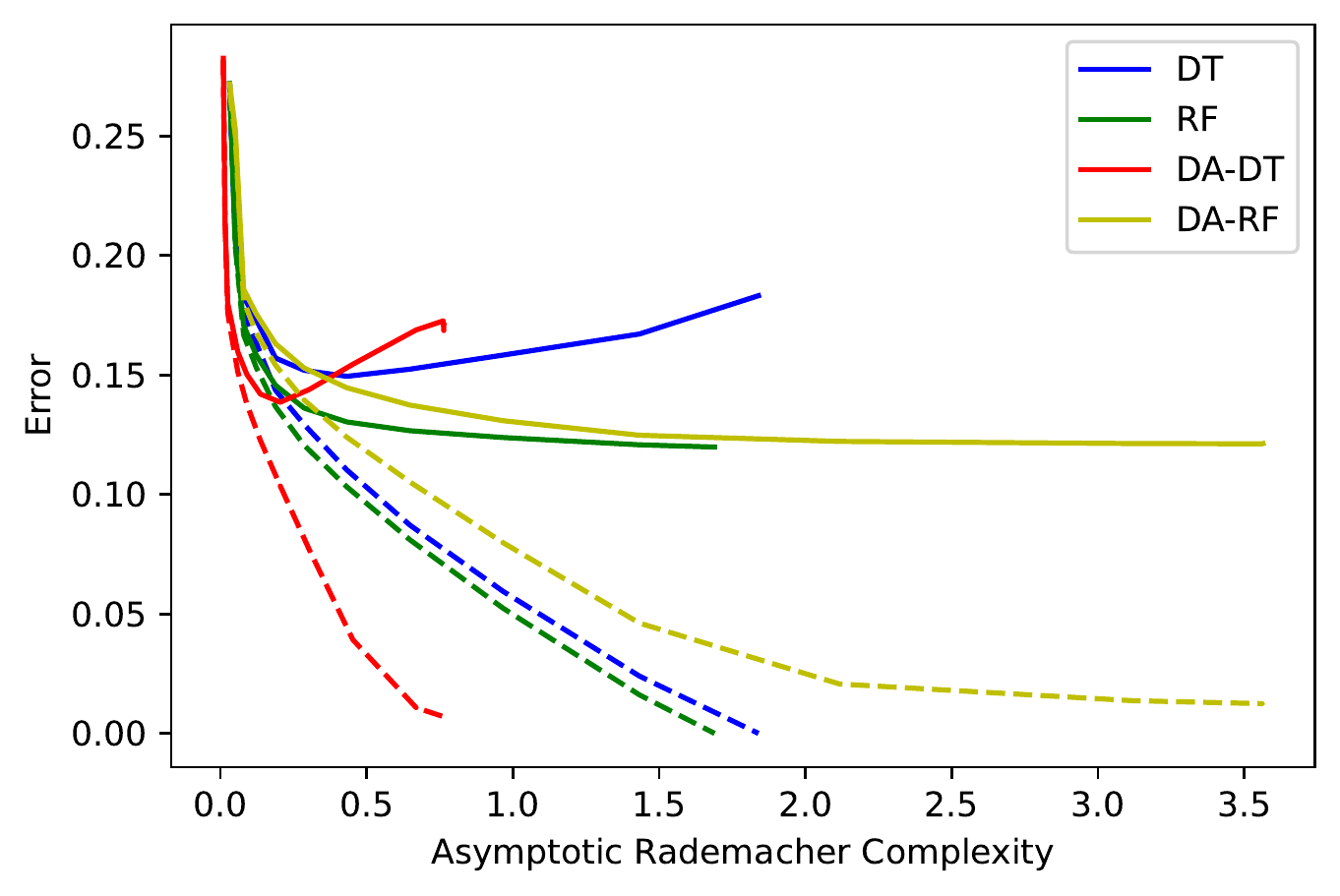}
    \includegraphics[width=5.5cm, height=4cm]{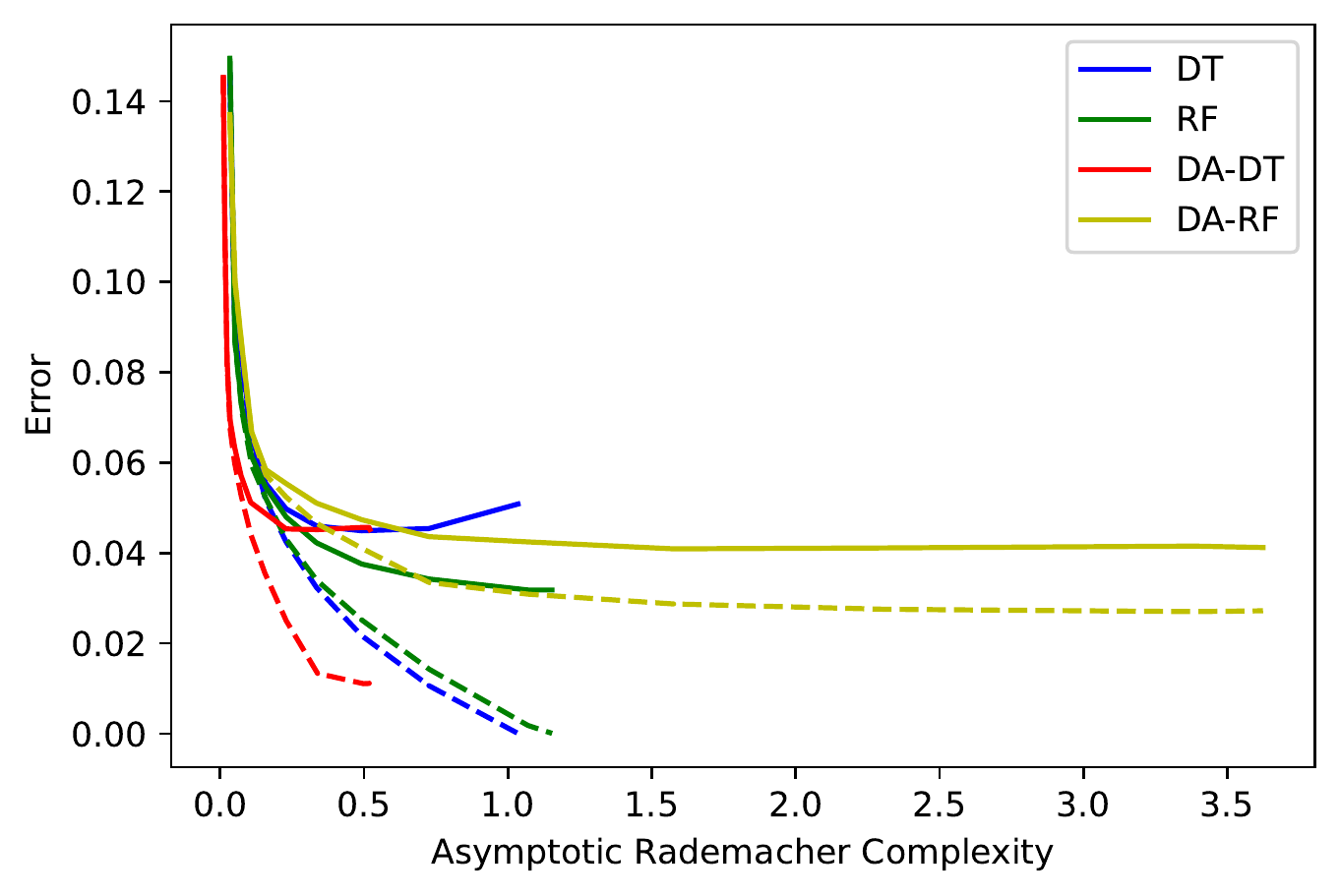}
    \caption{Test and training error of all methods.}\label{fig:2c}
\end{subfigure}

\caption{Test and training error of DA-RF and DA-DT (first column), the average Rademacher complexity (second column) and the test and training error of all methods. Each row depicts one dataset. Results are averaged over a 5-fold cross validation. Solid lines are the test error and dashed lines are the training error. Best viewed in color.} \label{fig:H2}
\end{figure*}

We repeat the above experiments with data augmentation training. Again we limit the maximum number of leaf nodes $n_l \in \{2,4,8,16,32,64,\dots,16384\}$. We train a RF with $M = 256$ trees and approximate it with a DT using $T = 10$ and $\varepsilon = 0.01$. We call this algorithm DT with Data Augmentation (DA-DT). Similar, we train a single DT and approximate it with a RF with $M=256$ trees, $T = 10$ and $\varepsilon = 0.01$ denoted as RF with Data Augmentation (DA-RF). Figure \ref{fig:2a} shows the error curves for this experiment. Again, note the logarithmic scale on the x-axis. First we see that the training error approaches zero for larger $n_l$ for both models as expected.
Second we see that the decision tree DA-DT despite fitting the decision boundary of a RF shows clear signs of overfitting. Third, and maybe even more remarkable, the forest DA-RF trained via data-augmentation on the \emph{bad, overfitted} labels from the DT still does \emph{not} overfit but also has a single descent. To gain a better picture we can again look at the Rademacher complexities of these two models in Figure \ref{fig:2b}. Similar to before there is a steep increase for both models. However, the DA-DT now converges against a \emph{smaller} Rademacher complexity compared to the DA-RF which now has a much \emph{larger} Rademacher complexity across all datasets despite the fact that DA-RF has a better test error. The forest does \emph{not} overfit in a u-shaped curved as expected but also shows a single descent whereas the DT still \emph{does} overfit in a u-shaped similar to before.
For a better comparison between the individual methods we combine them in a single plot. Figure \ref{fig:2c} shows the asymptotic Rademacher complexity over the test and train error of all methods. The dashed lines depict the training error, whereas the solid lines are the test error. Note that some curves stop early because their respective Rademacher complexities are not large enough to fill the entire plot.
As one can see, DT and RF have a comparably small maximum Rademacher complexity. RF seems to minimize the training error more aggressively and reaches a smaller error with smaller complexities, whereas DT starts to overfit comparably early. DA-DT seems to have the smallest Rademacher complexity but also overfits the most on some datasets (e.g. adult or magic). DA-RF has the largest complexity but does not seem to overfit at-all. It slowly converges against the original RF's performance, except on the adult dataset where it also does not overfit. Both, DT and DA-DT show a u-shaped curved whereas RF and DA-RF both show a single descent in most cases.
Clearly, the Rademacher complexity fails to explain the performance of the data augmented trees and forests. We argue that the reason for this lies in the \emph{algorithm} used to train the trees and \emph{not} in the model-class itself. 

\section{Negative Correlation Forests}
\label{sec:bias-variance}

The previous section implies that the model alone does not fully explain its performance. We argue that the learning algorithm also plays a crucial role in the generalization capabilities of the model and more specifically that the trade-off between bias and diversity plays a crucial role. We show that there is a large region of different diversity levels which are all equally good and it does not matter what specific trade-off is achieved as long as it falls into the same region. 

More formally, the bias-variance decomposition of the mean-squared-error $(f(x)-y)^2$ states that the expected error of a model $f$ can be decomposed into its bias and variance \cite{Markowitz/1952,Geman/etal/1992}: 
$$
\E{\mtiny{\substack{\theta \sim \Theta \\ x,y \sim \mathcal D}}}{f^{\theta}(x) - y)^2} = (\E{\mtiny{\substack{\theta \sim \Theta \\ x,y \sim \mathcal D}}}{f^{\theta}(x)} - y)^2 + \V{\mtiny{\substack{\theta \sim \Theta}}}{f^{\theta}(x)} 
$$
where $\Theta$ is a random process (e.g. due to bootstrap sampling.) induced by the algorithm that generates $f^{\theta}$,  $(\E{\mtiny{\substack{\theta \sim \Theta \\ x,y \sim \mathcal D}}}{f^{\theta}(x)} - y)^2$ is the bias of the algorithm and $\V{\mtiny{\substack{\theta \sim \Theta}}}{f^{\theta}(x)} $ is its variance. Considering the ensemble $f(x) = \frac{1}{M} \sum_{i=1}^M h_i(x)$, then the variance term can be further decomposed into a co-variance (see e.g. \cite{Brown/etal/2005} for more details):
\begin{equation}
(f(x) - y)^2 = \frac{1}{M}\sum_{i=1}^M (h_i(x) - y)^2 - \cov_i\left[{h_i(x)}\right]
\end{equation}
where we dropped $\theta$ for a better readability and $\cov_i$ is the co-variance of the predictions across the ensemble.

This decomposition does not directly relate the training error of a model to its generalization capabilities, but it shows how the individual training and testing losses are structured \cite{buschjager/etal/2020}. Although suspected for some time and exploited in numerous ensembling algorithms, the exact connection between the diversity of an ensemble and its generalization error was only established relatively recently. Germain et al. showed in \shortcite{germain/etal/15} that the diversity of an ensemble and its generalization error can be connected in a PAC-style bound shown in Theorem \ref{th:Cbound}.

\begin{theorem}[PAC-Style C-Bound \cite{germain/etal/15}]
	\label{th:Cbound}
	Let $\mathcal H = \bigcup_{j=1}^k \mathcal H_{j}$ denote a set of base classifiers and let $f = \frac{1}{M}\sum_{i=1}^M h_i$ be the ensemble. Then, for any $\delta > 0$, with probability at least $1-\delta$ over the choice of sample $\mathcal{S}$ of size $N$ drawn i.i.d. according to $\mathcal D$, the following inequality holds:
	\begin{align*}
    L_0(f) 
    &\le 1 - \frac{ \left( \max\left(0, \frac{1}{M} \sum\limits_{i=1}^M L_{0,\mathcal S}(h_i) - \tau_1(N,\delta)\right) \right)^2 }{
        \min\left(1, \cov_{i,\mathcal S}\left[h_i\right]+ \tau_2(N,\delta)\right)
    } 
    \end{align*}
    where $\tau_1(N,\delta), \tau_2(N,\delta) \to 0$ for $N \to \infty$ and any $\delta$
    and where $\cov_{i,\mathcal S}\left[h_i\right]$ is the co-variance of the ensemble evaluated on the sample $S$.
\end{theorem}
Intuitively, this result shows that an ensemble of powerful learners with a small bias that sometimes disagree will be better than a an ensemble with a comparable bias in which all models agree. 

While the original RF algorithm produces accurate ensembles its diversity is implicitly determined by the bootstrap and features samples. Hence it is difficult to precisely control its diversity. For a direct control over the diversity we will now introduce Negative Correlation Forest. Bound \ref{th:Cbound} is stated for the 0-1 loss which makes the direct minimization of this bound difficult as noted in \cite{germain/etal/15}. Luckily, the minimization over the bias-variance decomposition of the MSE is much more approachable and has already been studied in the form of Negative Correlation Learning (NCL). NCL offers a fine grained control over the diversity of an ensemble of neural networks by minimizing the following objective (see  \cite{Brown/etal/2005,Webb/etal/2019,buschjager/etal/2020} for more details):
\begin{equation}
    \label{eq:GNCL1}
    \ell_{\lambda}(f(x), y) := \frac{1}{M} \sum_{i=1}^M (h_i(x)-y)^2 - \frac{\lambda}{2M} \sum_{i=1}^M {d_i}^T D d_i
\end{equation}
where $d_i = (h_i(x) - f(x))$, $D = 2 \cdot I_C$ is the $C\times C$ identity matrix with $2$ on the main diagonal and $\lambda \in \mathbb R_+$ is the regularization strength. For $\lambda = 0$ this trains $M$ classifier independently and no further diversity among the ensemble members is enforced, for $\lambda > 0$ more diversity is enforced during training and for $\lambda < 0$ diversity is discouraged. While a large diversity helps to reduce the overall error it also implies that some trees must be wrong on some data points and thus means that their respective bias again increases. Finding a good balance between the bias and diversity is therefore crucial for a good performance. 


NCL was first introduced to train ensembles of neural networks because they can easily be optimized via gradient-based algorithms minimizing Eq. \ref{eq:GNCL1}. We adapt this approach to RF by training an initial RF with algorithm \ref{fig:RF} which is then refined by optimizing the NCL objective:
Recall that DTs use a series of axis-aligned splits of the form $\mathbbm{1}\{x_k \le t\}$ and $\mathbbm{1}\{x_k > t\}$ where $k$ is a feature index, $t$ is a threshold to determine the leaf nodes and each leaf node  has a (constant) prediction $\widehat y_{i,l} \in \mathbb R^C$ associated with it.
Let $\beta_i$ be the parameter vector of tree $h_i$ (e.g. containing split values, feature indices and leaf-predictions) and let $\beta = (\beta_1,\dots,\beta_M)$ be the parameter vector of the entire ensemble $f_{\beta}$. Then our goal is to solve
\begin{equation}
\arg\min_{\beta} \frac{1}{N}\sum_{(x,y) \in \mathcal S} \ell_{\lambda}\left(f_{\beta}(x),y\right)
\end{equation}
for a given trade-off $\lambda$. We propose to minimize this objective via stochastic gradient-descent. SGD is an iterative algorithm which takes a small step into the negative direction of the gradient in each iteration $t$ by using an estimation of the true gradient
\begin{equation}
\label{eq:sgd}
\beta^{t+1} \gets \beta^t - \alpha^t g_{\mathcal B}(\beta^t) 
\end{equation}
where 
\begin{equation}
g_{\mathcal B}(\beta^t) = \nabla_{\beta^t} \left( \sum_{(x,y) \in \mathcal B} \ell_{\lambda} \left(f_{\beta^t}(x),y\right) \right)
\end{equation}
is the gradient of $\ell$ wrt. to $\beta^t$ computed on a mini-batch $\mathcal B$.

Unfortunately, the axis-aligned splits of a DT are not differentiable and thus it is difficult to refine them further with gradient-based approaches. However, the leaf predictions $\widehat y_{i,l}$ are simple constants that can easily be updated via SGD. 
Formally, we use $\beta_i = (\widehat y_{i,1},\widehat y_{i,2}, \dots)$ leading to
\begin{equation}
\label{eq:grad}
g_{\mathcal B}(\beta^t_i) = \frac{1}{|\mathcal B|} \left( \sum_{(x,y)\in\mathcal B} \frac{\partial \ell_{\lambda}(f_{\beta^t}(x), y)}{\partial f_{\beta^t}(x)} w_i s_{i,l}(x)\right)_{l=1,2,\dots,L_i}
\end{equation}

\begin{algorithm}
	\begin{algorithmic}[1]
	\State{$h \gets \texttt{train\_rf}(M, n_l,d_i)$} \Comment{\parbox[t]{0.39\linewidth}{Algorithm \ref{fig:RF}}}
	\State{$w \gets (1/M, \dots, 1/M)$} \Comment{\parbox[t]{0.39\linewidth}{Use constant weights}}
	\For{$i=1,\dots,M$} \Comment{\parbox[t]{0.39\linewidth}{Init. leaf predictions}}
	    \State{$\beta_i \gets (\widehat y_{i,1},\widehat y_{i,2}, \dots)$}
	\EndFor
	\For{receive batch $\mathcal B$} \Comment{\parbox[t]{0.39\linewidth}{Perform SGD}}
      \For{$i=1,\dots,M$}
        \State{$\beta^t_i \gets \beta^t_i - \alpha^t g_{\mathcal B}(\beta^t_i)$} \Comment{\parbox[t]{0.39\linewidth}{Using Eq. \ref{eq:sgd} + Eq. \ref{eq:grad}}}
      \EndFor
    \EndFor
	\end{algorithmic}
	\caption{Negative Correlation Forest (NCForest).}
	\label{fig:NCForest}
\end{algorithm}

Algorithm \ref{fig:NCForest} summarizes the NCForest algorithm. First an initial RF is trained with $M$ trees using at most $n_l$ leaf nodes and $d_i$ features. Then, the leaf-predictions are extracted from the forest in line $4$ and SGD is performed in line $5$ to $7$. Given our previous experiments, the learning algorithm seems to play a crucial role in the performance of a model and Theorem \ref{th:Cbound} implies that a diverse ensemble should generalize better. However, too much diversity will also likely hurt the performance of the forest because then the bias increases.

\subsection{There is an optimal trade-off between bias and diversity}

Again we validate our hypothesis experimentally. We train an initial RF with $M=256$ trees with a maximum number of $n_l = 4096$ leaf nodes. Due to the bootstrap sampling and due to the feature sampling, this initial RF already has some diversity. Hence, we use negative $\lambda$ values to de-emphasize diversity and positive $\lambda$ values to emphasize diversity. Last, we noticed that between $\lambda = 1.0$ and $\lambda = 1.005$ there is a steep increase in the diversity because it starts to dominate the optimization (c.f. \cite{Brown/etal/2005} which reports a similar effect for Neural Networks). Hence we vary $\lambda \in \{-20,-19.9,-19.8,\dots,1.0,1.001,1.002,1.003,1.004,1.005\}$ and minimize the NCL objective over $50$ epochs using the Adam optimizer with a step size of $0.001$ and a batch size of $64$ implemented in PyTorch\cite{Paszke/etal/2019}. 
We also experimented with more leaf nodes, different $\lambda$ values and more epochs but the test-error would not improve further with different parameters.

\begin{figure*}

\begin{subfigure}[c]{\dimexpr0.30\textwidth+20pt\relax}
    \makebox[20pt]{\raisebox{60pt}{\rotatebox[origin=c]{90}{Adult}}}%
    \includegraphics[width=5.5cm, height=4cm]{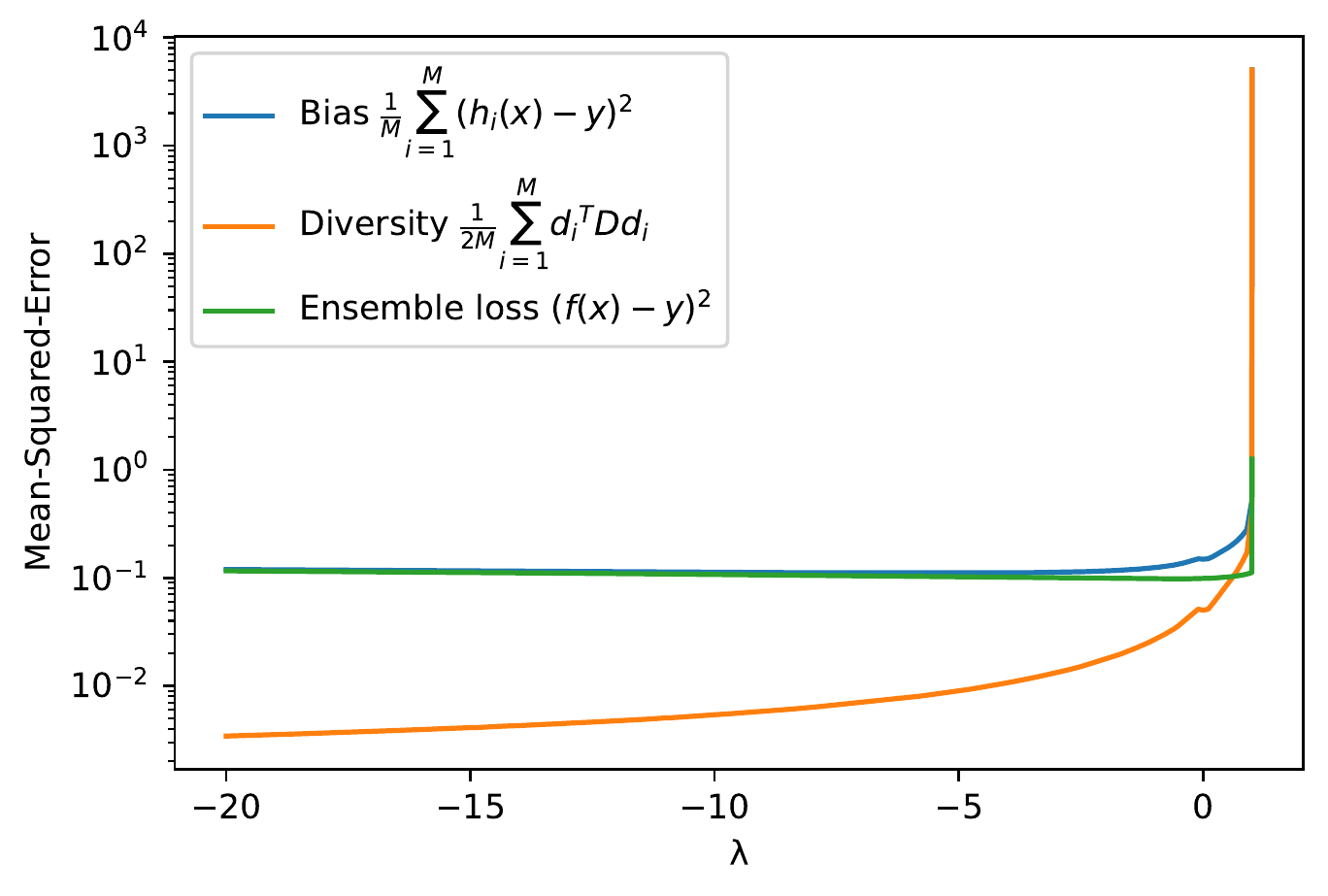}
    \makebox[20pt]{\raisebox{60pt}{\rotatebox[origin=c]{90}{Bank}}}%
    \includegraphics[width=5.5cm, height=4cm]{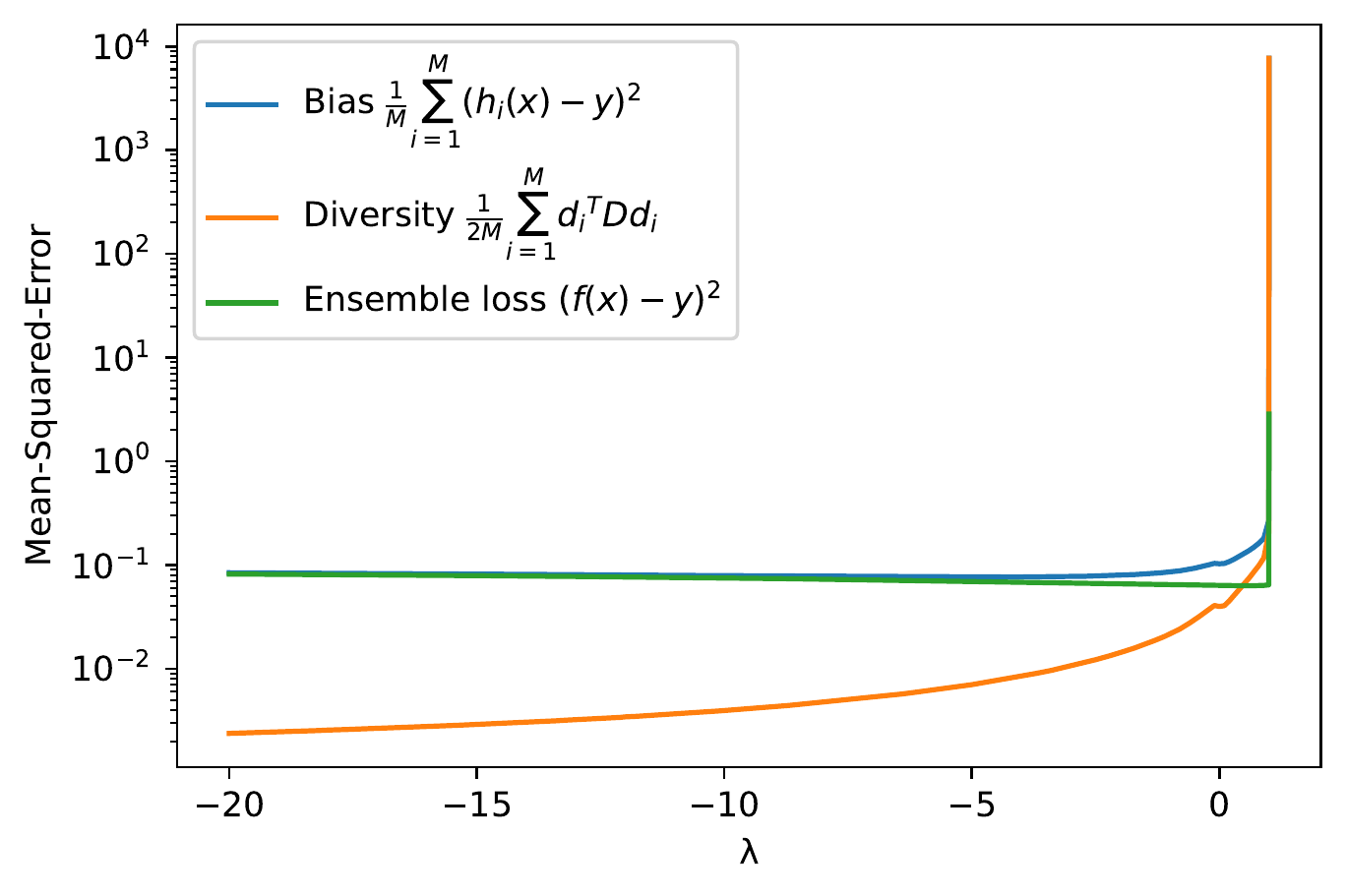}
    \makebox[20pt]{\raisebox{60pt}{\rotatebox[origin=c]{90}{EEG}}}%
    \includegraphics[width=5.5cm, height=4cm]{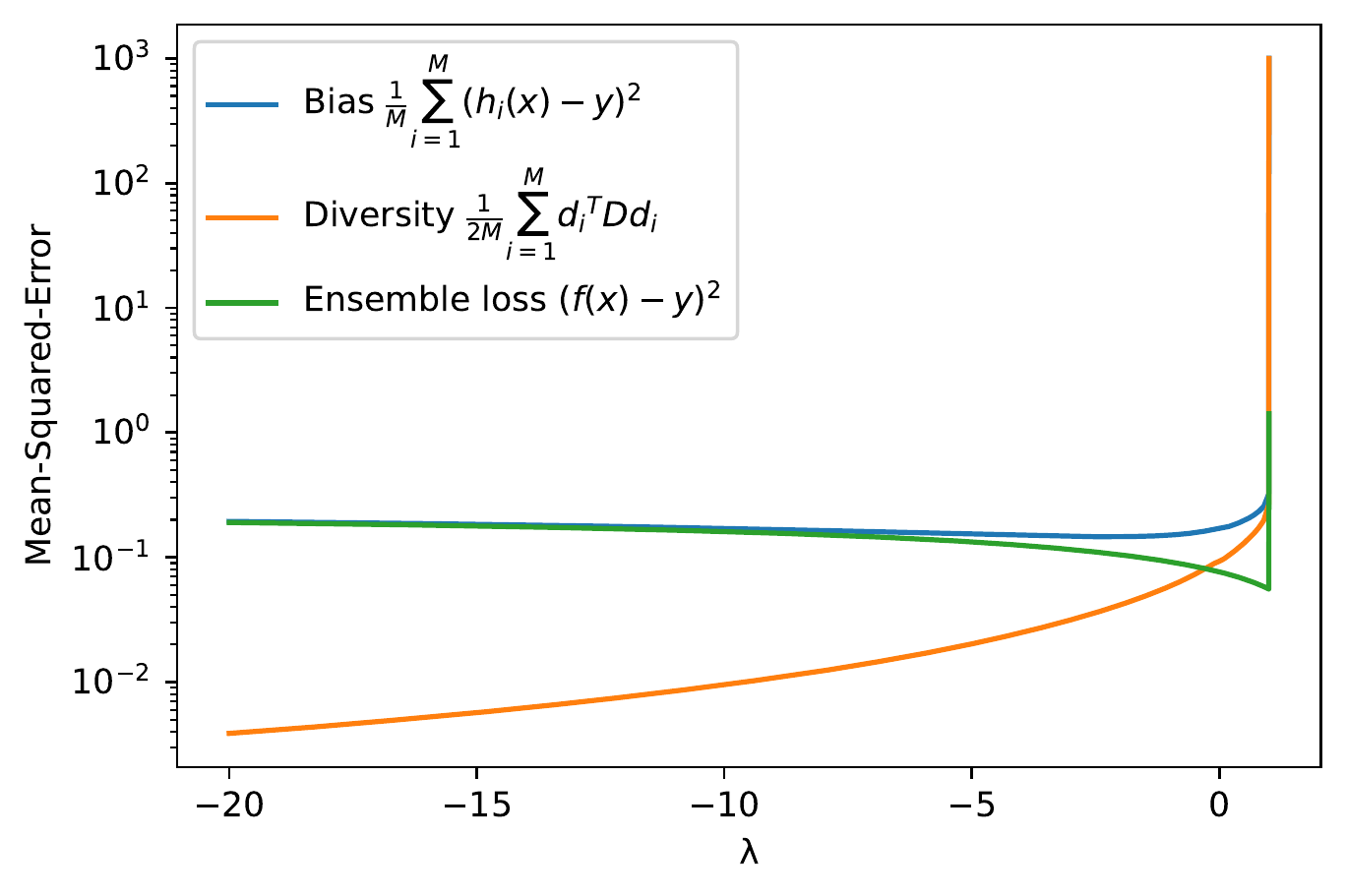}
    \makebox[20pt]{\raisebox{60pt}{\rotatebox[origin=c]{90}{Magic}}}%
    \includegraphics[width=5.5cm, height=4cm]{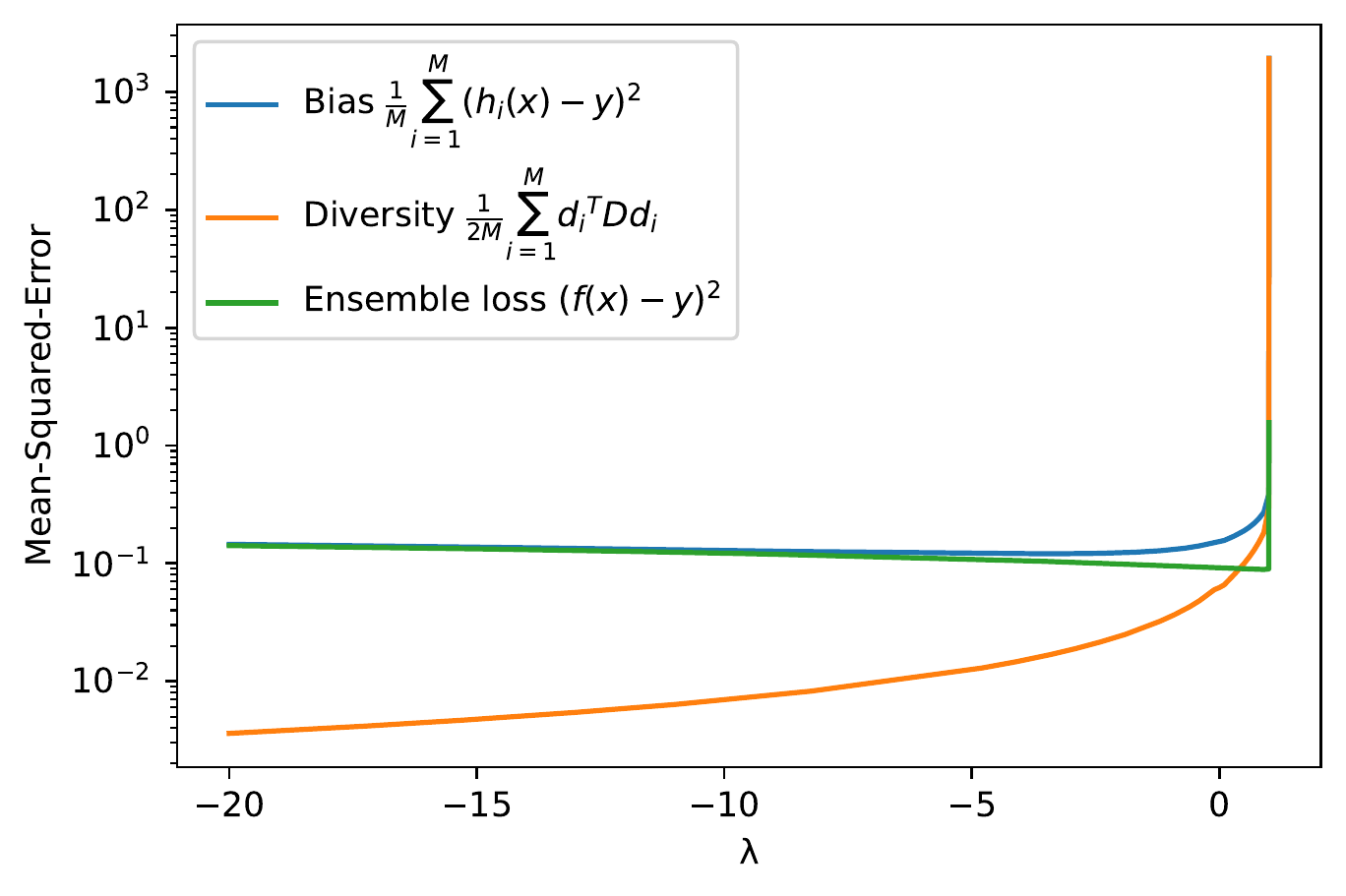}
    \makebox[20pt]{\raisebox{60pt}{\rotatebox[origin=c]{90}{Nomao}}}%
    \includegraphics[width=5.5cm, height=4cm]{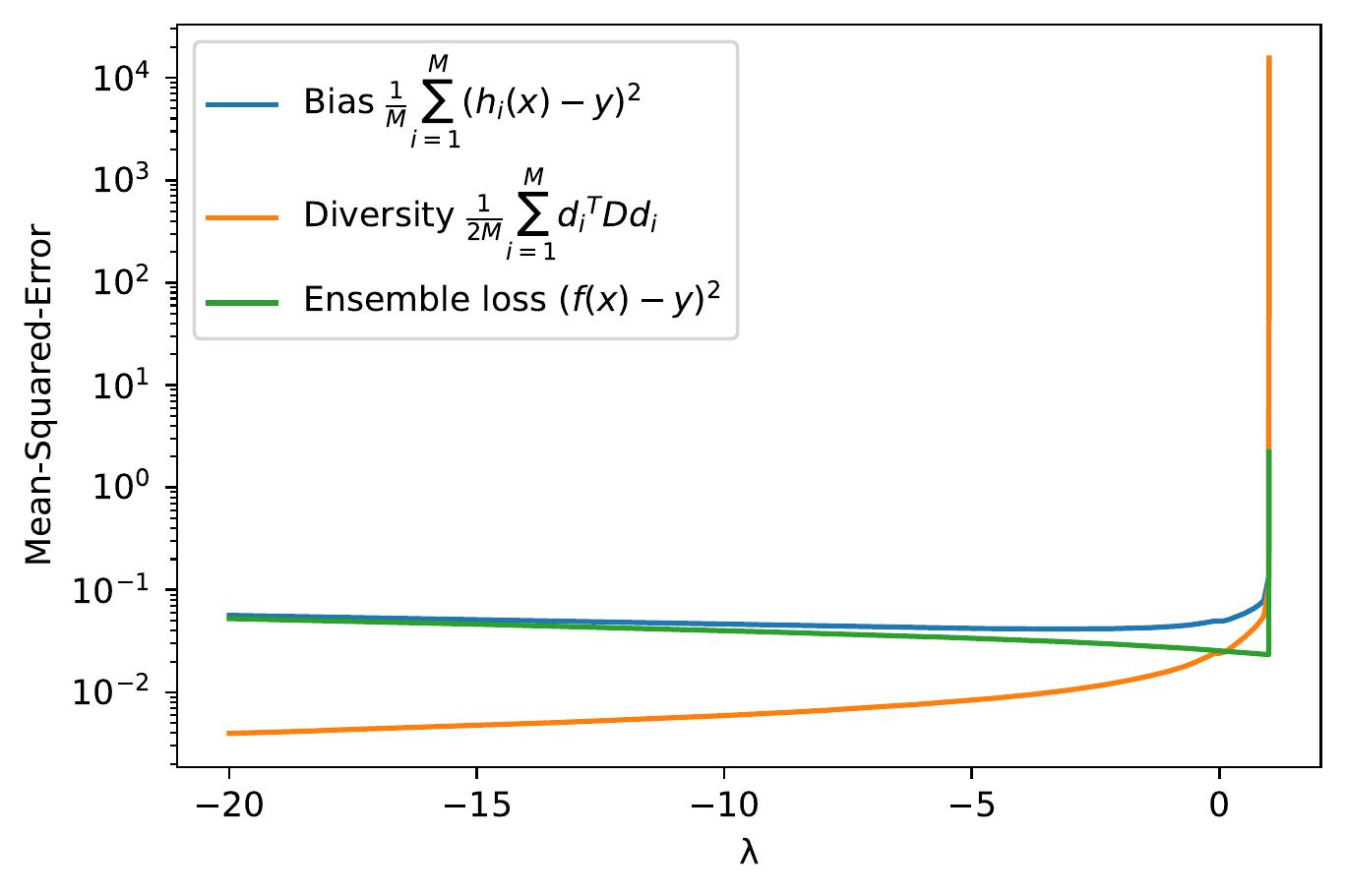}
    \caption{Mean Squared Error.}\label{fig:3a}
\end{subfigure}
\begin{subfigure}[c]{0.30\textwidth}
    \includegraphics[width=5.5cm, height=4cm]{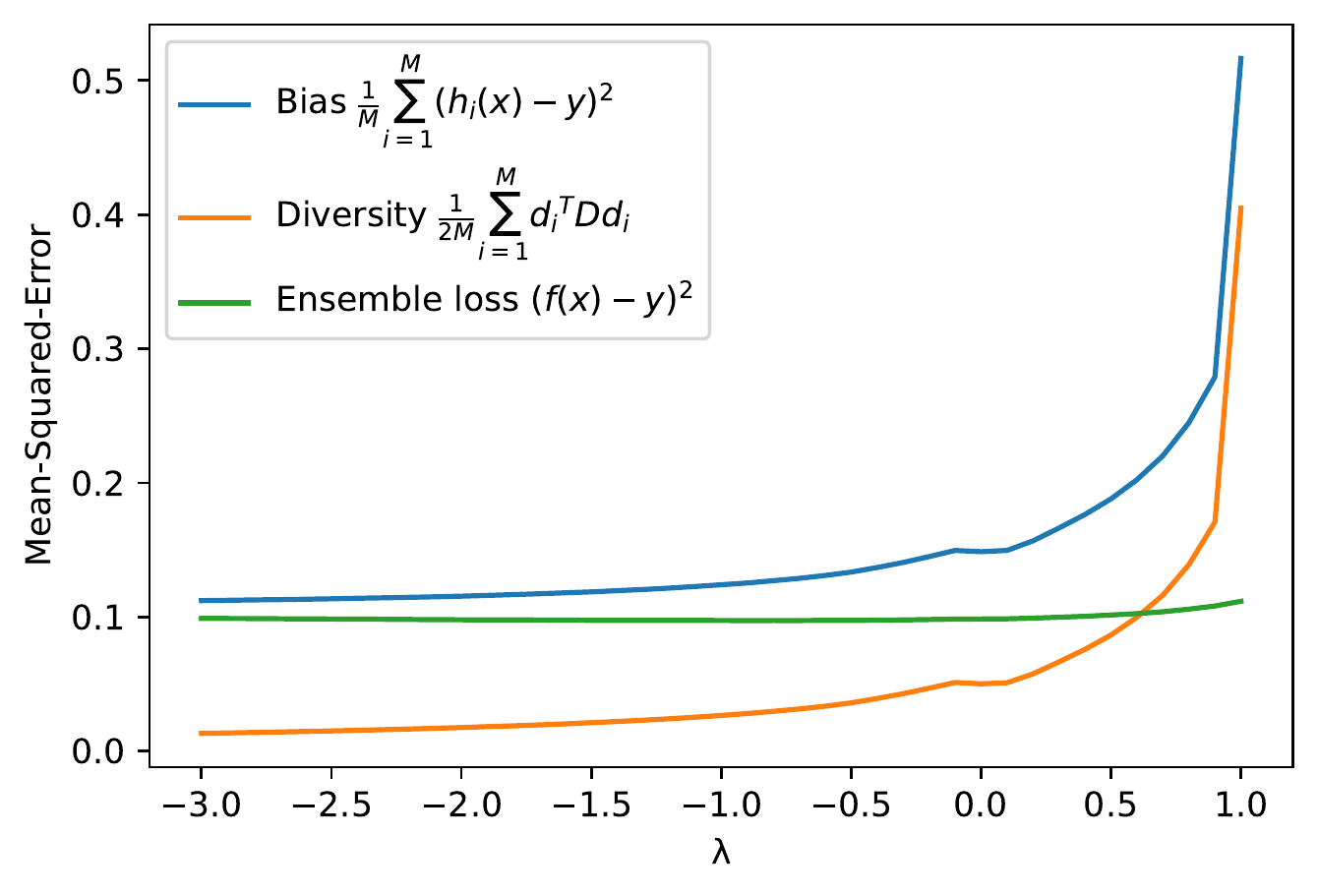}
    \includegraphics[width=5.5cm, height=4cm]{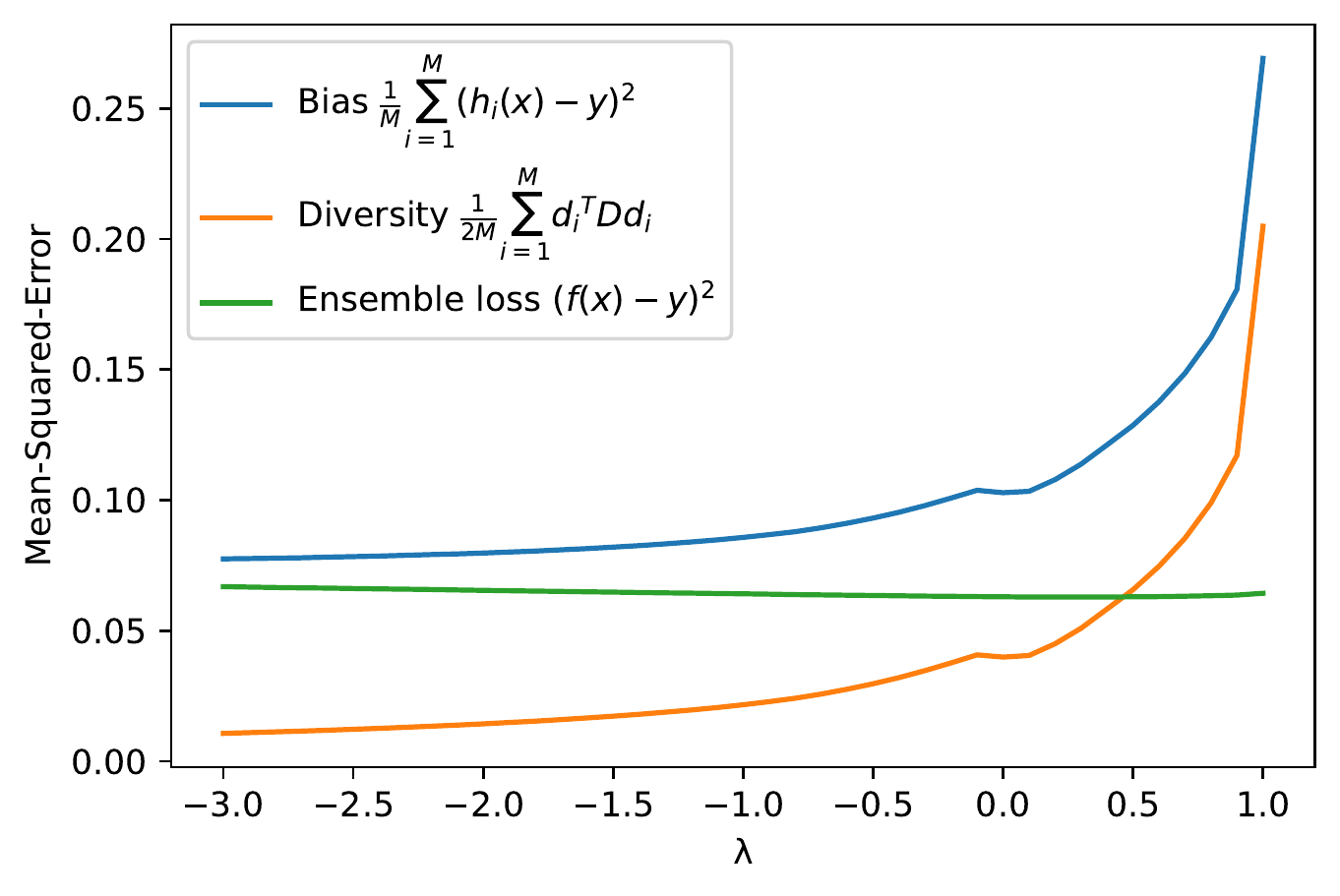}
    \includegraphics[width=5.5cm, height=4cm]{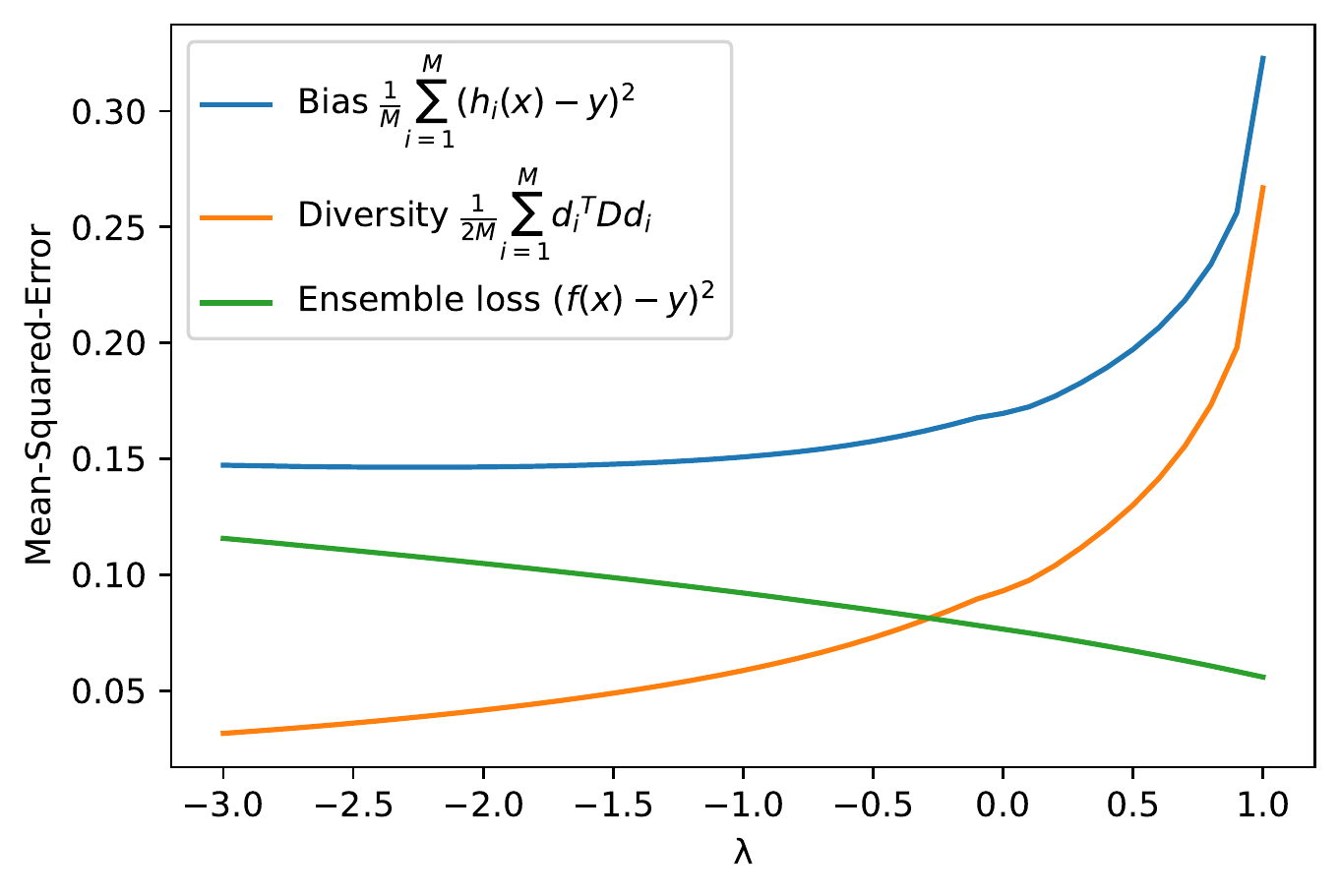}
    \includegraphics[width=5.5cm, height=4cm]{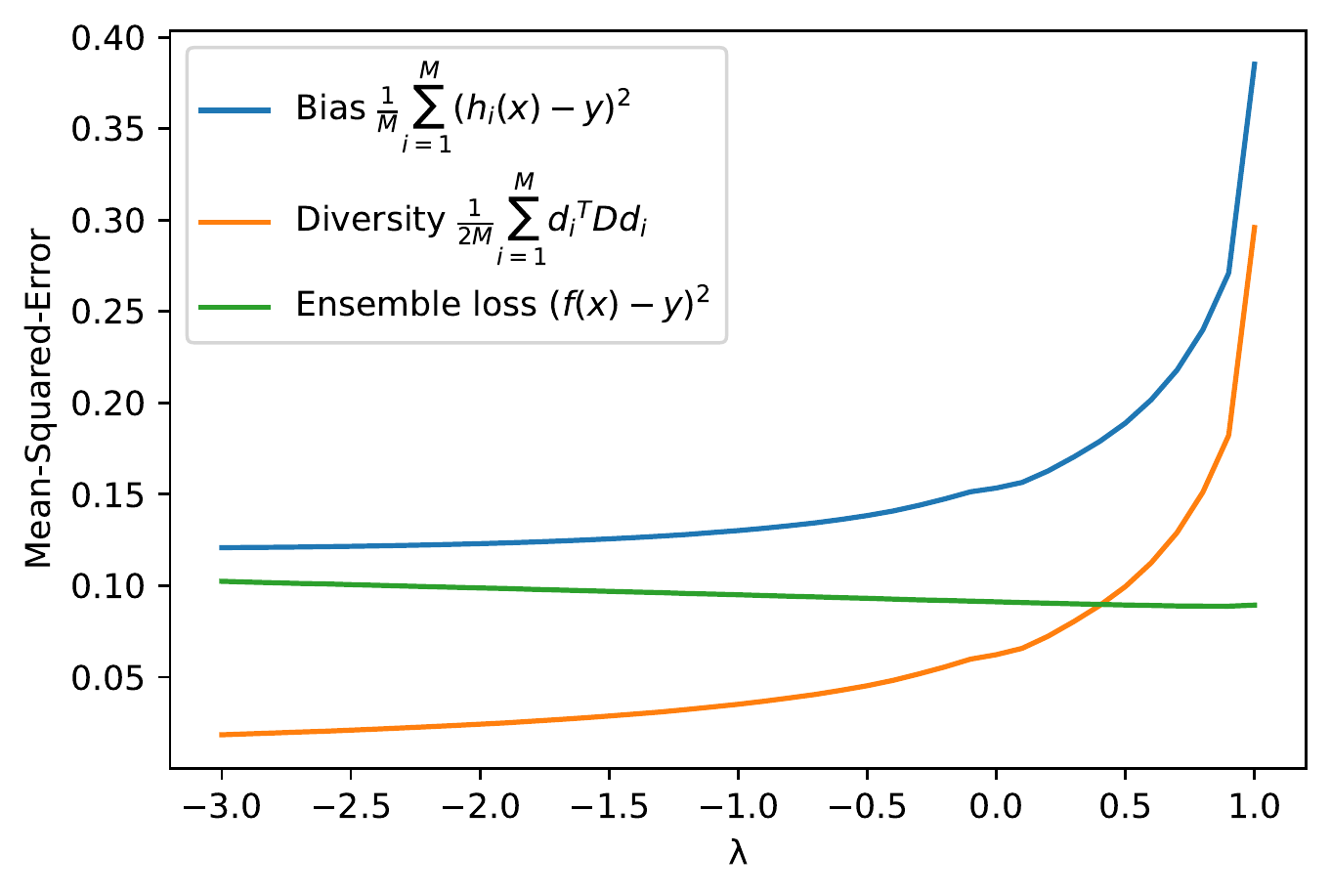}
    \includegraphics[width=5.5cm, height=4cm]{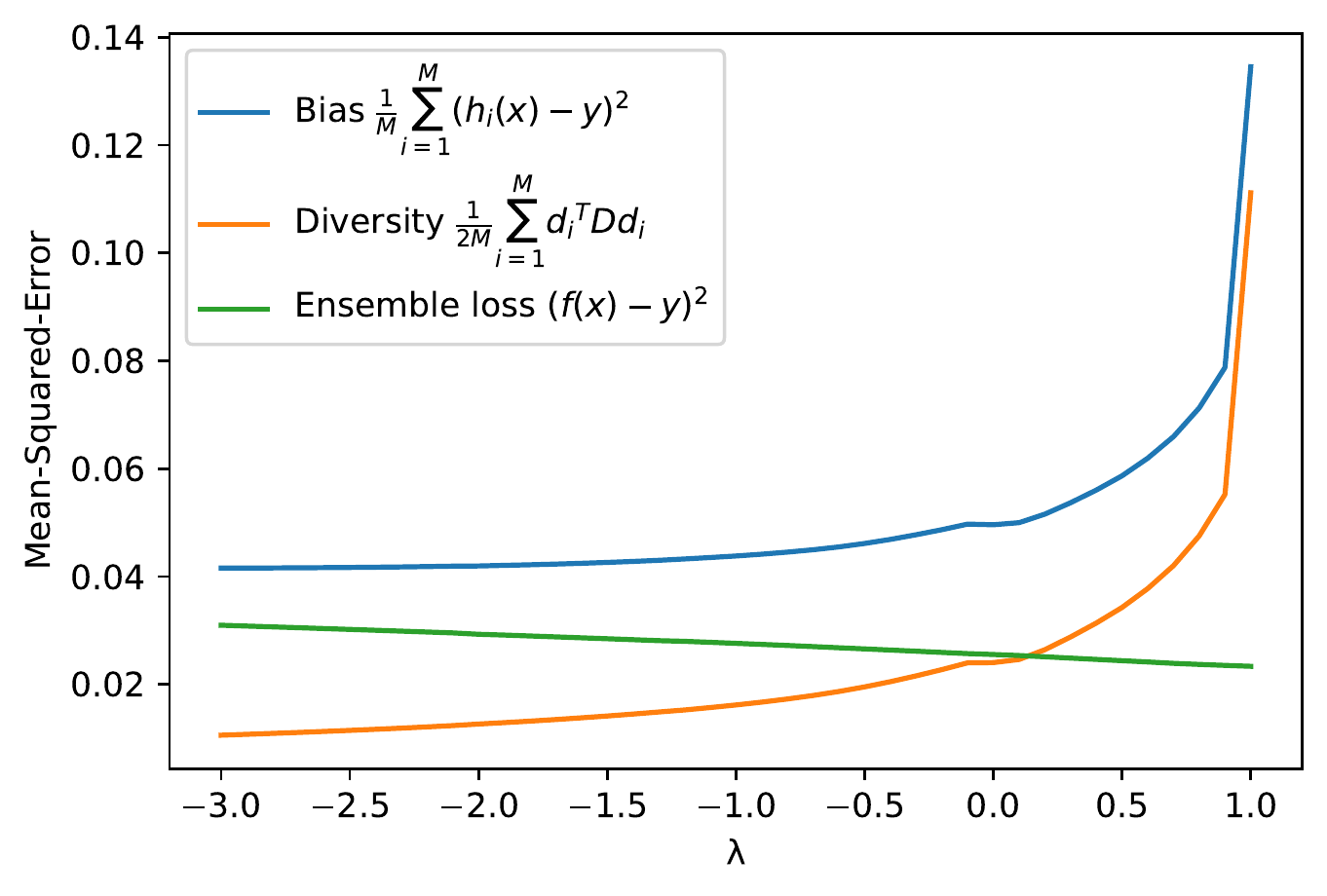}
    \caption{Mean Squared Error (zoom in).}\label{fig:3b}
\end{subfigure}
\begin{subfigure}[c]{0.30\textwidth}
    \includegraphics[width=5.5cm, height=4cm]{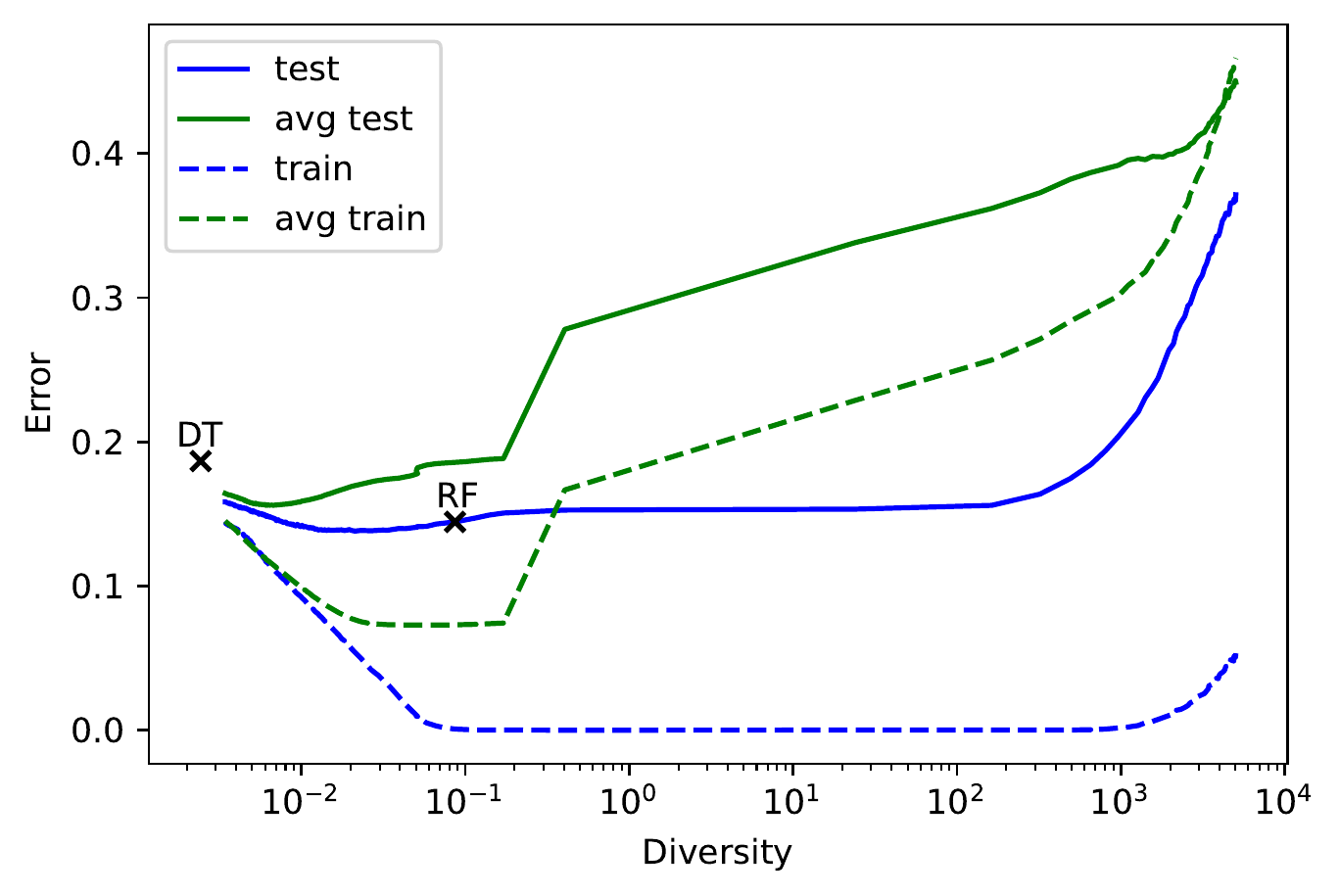}
    \includegraphics[width=5.5cm, height=4cm]{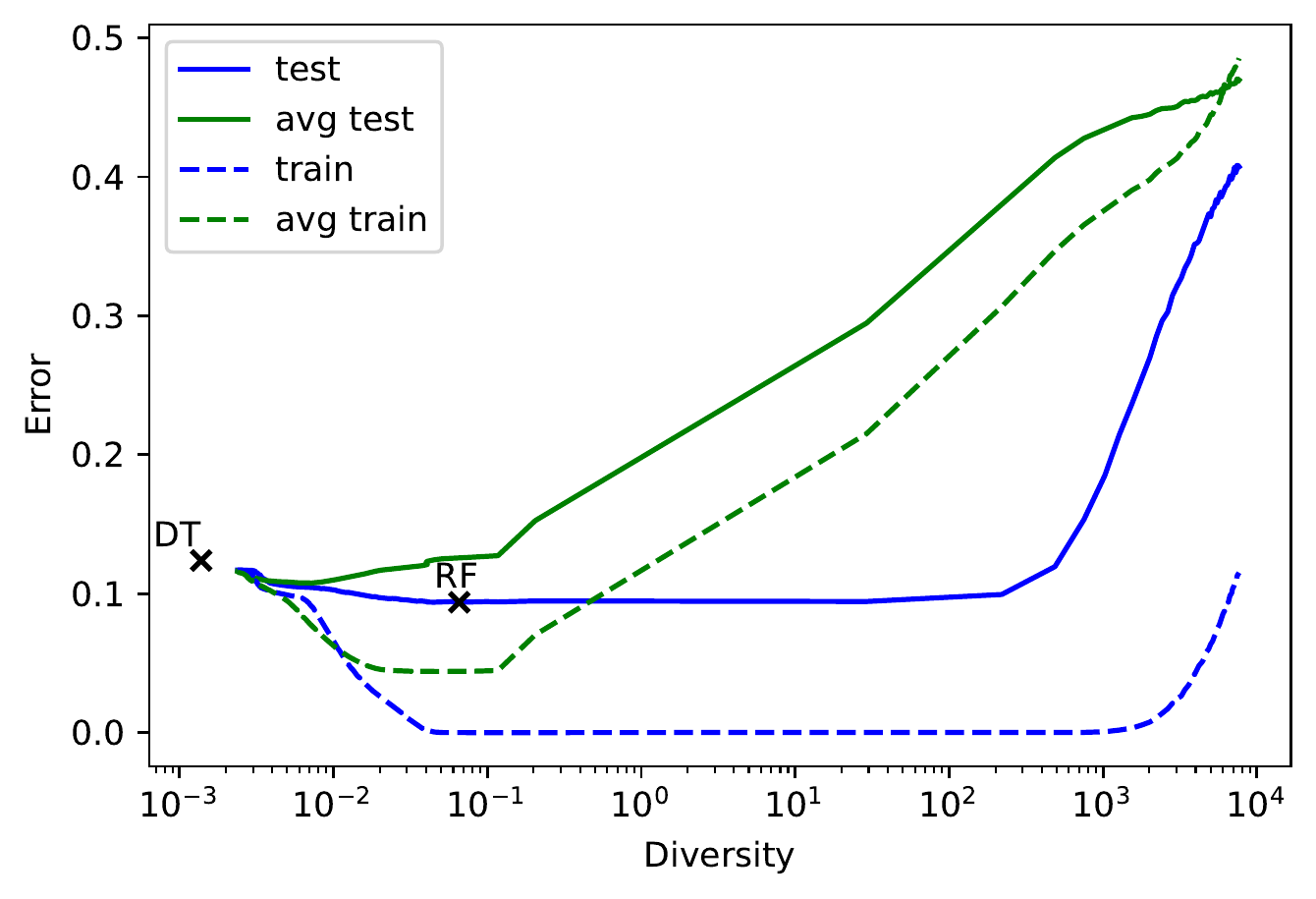}
    \includegraphics[width=5.5cm, height=4cm]{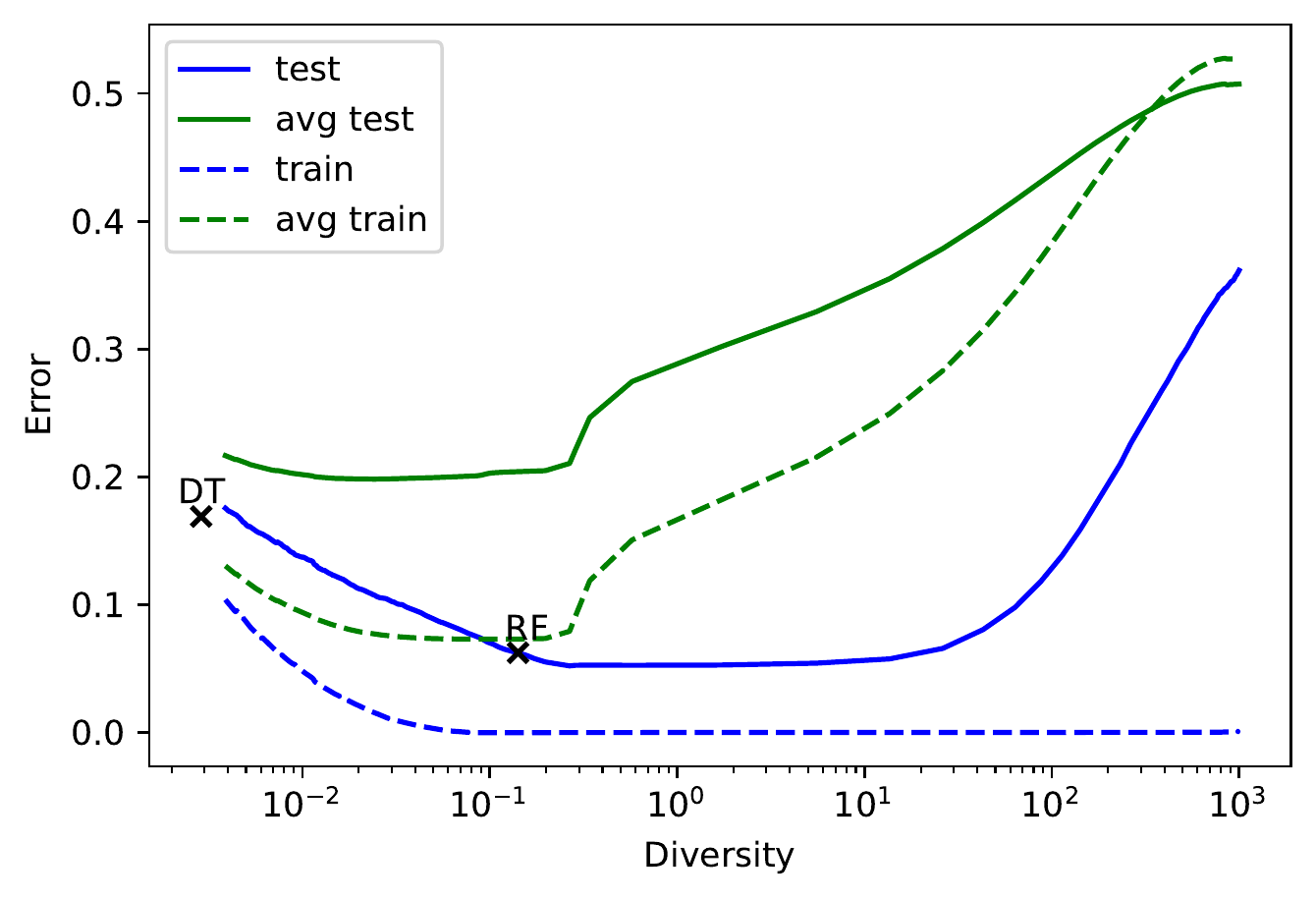}
    \includegraphics[width=5.5cm, height=4cm]{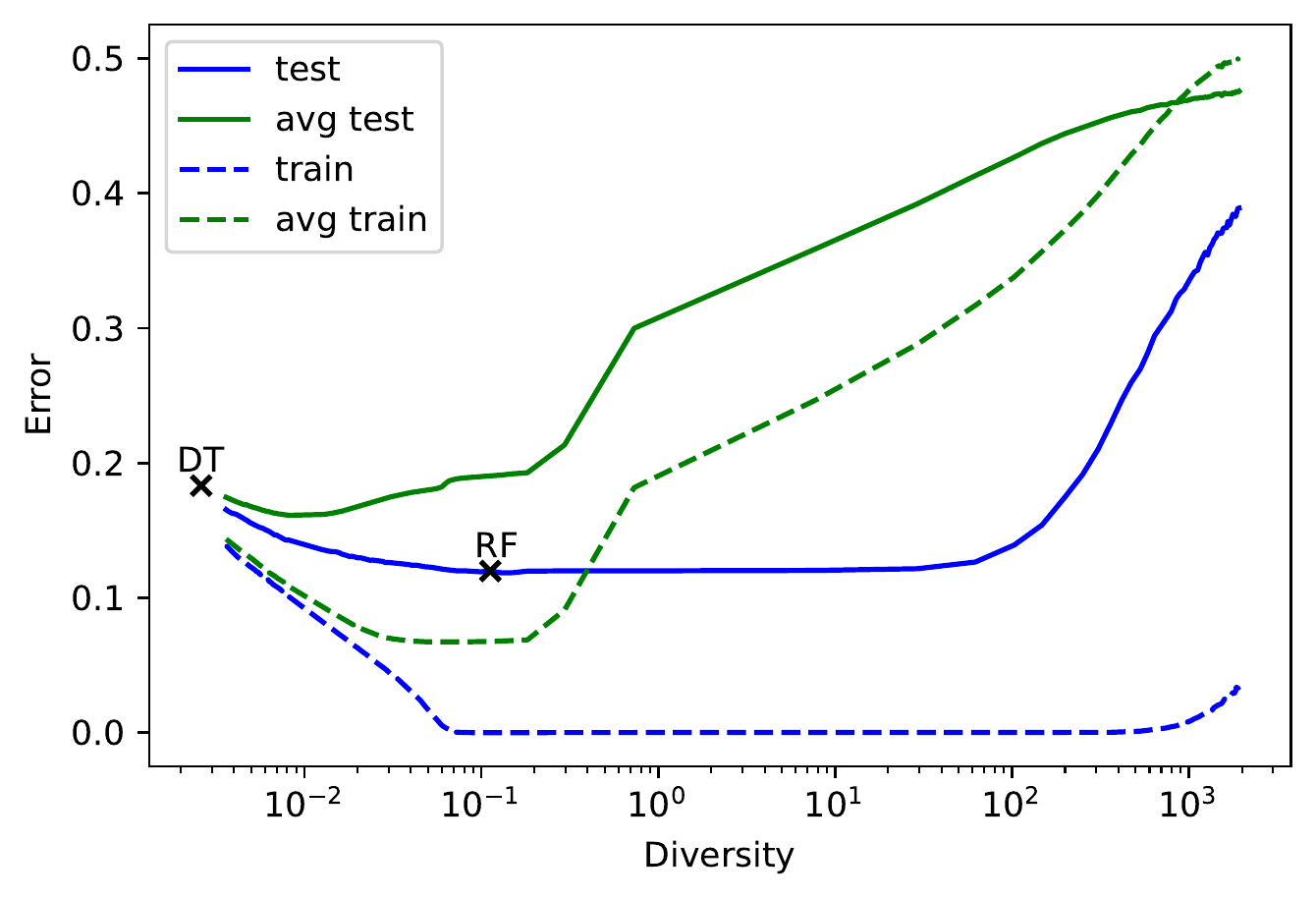}
    \includegraphics[width=5.5cm, height=4cm]{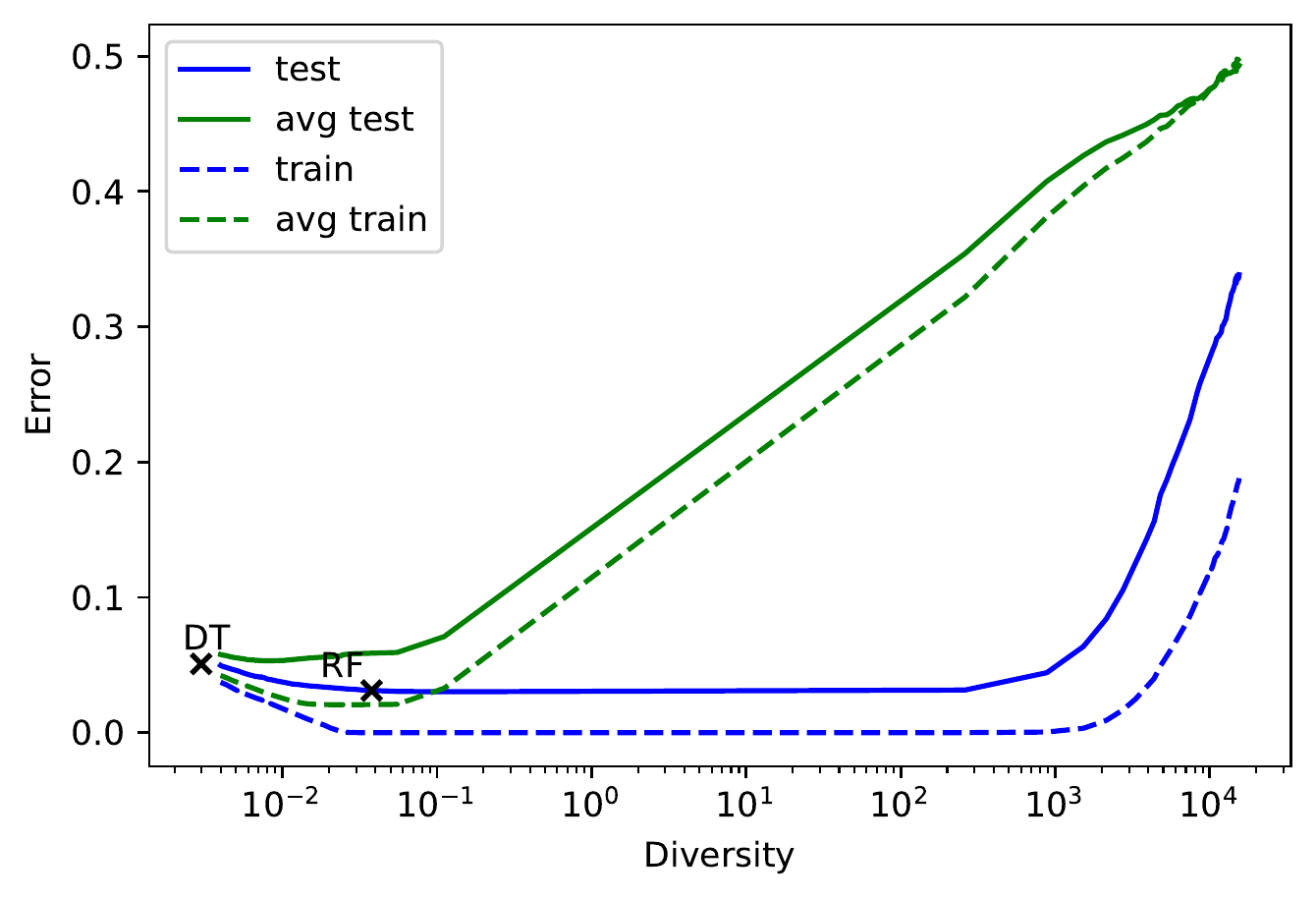}
    \caption{Test error.}\label{fig:3c}
\end{subfigure}
\caption{Mean-Squared error (first column, second column) over different $\lambda$ values and the test and train error (third column) over different diversities. Results are averaged over a 5-fold cross validation. Best viewed in color.} \label{fig:H3}
\end{figure*}

As seen in Figure \ref{fig:3a} and Figure \ref{fig:3b} our NCForest method indeed allows for a fine control over the diversity in the ensemble. Increasing $\lambda$ from $-20$ to $1.0$ leads to a larger bias and more diversity in the ensemble while the overall ensemble loss $\ell(f)$ nearly remains constant as expected. Increasing $\lambda > 1.0$ leads to a steep increase in both where the ensemble loss also increases because the diversity starts to dominate the optimization.

Looking at Figure \ref{fig:3c} we can see the average training and testing error of the trees in the ensemble as well as the test and train error of the entire ensemble. Again, dashed lines depict the train error and solid lins are the test error. Moreover, we marked the performance of a DT and a RF for better comparability\footnote{By definition a single DT does not have any diversity. For presentational purposes we assigned a non-zero diversity to it in order to not break the logarithmic axis.}. We can see two effects here. First, there seems to be a large region in which the diversity does not affect the ensemble error, but only the individual errors of the trees. In this region the performance of each individual trees changes, but the overall ensemble error remains nearly constant. The corresponding plots are akin to a bathtub: If the diversity is too small or too large, then performance suffers. Only in the middle of both extremes we find a good trade-off between both. For example, on the EEG dataset a diversity between $10^{-1} - 10^{2}$ seems to be a perfect trade-off with similar performance even though the average test error steeply increases for larger diversities.
Second we find that a RF seems to achieve a good balance between both quantities with its default parameters, although minor improvements are possible, e.g. on the adult or the EEG dataset.
We conclude that a larger diversity does not necessarily result in a better forest, but it can hurt the performance at some point. Likewise not enough diversity also leads to a bad performance and a good balance between both quantities must be achieved. Interestingly, there seems to be a comparably large region of similar performances where the \emph{exact} trade-off between bias and diversity does not matter. Hence, the diversity \emph{does} have some effect on the final model, but its effect might be overrated once this region is found by the algorithm. 

\section{Conclusion}
\label{sec:conclusion}

In this paper we revisited explanations for the success of Random Forests and showed that for most of these explanations an experiment can be constructed where they do not work or only offer little insight. First, given a proper definition of the complexity of forest, a RF does not exhibit a double descent, but rather has a single descent in which it simply fits the data better with increasing complexity. Second, a DT shows the `classic' u-shaped curve in which it starts to overfit at some point. We continued to show that a RF does not `inherit' these bad properties of a DT and retains a single-descent curve even if the RF is fitted on a ground-truth from an overfitted DT. Similar, a DT does not `inherit` the good properties of a RF, but also keeps its u-shaped overfitting curve when fitted on the ground-truth of a good, not overfitted RF. In all these experiments the Rademacher complexity did not accurately predict the performance of each classifier. In fact, DTs trained via data augmentation had a \emph{smaller} complexity than their RF counterparts but a much \emph{worse} test error. 
Hence, we argue that the training algorithm plays a crucial role in the performance of a model. We introduced Negative Correlation Forest (NCForest) which refines the leaf nodes of a forest to explicitly control the diversity among the trees. We hypothesized that a tree ensemble of diverse trees with sufficiently small bias should have a better generalization error than a homogeneous forest. And indeed there seems to be a bathtub-like correlation between the diversity and the test error. Having too few or having too much diversity hurts the performance, whereas the right balance of both quantities maximizes performance. Luckily there seems to be a comparably large region of different diversities which achieve this trade-off and a RF seems to find this region in most cases although some improvements are possible when this trade-off is further refined. 


\section*{Acknowledgments}
Part of the work on this paper has been supported by Deutsche Forschungsgemeinschaft (DFG) within the Collaborative Research Center SFB 876 "Providing Information by Resource-Constrained Analysis", DFG project number 124020371, SFB project A1, \url{http://sfb876.tu-dortmund.de}. Part of the work on this research has been funded by the Federal Ministry of Education and Research of Germany as part of the competence center for machine learning ML2R (01|18038A), \url{https://www.ml2r.de/}. Thank you Mirko Bunse and Lukas Pfahler for their helpful comments on this paper. 

\bibliography{literature}

\begin{thebibliography}{}

\bibitem[\protect\citeauthoryear{Asian, Yildiz, and
  Alpaydin}{2009}]{asian/etal/2009}
Asian, O.; Yildiz, O.~T.; and Alpaydin, E.
\newblock 2009.
\newblock Calculating the vc-dimension of decision trees.
\newblock In {\em 2009 24th International Symposium on Computer and Information
  Sciences},  193--198.
\newblock IEEE.

\bibitem[\protect\citeauthoryear{Belkin \bgroup et al\mbox.\egroup
  }{2019}]{belkin/etal/2019}
Belkin, M.; Hsu, D.; Ma, S.; and Mandal, S.
\newblock 2019.
\newblock Reconciling modern machine-learning practice and the classical
  bias--variance trade-off.
\newblock {\em Proceedings of the National Academy of Sciences}
  116(32):15849--15854.

\bibitem[\protect\citeauthoryear{Biau and Scornet}{2016}]{biau/Scornet/2016}
Biau, G., and Scornet, E.
\newblock 2016.
\newblock A random forest guided tour.
\newblock {\em Test} 25(2):197--227.

\bibitem[\protect\citeauthoryear{Biau}{2012}]{biau/2012}
Biau, G.
\newblock 2012.
\newblock Analysis of a random forests model.
\newblock {\em Journal of Machine Learning Research} 13(Apr):1063--1095.

\bibitem[\protect\citeauthoryear{Breiman \bgroup et al\mbox.\egroup
  }{1984}]{breiman/etal/1984}
Breiman, L.; Friedman, J.; Stone, C.; and Olshen, R.
\newblock 1984.
\newblock {\em Classification and Regression Trees}.
\newblock Taylor \& Francis.

\bibitem[\protect\citeauthoryear{Breiman}{2000}]{Breiman/2000}
Breiman, L.
\newblock 2000.
\newblock Some infinity theory for predictor ensembles.
\newblock Technical report, Technical Report 579, Statistics Dept. UCB.

\bibitem[\protect\citeauthoryear{Brown, Wyatt, and
  Tino}{2005}]{Brown/etal/2005}
Brown, G.; Wyatt, J.~L.; and Tino, P.
\newblock 2005.
\newblock {Managing Diversity in Regression Ensembles}.
\newblock {\em Jmlr}  1621--1650.

\bibitem[\protect\citeauthoryear{Buschj{\"a}ger, Pfahler, and
  Morik}{2020}]{buschjager/etal/2020}
Buschj{\"a}ger, S.; Pfahler, L.; and Morik, K.
\newblock 2020.
\newblock Generalized negative correlation learning for deep ensembling.
\newblock {\em arXiv preprint arXiv:2011.02952}.

\bibitem[\protect\citeauthoryear{Cortes, Mohri, and
  Syed}{2014}]{cortes/etal/2014}
Cortes, C.; Mohri, M.; and Syed, U.
\newblock 2014.
\newblock Deep boosting.
\newblock In {\em International conference on machine learning},  1179--1187.

\bibitem[\protect\citeauthoryear{Denil, Matheson, and
  De~Freitas}{2014}]{denil/2014}
Denil, M.; Matheson, D.; and De~Freitas, N.
\newblock 2014.
\newblock Narrowing the gap: Random forests in theory and in practice.
\newblock In {\em International conference on machine learning (ICML)}.

\bibitem[\protect\citeauthoryear{Geman, Bienenstock, and
  Doursat}{1992}]{Geman/etal/1992}
Geman, S.; Bienenstock, E.; and Doursat, R.
\newblock 1992.
\newblock {Neural Networks and the Bias/Variance Dilemma}.

\bibitem[\protect\citeauthoryear{Germain \bgroup et al\mbox.\egroup
  }{2015}]{germain/etal/15}
Germain, P.; Lacasse, A.; Laviolette, F.; March, M.; and Roy, J.-F.
\newblock 2015.
\newblock Risk bounds for the majority vote: From a pac-bayesian analysis to a
  learning algorithm.
\newblock {\em Journal of Machine Learning Research} 16(26):787--860.

\bibitem[\protect\citeauthoryear{Koltchinskii and
  Panchenko}{2002}]{Koltchinskii/Panchenko/2002}
Koltchinskii, V., and Panchenko, D.
\newblock 2002.
\newblock Empirical margin distributions and bounding the generalization error
  of combined classifiers.
\newblock {\em The Annals of Statistics} 30(1):1--50.

\bibitem[\protect\citeauthoryear{Leboeuf, LeBlanc, and
  Marchand}{2020}]{leboeuf/etal/2020}
Leboeuf, J.-S.; LeBlanc, F.; and Marchand, M.
\newblock 2020.
\newblock Decision trees as partitioning machines to characterize their
  generalization properties.
\newblock {\em Advances in Neural Information Processing Systems} 33.

\bibitem[\protect\citeauthoryear{Mansour}{1997}]{mansour/1997}
Mansour, Y.
\newblock 1997.
\newblock Pessimistic decision tree pruning based on tree size.
\newblock In {\em Proceedings of the Fourteenth International Conference on
  Machine Learning},  195--201.

\bibitem[\protect\citeauthoryear{Markowitz}{1952}]{Markowitz/1952}
Markowitz, H.
\newblock 1952.
\newblock {The Utility of Wealth}.
\newblock {\em Journal of Political Economy}.

\bibitem[\protect\citeauthoryear{Oshiro, Perez, and
  Baranauskas}{2012}]{oshiro/etal/2012}
Oshiro, T.~M.; Perez, P.~S.; and Baranauskas, J.~A.
\newblock 2012.
\newblock How many trees in a random forest?
\newblock In {\em International workshop on machine learning and data mining in
  pattern recognition},  154--168.
\newblock Springer.

\bibitem[\protect\citeauthoryear{Paszke \bgroup et al\mbox.\egroup
  }{2019}]{Paszke/etal/2019}
Paszke, A.; Gross, S.; Massa, F.; Lerer, A.; Bradbury, J.; Chanan, G.; Killeen,
  T.; Lin, Z.; Gimelshein, N.; Antiga, L.; Desmaison, A.; K{\"{o}}pf, A.; Yang,
  E.; DeVito, Z.; Raison, M.; Tejani, A.; Chilamkurthy, S.; Steiner, B.; Fang,
  L.; Bai, J.; and Chintala, S.
\newblock 2019.
\newblock {PyTorch: An imperative style, high-performance deep learning
  library}.
\newblock In {\em Advances in Neural Information Processing Systems}.

\bibitem[\protect\citeauthoryear{Pedregosa and
  others}{2011}]{Pedregosa/etal/2001}
Pedregosa, F., et~al.
\newblock 2011.
\newblock Scikit-learn: Machine learning in {P}ython.
\newblock {\em Journal of Machine Learning Research} 12:2825--2830.

\bibitem[\protect\citeauthoryear{Quinlan}{1986}]{quinlan/1986}
Quinlan, J.~R.
\newblock 1986.
\newblock Induction of decision trees.
\newblock {\em Machine learning} 1(1):81--106.

\bibitem[\protect\citeauthoryear{Shalev-Shwartz and
  Ben-David}{2014}]{shalevshwartz/bendavid/2014}
Shalev-Shwartz, S., and Ben-David, S.
\newblock 2014.
\newblock {\em Understanding machine learning: From theory to algorithms}.
\newblock Cambridge university press.

\bibitem[\protect\citeauthoryear{Webb \bgroup et al\mbox.\egroup
  }{2019}]{Webb/etal/2019}
Webb, A.~M.; Reynolds, C.; Iliescu, D.-A.; Reeve, H.; Lujan, M.; and Brown, G.
\newblock 2019.
\newblock {Joint Training of Neural Network Ensembles}.

\bibitem[\protect\citeauthoryear{Zhang \bgroup et al\mbox.\egroup
  }{2021}]{zhang/etal/2021}
Zhang, C.; Bengio, S.; Hardt, M.; Recht, B.; and Vinyals, O.
\newblock 2021.
\newblock Understanding deep learning (still) requires rethinking
  generalization.
\newblock {\em Communications of the ACM} 64(3):107--115.

\end{thebibliography}
\bibliographystyle{aaai}

\end{document}